\documentclass{article}

\usepackage[preprint]{neurips_2025}

\usepackage{multirow}
\usepackage{multicol}
\usepackage{xcolor}
\usepackage{multirow}
\usepackage{multicol}
\usepackage{colortbl}
\usepackage{verbatim}
\usepackage{graphicx}
\usepackage{booktabs}
\usepackage{multirow}
\usepackage{listings}
\usepackage{xcolor}
\usepackage[T1]{fontenc}
\usepackage[utf8]{inputenc}
\usepackage{array}
\usepackage{caption}

\usepackage[
    colorlinks=true,     
    linkcolor=blue,     
    citecolor=blue,     
    urlcolor=blue        
]{hyperref}
\usepackage{amsmath}
\usepackage{cleveref}
\usepackage{enumitem}
\usepackage{pifont}
\setlist{
    topsep=0.3ex,      
    itemsep=0.2ex,     
    parsep=0pt,        
    leftmargin=1.5em,  
    labelindent=0.5em  
}
\usepackage{amsfonts}

\usepackage{titlesec}
\titlespacing*{\section}{0pt}{0.6ex}{0.3ex}
\titlespacing*{\subsection}{0pt}{0.3ex}{0.2ex}
\titlespacing*{\subsubsection}{0pt}{0.1ex}{0.1ex}
\titlespacing*{\paragraph}{0ex}{0.1ex}{1ex}

\title{Rex-Thinker: Grounded Object Referring via Chain-of-Thought Reasoning}

\begin{document}

\author{Qing Jiang$^{1,2*}$ , Xingyu Chen$^{3*}$ , Zhaoyang Zeng$^{1}$ ,  Junzhi Yu$^{3}$ , Lei Zhang$^{1,2\dagger}$ \\
$^1$International Digital Economy Academy (IDEA) \\
$^2$South China University of Technology \\ 
$^3$Peking University\\
{\tt\small \{jiangqing, chenxingyu, leizhang\}@idea.edu.cn} \\
\url{https://rexthinker.github.io/}
}

\makeatletter
\def\@makefnmark{} 
\makeatother
\footnotetext{$^*$Equal contributions, work done during internship or academic visit at IDEA. $^\dagger$Corresponding author.}

\maketitle

\begin{abstract}

Object referring aims to detect all objects in an image that match a given natural language description. We argue that a robust object referring model should be grounded, meaning its predictions should be both explainable and faithful to the visual content. Specifically, it should satisfy two key properties: \textbf{1) Verifiable}, by producing interpretable reasoning that justifies its predictions and clearly links them to visual evidence; and \textbf{2) Trustworthy}, by learning to abstain when no object in the image satisfies the given expression. However, most methods treat referring as a direct bounding box prediction task, offering limited interpretability and struggling to reject expressions with no matching object.
In this work, we propose Rex-Thinker, a model that formulates object referring as an explicit Chain-of-Thought (CoT) reasoning task. Given a referring expression, we first identify all candidate object instances corresponding to the referred object category. Rex-Thinker then performs step-by-step reasoning over each candidate to assess whether it matches the given expression, before making a final prediction.
To support this paradigm, we construct a large-scale CoT-style referring dataset named HumanRef-CoT by prompting GPT-4o on the HumanRef dataset. Each reasoning trace follows a structured planning, action, and summarization format, enabling the model to learn decomposed, interpretable reasoning over object candidates. We then train Rex-Thinker in two stages: a cold-start supervised fine-tuning phase to teach the model how to perform structured reasoning in our defined CoT format, followed by GRPO-based reinforcement learning to further improve accuracy and generalization. Experiments show that our CoT-based approach outperforms standard baselines in both precision and interpretability on in-domain evaluation, while also demonstrating improved ability to reject hallucinated outputs and strong generalization in out-of-domain settings. Code is available at \url{https://github.com/IDEA-Research/Rex-Thinker}.
\end{abstract}

\section{Introduction}


Object Referring, also known as Referring Expression Comprehension (REC)~\cite{qiao2020referring,Datasets:REFCOCO,Datasets:REFCOCOG,kazemzadeh2014referitgame,zhang2019referring,luo2020multi,yu2018mattnet,yu2018mattnet,yang2019dynamic,liao2020real}, aims to predict the bounding boxes for objects in an image that match a given natural language description, which may refer to visual attributes, spatial relations, or interactions. This task has broad applications; however, compared to standard open-vocabulary object detection~\cite{jiang2025t, ren2024dino, ren2024grounding, liu2024grounding, yao2022detclip, jiang2024t,jiang2023t, li2024visual, ren2024dino, li2022grounded,cheng2024yolo,minderer2022simple,zareian2021open,wu2023aligning}, REC is significantly more challenging, as it requires both fine-grained visual grounding and more complicated language understanding.

Benefiting from the strong language comprehension capabilities of large language models (LLMs), multimodal large language models (MLLMs) have demonstrated impressive performance on this task. There are mainly two paradigms: one treats bounding box coordinates as text tokens and predicts them directly~\cite{chen2023shikra,you2023ferret,zhang2024ferret,wang2023cogvlm,zhan2025griffon,zhan2024griffon,bai2025qwen2, wu2024deepseek, chen2024expanding, VLM:MM1}, while the other adopts a retrieval-based strategy~\cite{ma2024groma, jiang2024chatrex, jiang2025referringperson}, where the model is given a set of candidate boxes and predicts the box indices that match the expression. Although both approaches have shown promising results, they remain fundamentally implicit, lacking interpretable reasoning steps that reveal how the model arrives at its final prediction. Furthermore, these models are prone to hallucination~\cite{jiang2025referringperson}, often producing outputs for objects that do not exist in the image, thereby limiting their reliability in real-world applications.

We argue that a robust referring system should be \textit{grounded}, i.e., its predictions must be both explainable and tightly linked to visual evidence. This requires two essential properties: \textbf{1) Verifiable}, by providing an explicit reasoning process that allows its decisions to be examined and traced to specific image regions; and \textbf{2) Trustworthy}, by minimizing hallucinated outputs and learning to reject when no object in the image satisfies the given description.
To meet these criteria, we draw inspiration from how humans naturally approach referring expressions. For example, when asked to locate ``the person wearing a blue shirt'', humans would typically first identify all people in the image, then examine each one to determine whether it matches the described attribute. This step-by-step approach reflects a grounded reasoning process, i.e., first localizing relevant object candidates, and then carefully verifying each one against the expression.

\begin{figure*}[t]\centering\vspace{-1em}
\includegraphics[width=1\linewidth]{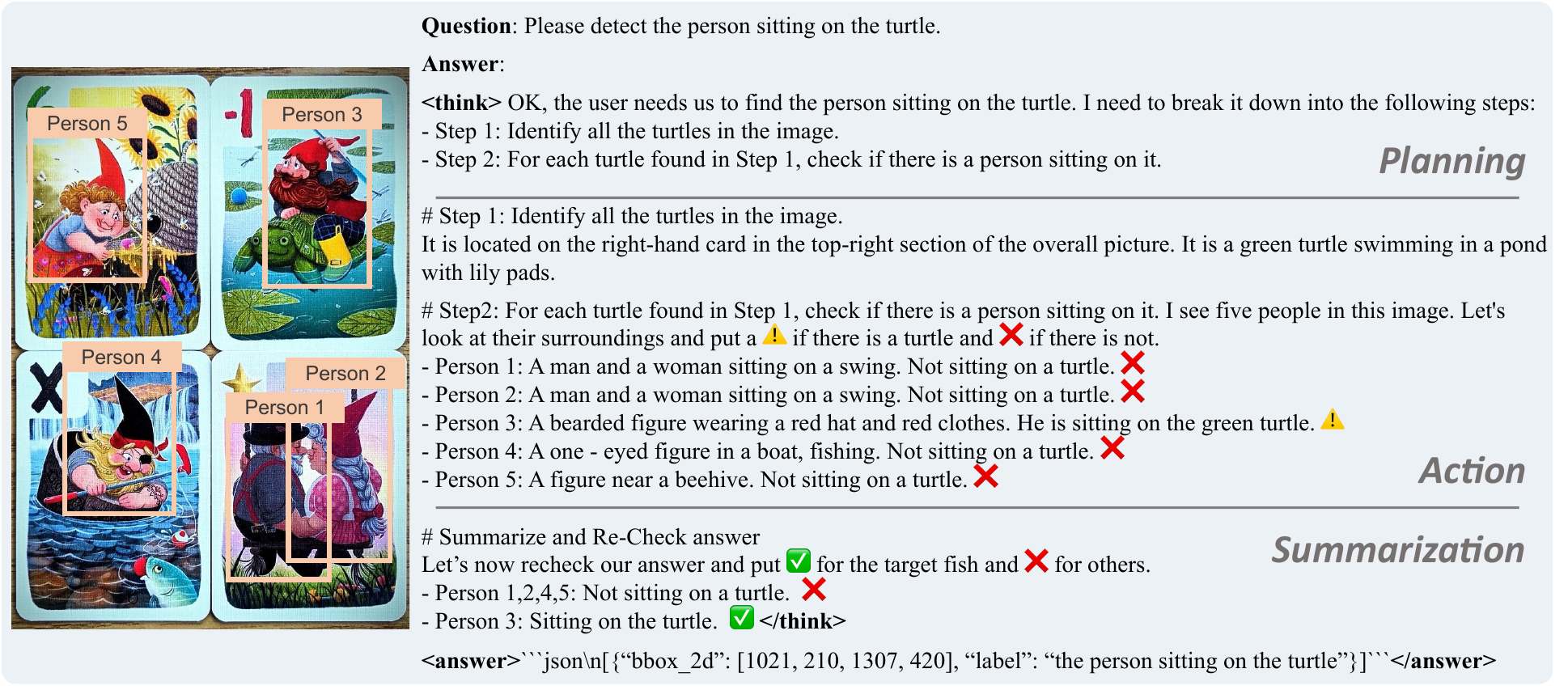}\vspace{-1mm}
\caption{An example of Rex-Thinker for object referring with CoT reasoning of planning (task decomposition), action (evaluating each candidate), and summarization (final decision). Each step is grounded in a specific hint box (as denoted in the left image), enabling interpretable predictions.}
\label{fig:teaser}
\vspace{-1em}
\end{figure*}

Motivated by this observation, we propose Rex-Thinker, an MLLM that performs object referring through explicit Chain-of-Thought (CoT) reasoning. Specifically, given an image and a referring expression, we first use an open-vocabulary object detector~\cite{liu2024grounding} to extract all candidate object boxes corresponding to the referred category. These candidate boxes, along with the image and the expression, are then passed into the model for step-by-step reasoning.
Rex-Thinker follows a structured CoT framework consisting of three key stages as shown in Figure~\ref{fig:teaser}:
\textbf{1) Planning}, where the model decomposes the referring expression into subgoals;
\textbf{2) Action}, where the model examines each candidate box to determine whether it satisfies its current subgoal;
\textbf{3) Summarization}, where it aggregates the intermediate decisions to produce the final prediction.
Following DeepSeek-R1~\cite{guo2025deepseek}, we instruct the model to place its reasoning steps within a \texttt{<think>}...\texttt{</think>} block and to output the final prediction inside a \texttt{<answer>}...\texttt{</answer>} block. This structured reasoning process not only improves interpretability, but also enables transparent and verifiable predictions, as each reasoning step is grounded in a specific candidate region in the image.

To support this CoT framework, we construct a CoT-style referring dataset named HumanRef-CoT, containing 90,824 samples generated by prompting GPT-4o~\cite{hurst2024gpt} on the HumanRef~\cite{jiang2025referringperson} dataset. Each example is annotated with a structured reasoning trace following the planning, action, and summarization paradigm, enabling explicit supervision for step-by-step reasoning. We train our model in two stages: a cold-start supervised fine-tuning phase to teach the model how to perform structured reasoning, followed by reinforcement learning (RL) based on Group Relative Policy Optimization (GRPO)~\cite{deepseekmath} to further improve accuracy and generalization. Experiments demonstrate that our CoT-based approach consistently outperforms direct coordinate prediction baselines. On the in-domain HumanRef benchmark, our model achieves state-of-the-art results with higher detection accuracy and significantly fewer hallucinated outputs, especially on rejection cases. In out-of-domain evaluations on RefCOCOg~\cite{Datasets:REFCOCOG}, the model trained only on HumanRef-CoT shows strong zero-shot generalization. Further fine-tuning with GRPO on RefCOCOg yields additional performance gains while preserving the model’s ability to perform grounded CoT reasoning across arbitrary object categories. To summarize, our contributions are threefold:
\vspace{-0.2em}
\begin{itemize}
    \item We formulate the grounded object referring task as a \textit{planning–action–summarization} problem, leveraging Chain-of-Thought reasoning to build a verifiable and trustworthy system.
    \item We introduce HumanRef-CoT, the first dataset for grounded object referring with step-by-step reasoning annotations, enabling the supervised training of model interpretability.
    \item We propose Rex-Thinker, a grounded object referring model trained via cold-start SFT and GRPO-based reinforcement learning. Rex-Thinker achieves SOTA performance on the HumanRef benchmark and demonstrates strong generalization on out-of-domain scenes and objects.
\end{itemize}

\section{Related Work}

\paragraph{MLLM-based Object Referring Methods.}
Recent progress in multimodal large language models (MLLMs)~\cite{gpt4v, bai2025qwen2, wu2024deepseek, chen2024expanding, alayrac2022flamingo, li2024aria, deitke2024molmo, agrawal2024pixtral, wang2024qwen2, li2024llava, li2025eagle, VLM:LLaVA, yangkptllm, zhu2025internvl3, chen2025eagle, steiner2024paligemma, guo2025seed1, chen2025eagle} has led to strong performance in referring expression comprehension. Existing approaches typically follow two paradigms. One line of work treats bounding box coordinates as textual tokens~\cite{chen2021pix2seq} and directly generates them during decoding~\cite{chen2023shikra, you2023ferret, wang2023cogvlm, zhan2025griffon, zhang2024ferret}. The other line formulates the task as retrieval~\cite{jiang2024chatrex, ma2024groma, jiang2025referringperson}, where a detector proposes candidate regions and the model selects the best-matching box indices based on the input expression. This decouples localization from semantic understanding and simplifies learning.

While both paradigms achieve strong results on standard benchmarks such as RefCOCO/+/g~\cite{Datasets:REFCOCOG, Datasets:REFCOCO}, they face key limitations: a lack of interpretability and an inability to abstain when no object in the image matches the expression~\cite{jiang2025referringperson}. To address this, we introduce a Chain-of-Thought reasoning framework that enables step-by-step evaluation over candidate boxes. This improves interpretability, reduces hallucinations, and grounds the model’s predictions in the input image.

\paragraph{Reasoning-based LLMs and MLLMs.}
Recent work in large language models~\cite{jaech2024openai, guo2025deepseek, team2025kimi, muennighoff2025s1simpletesttimescaling, xiang20252reasoningllmslearning, xiong2025selfrewardingcorrectionmathematicalreasoning, chu2025sft, openai2025competitiveprogramminglargereasoning} has demonstrated that reasoning ability can be significantly enhanced through Chain-of-Thought (CoT) training or reinforcement learning-based post-training. OpenAI o1~\cite{jaech2024openai} model demonstrates that inference-time scaling can greatly enhance performance on complex tasks like math and coding. DeepSeek-R1~\cite{guo2025deepseek} introduces GRPO~\cite{deepseekmath} as a post-training method to improve reasoning without requiring costly critic models. 

In the multimodal domain,   efforts such as LLaVA-CoT~\cite{xu2024llava} and LlamaV-o1~\cite{thawakar2025llamav} aim to enhance reasoning by constructing CoT-style data or employing multi-step curriculum learning, without relying on reinforcement learning. More recently, inspired by DeepSeek-R1~\cite{guo2025deepseek}, a growing number of works adopt GRPO-based post-training to endow MLLMs with reasoning capabilities. GRPO has been successfully applied to enhance multimodal reasoning across a wide range of domains, including mathematical problem solving~\cite{yang2025r1, peng2025skywork, zhang2025r1, deng2025openvlthinker, wei2025skywork}, video understanding~\cite{feng2025video, liao2025improved}, and perception tasks~\cite{liu2025seg, liu2025visual, ma2025deepperception, shen2025vlm, yu2025perception} such as object detection, segmentation, and referring expression comprehension. Following the DeepSeek-R1 paradigm, we first fine-tune Rex-Thinker on structured CoT data to teach the model how to perform grounded object reasoning. GRPO is then applied in a second stage to further improve accuracy and generalization.
\section{Chain-of-Thought Reasoning Referring Data}

High-quality supervision is critical for teaching the model to reason explicitly. To this end, we develop a data engine that generates structured referring annotations aligned with our Chain-of-Thought formulation. In this section, we introduce the design principles of our CoT reasoning structure and present the data construction pipeline that transforms existing REC annotations into step-by-step reasoning traces suitable for supervised training.

\subsection{CoT Formulation}
The core idea behind our CoT formulation for REC is to transform the task into a structured, grounded reasoning process over a set of candidate objects. Rather than directly predicting the referred object, the model evaluates each candidate in sequence, guided by input box hints that localize specific regions in the image. We decompose this CoT process into three key stages:
\vspace{-0.2em}
\begin{itemize}
\item \textbf{Planning:} The model analyzes the complexity of the referring expression and determines how many reasoning steps are needed. For simple expressions, it may plan a single step to directly match an attribute such as color or size. For more complex expressions, the model generates a multi-step plan, where each step focuses on resolving a specific sub-aspect.
\item \textbf{Action:} Based on the reasoning plan, the model checks whether each candidate region, grounded via its input box hint, satisfies the current subgoal. This makes the reasoning clear and directly tied to specific regions in the image.
\item \textbf{Summarization:} Finally, the model reviews the reasoning results across all steps and determines which objects best match the overall expression and outputs the final prediction.
\end{itemize}
This structured CoT process improves both interpretability and verifiability. Each candidate is evaluated corresponding to the input box hints, allowing every reasoning step to be explicitly grounded to a specific region of the image. This makes the model’s decisions transparent and easy to trace. Additionally, breaking complex expressions into sub-tasks enables step-by-step reasoning, which enhances accuracy and reflects how humans typically process such tasks.

\begin{figure*}[t]\centering\vspace{-2em}
\includegraphics[width=1\linewidth]{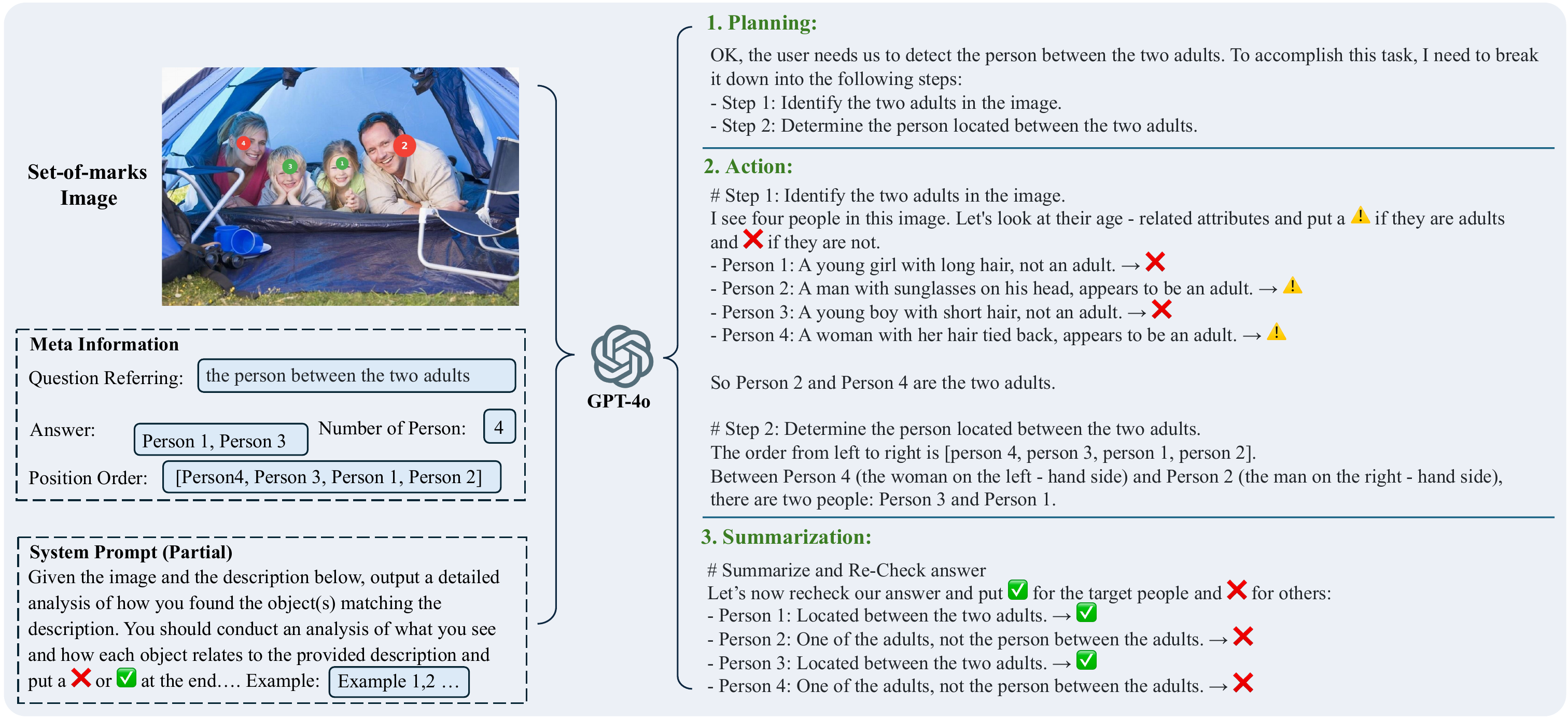}\vspace{-1mm}
\caption{Overview of the proposed CoT reasoning referring data engine. We prompt GPT-4o to generate a three-step CoT reasoning process, including planning, action, and summarization.}
\label{fig:data_engine}
\vspace{-2em}
\end{figure*}

\subsection{Data Engine Pipeline}
Building on the structured CoT formulation, we develop a data engine that leverages GPT-4o~\cite{hurst2024gpt} to generate high-quality CoT annotations tailored to the referring task.

\subsubsection{Data Acquisition}
We construct our CoT dataset based on HumanRef~\cite{jiang2025referringperson}, a recently proposed dataset specifically designed for REC in human-centric scenarios. Unlike prior REC datasets such as RefCOCO/+/g~\cite{Datasets:REFCOCOG, Datasets:REFCOCO}, HumanRef emphasizes multi-instance referring expressions, where a single expression may refer to multiple target persons. It also categorizes expressions into six distinct subsets: attribute, position, interaction, reasoning, celebrity recognition, and rejection. Since the HumanRef dataset provides all person boxes in an image, it can be directly used in our CoT annotation pipeline.

\subsubsection{GPT-4o Annotation}
To generate high-quality CoT annotations, we employ in-context prompting with GPT-4o~\cite{hurst2024gpt} as shown in Figure~\ref{fig:data_engine}. Given an image and the bounding boxes of all persons within it, we apply the Set-of-Mark~\cite{yang2023set} strategy: each individual is labeled with an indexed visual marker, where ground-truth targets are marked in green and others in red. This design grounds the answer and guides GPT-4o to reason along the correct path. The prompt includes three key components: 1) meta-information such as the referring question, the number of people, their left-to-right spatial order, and the correct answer; 2) a system prompt specifying the desired planning–action–summarization structure; and 3) several in-context examples written by humans to illustrate the expected reasoning format. In essence, we provide GPT-4o with both the referring expression and its ground-truth answer, and prompt it to generate step-by-step reasoning in our CoT format. To ensure annotation quality, we retain only examples where GPT-4o’s final prediction matches the ground-truth label.

We construct a total of 90,824 high-quality CoT annotations based on the HumanRef dataset, which we refer to as HumanRef-CoT. This diverse and large-scale dataset serves as the foundation for both our initial cold-start SFT and GRPO-based post-training.
\section{Method}
To leverage the CoT-style referring data, we present Rex-Thinker, a retrieval-based model that performs object referring through explicit Chain-of-Thought reasoning.

\subsection{Retrieval-based Object Referring}

\begin{figure*}[t]\centering\vspace{-2em}
\includegraphics[width=1\linewidth]{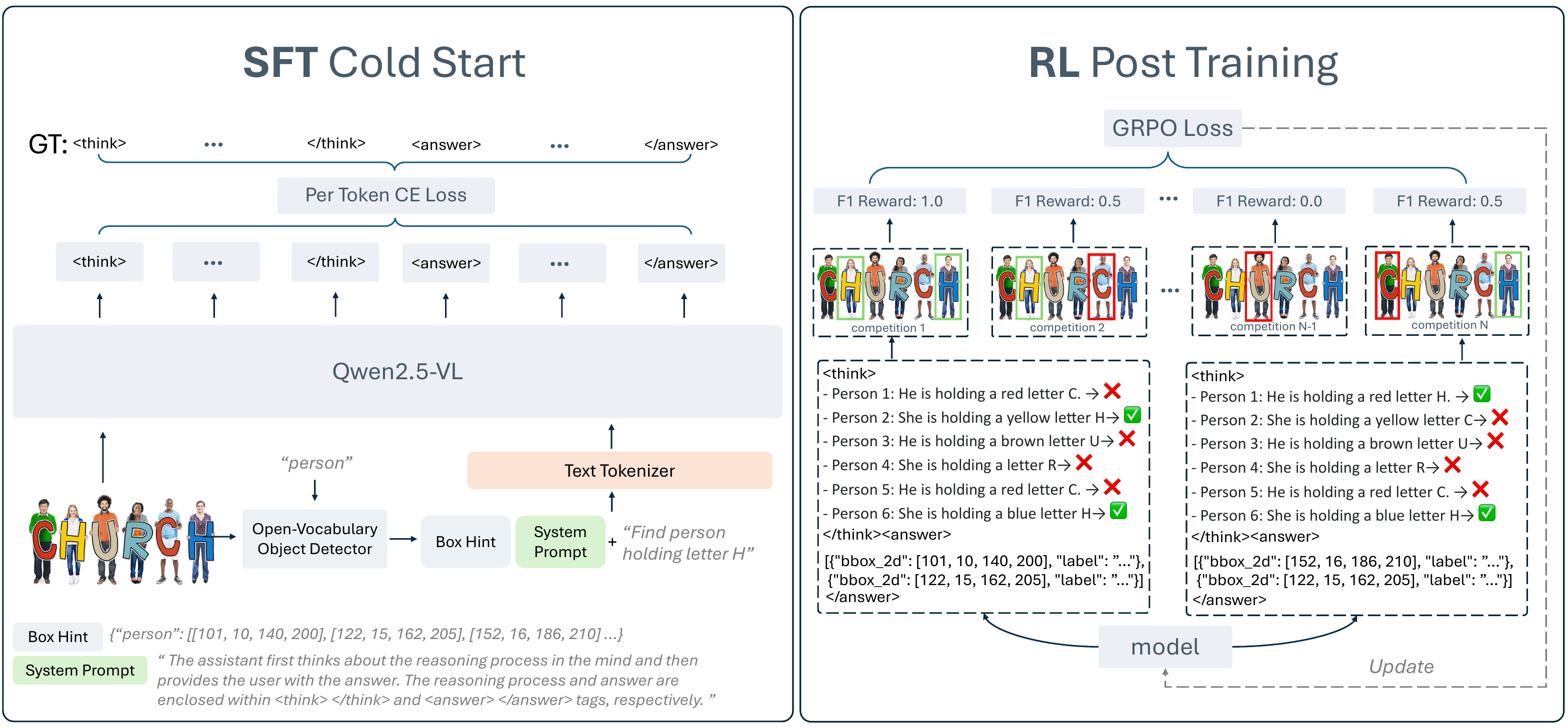}\vspace{-2mm}
\caption{Overview of the Rex-Thinker architecture and our two-stage training methods}
\label{fig:model}
\vspace{-1em}
\end{figure*}

To support explicit Chain-of-Thought (CoT) reasoning, we reformulate referring expression comprehension as a retrieval-based task. As shown in Figure~\ref{fig:model}, rather than directly regressing bounding boxes, we first use an open-vocabulary detector~\cite{liu2024grounding} to extract a set of candidate object boxes corresponding to the referred object category. These candidate boxes serve as \textit{box hints} to guide both the reasoning path and final decision of the model. This retrieval-based formulation brings two key advantages. First, during the reasoning phase, the model evaluates each candidate region in the order they appear in the input box hints (e.g., ``Person 1'' corresponds to the first input box). This alignment ensures that each step in the CoT trace is explicitly grounded to a specific region in the image, making the reasoning process interpretable and visually verifiable. Second, during the prediction phase, the model can directly select from the input box hints when producing the final output, thereby easing the challenge of precise coordinate regression.

\begin{table*}[t]
\centering\small
  \resizebox{1.0\linewidth}{!}{
    \begin{tabular}{@{}p{\textwidth}@{}}
\toprule
\texttt{<image>}. A conversation between User and Assistant. The user asks a question, and the Assistant solves it. The assistant first thinks about the reasoning process in the mind and then provides the user with the answer. The reasoning process and answer are enclosed within \texttt{<think>} \texttt{</think>} and \texttt{<answer>} \texttt{</answer>} tags, respectively, i.e., \texttt{<think>} reasoning process here \texttt{</think>} \texttt{<answer>} answer here \texttt{</answer>}. Hint: Object and its coordinates in this image: \textcolor{red}{Box Hint}. User: Locate \textcolor{red}{Referring}. Assistant: \\
\bottomrule
\end{tabular}}\vspace{-1mm}
\caption{Prompt Template for Rex-Thinker. \textcolor{red}{Box Hint} and \textcolor{red}{Referring} will be replaced with the input candidate boxes and the referring expression, respectively.}
\label{tab:prompt}
\vspace{-2em}
\end{table*} 

We build Rex-Thinker on top of Qwen2.5-VL-7B~\cite{bai2025qwen2}, preserving its original architecture and using JSON-format bounding box coordinates as the final output. The model input includes the image, the box hint, the referring expression, and a system prompt that guides the reasoning process. We adopt a similar input prompt format in DeepSeek-R1~\cite{guo2025deepseek} as shown in Table~\ref{tab:prompt}.

\subsection{Training}

Following DeepSeek-R1~\cite{guo2025deepseek}, we adopt a two-stage training strategy consisting of supervised fine-tuning for cold start and GRPO-based reinforcement learning for post-training.

\subsubsection{SFT Cold Start}

We begin by fine-tuning Rex-Thinker on the HumanRef-CoT dataset to instill the ability to perform structured reasoning following our defined planning, action, and summarization format. We apply cross-entropy loss at the token level to both the reasoning trace and the final answer, providing strong supervision across the entire generation process. This stage teaches the model how to reason step-by-step in a CoT manner and also how to utilize the provided box hints to guide its final predictions.
\subsubsection{GRPO Post Training}

While SFT teaches the model to follow our grounded CoT format, its strict token-level supervision may constrain the model to explore alternative reasoning traces and generalize beyond the training data. To enhance generalization beyond the limitations of supervised learning, we employ GRPO-based~\cite{deepseekmath} reinforcement learning for post-training. GRPO optimizes model performance by 1) sampling multiple candidate responses for each question and 2) selectively reinforcing responses that achieve higher task-level rewards.

In our setting, given an image and a referring expression $(I,x)$, the model generates a group of $G$ complete responses ${o_1, o_2, \dots, o_G}$ from the current model $\pi_\theta$. Each response contains a full reasoning trace and a final predicted bounding box set. For each $o_i$, we compute a scalar reward $r_i$ (detailed in Section~\ref{sec:reward-modeling}), and normalize these rewards to estimate group-relative advantages:
\begin{equation}
A_i = (r_i - \mathrm{mean}({r_1, \dots, r_G}))/\mathrm{std}({r_1, \dots, r_G}).
\end{equation}

Define the token-level advantage estimates $\hat{A}_{i,t} = A_i$, and the importance ratio at each decoding step as follows,
\begin{equation}
\rho_{i,t} = \frac{\pi_\theta(o_{i,t} \mid (I,x), x, o_{i,<t})}{\pi_{\theta_{\text{old}}}(o_{i,t} \mid (I,x), x, o_{i,<t})},
\end{equation}
where $\pi_{\theta_{\text{old}}}$ is the model before the current update. Then, the GRPO objective is given as follows,
\begin{equation}
\mathcal{J}_{\text{GRPO}}(\theta) =\frac{1}{G} \sum_{i=1}^G \frac{1}{|o_i|}\sum_{t=1}^{|o_i|}\left[\min \left(\rho_{i,t} \hat{A}_{i,t}, \text{clip}\left(\rho_{i,t}, 1 - \epsilon, 1 + \epsilon\right) \hat{A}_{i,t}\right) - \beta \mathbb{D}_{\text{KL}}\left[\pi_\theta \| \pi_{\text{ref}}\right]\right]
\end{equation}
\begin{equation}
\mathbb{D}_{\text{KL}}\left[\pi_\theta \| \pi_{\text{ref}}\right] = \frac{\pi_\theta(o_{i,t} \mid (I,x), x, o_{i,<t})}{\pi_{\text{ref}}(o_{i,t} \mid (I,x), x, o_{i,<t})} - \log\frac{\pi_\theta(o_{i,t} \mid (I,x), x, o_{i,<t})}{\pi_{\text{ref}}(o_{i,t} \mid (I,x), x, o_{i,<t})} - 1,
\end{equation}
where $\epsilon$ is a hyperparameter controlling the clipping range, $\pi_{\text{ref}}$ is the model fixed after SFT stage, and $\beta$ is the KL penalty coefficient.

We argue that this formulation is suited to policy exploration in our reasoning-driven task. 
Given that the model is already capable of producing structured reasoning traces after SFT, GRPO allows it to freely explore different reasoning paths. In each iteration, the model generates diverse reasoning strategies that may lead to different predicted object sets. The reward function then guides the model to reinforce reasoning paths that yield accurate predictions.

\subsubsection{Reward Modeling}
\label{sec:reward-modeling}

\paragraph{Accuracy Reward. }
We use the F1 score to jointly evaluate the precision and recall of the model’s predictions. Given a set of predicted boxes $\hat{B}$ and the ground-truth set $B^*$, since box hints are provided as input, we define a match only when a predicted box exactly overlaps with a ground-truth box (i.e., IoU = 1), which encourages the model to select final outputs directly from the box hints. Let $M = \hat{B} \cap B^*$ denote the set of matched box pairs under this criterion. We compute precision, recall, and the F1 reward as:
\begin{equation}
\text{Precision} = \frac{|M|}{|\hat{B}|}, \quad
\text{Recall} = \frac{|M|}{|B^*|}, \quad
r^{\text{F1}} = \frac{2 \cdot \text{Precision} \cdot \text{Recall}}{\text{Precision} + \text{Recall}}.
\end{equation}

\paragraph{Format Reward.}
To encourage interpretable and well-structured output, we define a format reward $r^{\text{fmt}}$ that equals 1 if the output follows the required structure: the reasoning must be enclosed in \texttt{<think>}...\texttt{</think>} and the final result in \texttt{<answer>}...\texttt{</answer>}, and 0 otherwise.

The total reward is a weighted combination of the accuracy and format rewards, i.e., 
    $r_i = \lambda \cdot r_i^{\text{F1}} + (1 - \lambda) \cdot r_i^{\text{fmt}}$,
where $\lambda = 0.9$ to emphasize correct detection while still enforcing output structure.

\section{Experiments}

In this section, we evaluate the effectiveness of our CoT-based reasoning approach for object referring. We first introduce the experimental setup, then present in-domain results on the HumanRef benchmark, followed by out-of-domain evaluation on the RefCOCOg benchmark. Lastly, we conduct ablation studies to analyze key design choices.

\subsection{Experimental Setup}
\label{sec:setup}

\paragraph{Model Setting.}
We use Qwen2.5-VL-7B-Instruct as our base model. Qwen2.5-VL outputs absolute bounding box coordinates rather than quantized tokens, which provides better localization accuracy for detection tasks. We adopt this native decoding format for final bounding box predictions.

\paragraph{SFT Training.}
We fine-tune the model on the full HumanRef-CoT dataset using supervised learning. We use a learning rate of 2e-5, weight decay of 0.01, and cosine decay scheduling. The maximum generation length is set to 2048 tokens. During SFT, the vision encoder and MLP projector are frozen, and we update only the LLM parameters. For each training instance, we use all person bounding boxes in the image as box hints.

\paragraph{GRPO Training.}
After SFT, we apply GRPO for reward-driven post-training. We continue training on HumanRef-CoT, but randomly shuffle the box hint order in each training data to create novel input configurations. This leads the model to explore different reasoning paths than those seen during SFT. During this phase, we train only the LLM. We use a learning rate of 1e-6, 8 rollout samples per input, a batch size of 8, and gradient accumulation steps of 2. The KL penalty coefficient $\beta$ is set to 0.04, the sampling temperature to 1.0, and the output length remains 2048 tokens.

\paragraph{Evaluation Protocol.}
For in-domain evaluation, we evaluate our model on the HumanRef benchmark, which consists of six subsets: attribute, position, interaction, reasoning, celebrity recognition, and rejection. Following~\cite{jiang2025referringperson}, we report Recall (R), Precision (P), and DensityF1 (DF1) scores averaged over IoU thresholds from 0.5 to 0.95. For the rejection subset, we report the rejection score, defined as the proportion of 1,000 images where the model correctly outputs no bounding box when the object described by the referring expression is not present in the image. For out-of-domain evaluation, we evaluate our model on the RefCOCOg dataset and report accuracy at an IoU threshold of 0.5. We compare three variants: 1) Rex-Thinker-Plain, which is trained on HumanRef-CoT using SFT only on the final detection outputs, without reasoning supervision; 2) Rex-Thinker-CoT, which is trained with SFT on both the reasoning process and the final answer; and 3) Rex-Thinker-GRPO, which is initialized from Rex-Thinker-CoT and further optimized with GRPO training.

\begin{table*}[t]
\centering\vspace{-2em}
  \resizebox{1.0\linewidth}{!}{
    \begin{tabular}{c|ccc|ccc|ccc|ccc|ccc|ccc|c}
\toprule
\multirow{2}{*}{Method} & \multicolumn{3}{c|}{Attribute} & \multicolumn{3}{c|}{Position} & \multicolumn{3}{c|}{Interaction} & \multicolumn{3}{c|}{Reasoning} & \multicolumn{3}{c|}{Celebrity} & \multicolumn{3}{c|}{Average}                                             & \multicolumn{1}{l}{Rejection} \\ \cline{2-20} 
                        & R        & P        & DF1     & R        & P       & DF1     & R         & P        & DF1      & R        & P        & DF1     & R        & P        & DF1      & \multicolumn{1}{c}{R} & \multicolumn{1}{c}{P} & \multicolumn{1}{c|}{DF1} & \multicolumn{1}{c}{Score}     \\ \midrule
DINOX~\cite{ren2024dino}                   & 59.5     & 28.8     & 20.9    & 78.8     & 28.1    & 17.6    & 67.3      & 28.5     & 18.9     & 76.2     & 32.1     & 22.2    & 94.1     & 48.0     & 37.0     & 75.2                  & 33.1                  & 23.3                     & 36.0                          \\
InternVL-2.5-8B~\cite{chen2025expandingperformanceboundariesopensource}         & 23.5     & 39.0     & 27.1    & 23.0     & 28.0    & 24.3    & 27.8      & 40.1     & 31.3     & 17.5     & 22.8     & 18.9    & 57.4     & 59.3     & 58.0     & 29.8                  & 37.8                  & 31.9                     & 54.9                          \\
Ferret-7B~\cite{you2023ferret}               & 27.9     & 44.4     & 30.4    & 30.2     & 36.2    & 29.8    & 30.8      & 41.8     & 31.2     & 19.7     & 33.7     & 22.8    & 63.2     & 60.0     & 57.5     & 34.4                  & 43.2                  & 34.3                     & 2.0                           \\
Groma-7B~\cite{ma2024groma}                & 67.5     & 47.8     & 38.6    & 63.2     & 43.1    & 37.2    & 66.6      & 48.1     & 40.6     & 59.1     & 41.4     & 34.8    & 73.2     & 63.3     & 59.1     & 65.9                  & 48.7                  & 42.1                     & 0.0                           \\
ChatRex-7B~\cite{jiang2024chatrex}              & 44.3     & 78.0     & 51.8    & 48.0     & 66.7    & 52.5    & 49.6      & 74.8     & 56.5     & 36.6     & 65.1     & 42.8    & 73.7     & 76.5     & 74.2     & 50.4                  & 72.2                  & 55.6                     & 0.0                           \\
Qwen2.5-VL-7B~\cite{bai2025qwen2}           & 49.1     & 71.3     & 54.4    & 50.2     & 61.7    & 52.8    & 48.2      & 66.3     & 53.2     & 34.6     & 61.2     & 40.3    & 80.3     & 81.9     & 80.1     & 52.5                  & 68.5                  & 56.2                     & 7.1                           \\
DeepSeek-VL2-small~\cite{wu2024deepseek}      & 52.3     & 78.0     & 57.7    & 56.4     & 66.1    & 58.1    & 55.4      & 75.7     & 60.7     & 46.6     & 61.7     & 50.1    & 85.9     & 74.3     & 70.7     & 59.3                  & 71.2                  & 59.5                     & 3.1                           \\
Molmo-7B-D~\cite{deitke2024molmo}              & 82.7     & 86.4     & 76.3    & 78.0     & 80.6    & 72.4    & 69.9      & 77.7     & 66.1     & 72.1     & 80.4     & 65.5    & 85.9     & 87.5     & 82.9     & 77.7                  & 82.5                  & 72.6                     & \bf 68.6                          \\
RexSeek-7B~\cite{jiang2025referringperson}              & \underline{87.2}     & 86.8     & 81.5    & 86.1     & 86.3    & 83.8    & \bf 84.8      & \underline{84.6}     & \bf 80.7     & \bf 87.8     & \underline{84.7}     & \underline{81.5}    & 83.4     & 86.5     & 84.2     & \underline{85.9}                  & 85.8                  & \underline{82.3}                     & 54.1                          \\ \midrule
\rowcolor{gray!15}Rex-Thinker-Plain        & 83.0     & \bf 88.7     & 81.4    & 82.5     & 83.9    & 81.3    & 80.1      & \bf 85.6     & \underline{80.2}     & 80.5     & 82.2     & 77.3    & 86.7     & 88.7     & 86.8     & 82.6                  & 85.8                  & 81.4                     & 53.5                          \\
\rowcolor{gray!15}Rex-Thinker-CoT          & 86.6     & 87.7     & \underline{82.7}    & \underline{86.5}     & \underline{87.0}    & \underline{84.3}    & 79.6      & 81.7     & 77.2     & 85.7     & 83.8     & 80.3    & \underline{87.6}     & \bf 89.5     & \bf 87.2     & 85.2                  & \underline{85.9}                  & \underline{82.3}                     & 67.3                          \\
\rowcolor{gray!15}Rex-Thinker-GRPO         & \bf 88.5     & \bf 88.7     & \bf 84.1    & \bf 87.2     & \bf 87.1    & \bf 84.6    & \underline{81.5}      & 83.5     & 79.1     & \underline{87.7}     & \bf 85.4     & \bf 82.3    & \bf 88.0     & \underline{89.3}     & \bf 87.2     & \bf 86.6                  & \bf 86.8                 & \bf 83.5                     & \underline{68.2}                          \\ \bottomrule
\end{tabular}}
  \caption{In-domain evaluation results on the HumanRef benchmark. R, P, and DF1 represent Recall, Precision, and DensityF1. The \textbf{blod} and \underline{underline} fonts indicate the best and second numbers.}
  \label{tab:HumanRef_benchmark}
  \vspace{-2mm}
  \end{table*}

\subsection{In-domain Evaluation Results}
We begin by evaluating in-domain performance on the HumanRef benchmark to assess referring accuracy within the person domain. As shown in Table~\ref{tab:HumanRef_benchmark}, Rex-Thinker-CoT, trained with structured CoT supervision, consistently outperforms Rex-Thinker-Plain across most evaluation subsets. Specifically, it achieves average improvements of +2.6 Recall, +0.1 Precision, and +0.9 DensityF1, confirming that step-by-step reasoning leads to more accurate and well-grounded predictions. Most notably, the CoT-trained model shows a remarkable 13.8 point improvement in terms of Rejection Score on the rejection subset, indicating substantially reduced hallucination rates and enhanced ability to appropriately abstain from predictions when no valid target exists, which is a critical capability for real-world applications requiring high reliability.

Additional performance gains are realized through GRPO-based reinforcement learning. Rex-Thinker-GRPO demonstrates consistent improvements over Rex-Thinker-CoT, achieving gains of +1.4 Recall, +0.9 Precision, and +1.2 DensityF1. These results demonstrate the effectiveness of reward-based optimization in improving both the model’s reasoning process and prediction accuracy. While supervised CoT training provides strong guidance and teaches the model how to reason step by step, it may constrain the model to follow fixed patterns and limit its ability to explore more optimal reasoning strategies. In contrast, GRPO enables dynamic exploration of alternative reasoning strategies that better optimize for task-level objectives, leading to improved performance.

Among the six subsets in HumanRef benchmark, Rex-Thinker-GRPO achieves leading metrics in most categories. Notably, in the \textit{Reasoning} subsets, it outperforms Rex-Thinker-Plain by a +5 DensityF1, attributed to its CoT capabilities for reasoning tasks. The only subset with relatively weaker performance is \textit{Interaction} subset, please refer to Section~\ref{sec:con} for analysis. 
 
\subsection{Out-of-domain Evaluation Results}
To evaluate the generalization of Rex-Thinker to unseen object categories, we conduct experiments on the out-of-domain RefCOCOg dataset. We first adopt a zero-shot setting: the model is trained only on HumanRef-CoT and directly evaluated on RefCOCOg. Given a referring expression and its original COCO category label, we use Grounding DINO~\cite{liu2024grounding} to detect all instances of the target object category and use the detected results as box hints to Rex-Thinker.

\begin{figure*}[t]\centering\vspace{-2em}
\includegraphics[width=1\linewidth]{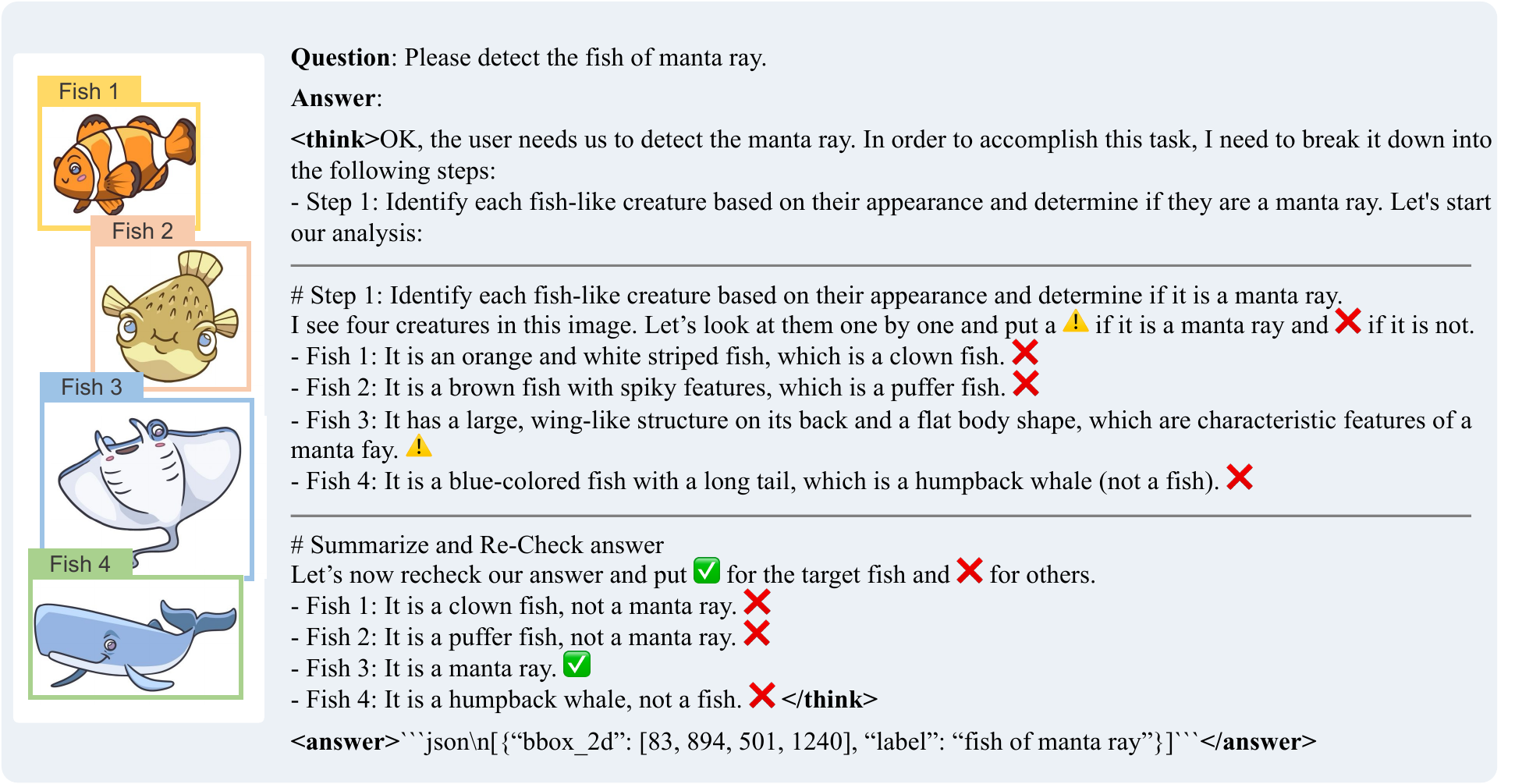}\vspace{-1mm}
\caption{The out-of-domain result. We use Rex-Thinker-GPRO trained on HumanRef-CoT to infer an unseen category (i.e., fish), resulting in a strong generalization. Boxes in the image denote hints.}
\label{fig:human2obj}
\vspace{-0.5em}
\end{figure*}

\begin{figure*}[t]
\centering
\begin{minipage}[t]{0.3\linewidth}
  \vspace{0pt} 
  \centering
  \resizebox{\linewidth}{!}{
    \begin{tabular}{c|cc}
\toprule
\multirow{2}{*}{Model} & \multicolumn{2}{c}{RefCOCOg} \\ \cline{2-3} 
                       & val           & test         \\ \midrule
RexSeek-7B~\cite{jiang2025referringperson}  & 84.0    & 84.4 \\
Grounding DINO~\cite{liu2024grounding}         & 86.1          & 87.0         \\
QwenVL-2.5-7B~\cite{bai2025qwen2}          & 87.2          & 87.2         \\
ChatRex-7B~\cite{jiang2024chatrex}             & 89.8          & 90.0         \\ \midrule
\rowcolor{gray!15}Rex-Thinker-CoT        & 81.2          &  80.3            \\
\rowcolor{gray!15}Rex-Thinker-GRPO        & 83.2         &  83.3          \\
\rowcolor{gray!15}Rex-Thinker-GRPO$^*$       & 89.2         & 88.8             \\ \bottomrule
    \end{tabular}
  }
  \captionof{table}{Out-of-domain evaluation results on RefCOCOg. $^*$Fine-tuned on RefCOCOg using GRPO.}
  \label{tab:refcocog_results}
\end{minipage}
\hfill
\begin{minipage}[t]{0.65\linewidth}
  \vspace{0pt} 
  \centering
  \includegraphics[width=\linewidth]{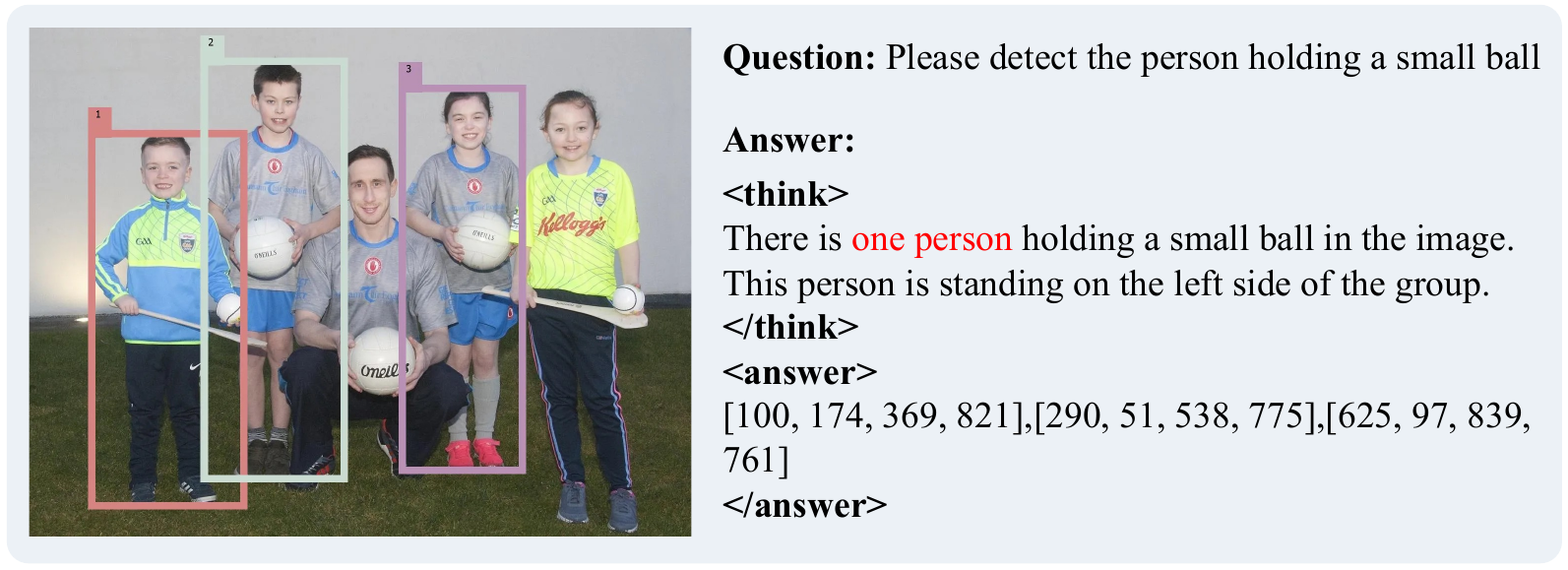}
  \caption{Predictions from a model that was trained with GRPO only, without CoT-based supervised fine-tuning as cold-start initialization. Boxes in the image denote answers.}
  \label{fig:no_cold_start}
\end{minipage}
\vspace{-1em}
\end{figure*}

As shown in Table~\ref{tab:refcocog_results}, the CoT-trained model already performs competitively without any task-specific tuning. Further gains are achieved by applying GRPO for post training, demonstrating that reward-driven training enhances the model’s ability to generalize beyond the training domain. Interestingly, we find that Rex-Thinker maintains its structured CoT behavior even on novel categories. As illustrated in Figure~\ref{fig:human2obj}, Rex-Thinker-GRPO successfully generalizes to detect the fish of manta ray (with ``fish'' bounding boxes as hints) while adhering to its \textit{planning-action-summarization} reasoning paradigm. Notably, the model demonstrates self-correction ability: when provided with an incorrect hint label (e.g., a whale was incorrectly labeled as a "fish" in hint boxes), Rex-Thinker rectifies the error through logical reasoning and explicitly rejects the misclassification.

To further explore the upper bound of the model, we fine-tune Rex-Thinker-CoT  using GRPO directly on RefCOCOg. This leads to additional performance improvements, achieving results comparable to state-of-the-art referring models. The experiment results highlight the adaptability of our reasoning paradigm across domains and the effectiveness of reward-based optimization in extending CoT reasoning to unseen categories.

\subsection{Ablations}

\paragraph{Effect of Retrieval-based Referring.}
Our approach adopts a retrieval-based formulation of object referring by providing the model with candidate object boxes as box hints. This design serves two key purposes: first, it allows the model to reason over each candidate region individually, aligning each step of the reasoning process with a specific image region and thereby ensuring grounded, interpretable outputs; second, it enables the model to reference these box hints when producing the final prediction, reducing the difficulty of direct coordinate regression.

To evaluate the impact of this retrieval-based design on referring accuracy, we conduct an ablation study by fine-tuning Qwen2.5-VL-7B on HumanRef-CoT with and without box hints. In this experiment, we do not include CoT supervision, as CoT reasoning inherently depends on the presence of box hints. As shown in Table~\ref{tab:ab1_retrieval_method}, incorporating box hints as input leads to substantial performance improvements across all major metrics, with average increases of 13.2, 11.7, and 10.8 points in Recall, Precision, and Density F1, respectively. While the model without box hints shows higher performance on the rejection subset, we attribute this phenomenon primarily to its over-rejection behavior. By analyzing the full test set, we observe that the no-hint model incorrectly abstains from prediction on 189 samples across the five non-rejection subsets, compared to only 134 for the box-hint variant. These results indicate that box hints facilitate more accurate predictions by reducing the difficulty of direct coordinate regression.

\paragraph{Impact of CoT-based Cold Start on GRPO.}
In Rex-Thinker, we adopt a two-stage training strategy where the model is first supervised using CoT-annotated data, followed by GRPO-based reinforcement learning. To assess the importance of this CoT-based initialization, we compare GRPO training with and without the cold-start SFT stage.

As shown in Table~\ref{tab:cot_ablation}, the model with CoT-based SFT achieves higher final performance than the direct GRPO model, indicating that the initial exposure to structured reasoning patterns provides a more effective starting point for reward-driven learning. Furthermore, as illustrated in Figure~\ref{fig:no_cold_start}, models trained without CoT supervision tend to generate unstructured or incoherent reasoning traces, lacking the verifiable and trustworthy qualities we aim to promote. In contrast, CoT-pretrained models produce well-formed thinking steps aligned with our planning, action, and summarization framework.

\begin{table*}[t]
\centering\vspace{-2em}
  \resizebox{1.0\linewidth}{!}{
    \begin{tabular}{c|ccc|ccc|ccc|ccc|ccc|ccc|c}
\toprule
\multirow{2}{*}{\begin{tabular}[c]{@{}c@{}}With \\ Box Hint\end{tabular}} & \multicolumn{3}{c|}{Attribute} & \multicolumn{3}{c|}{Position} & \multicolumn{3}{c|}{Interaction} & \multicolumn{3}{c|}{Reasoning} & \multicolumn{3}{c|}{Celebrity} & \multicolumn{3}{c|}{Average} & Rejection \\ \cline{2-20} 
                               & R        & P        & DF1     & R        & P       & DF1     & R         & P        & DF1      & R        & P        & DF1     & R        & P        & DF1      & R        & P       & DF1     & Score     \\ \midrule
No                             & 66.4     & 74.3     & 67.2    & 69.3     & 71.9    & 69.5    & 65.2      & 72.1     & 66.4     & 63.6     & 67.5     & 62.2    & 82.4     & 84.6     & 82.7     & 69.4     & 74.1    & 69.6    & \bf 71.7      \\
Yes                            & \bf 83.0     & \bf 88.7     & \bf 81.4    & \bf 82.5     & \bf 83.9    & \bf 81.3    & \bf 80.1      & \bf 85.6     & \bf 80.2     & \bf 80.5     & \bf 82.2     & \bf 77.3    & \bf 86.7     & \bf 88.7     & \bf 86.8     & \bf 82.6     & \bf 85.8    & \bf 81.4    & 53.5      \\ \bottomrule
\end{tabular}}
  \caption{Ablation study on the retrieval-based design of our model. We compare performance with and without box hints to assess their impact on referring accuracy.}
  \label{tab:ab1_retrieval_method}
  \vspace{-2mm}
  \end{table*} 

\begin{table*}[t]
\centering
  \resizebox{1.0\linewidth}{!}{
    \begin{tabular}{c|ccc|ccc|ccc|ccc|ccc|ccc|c}
\toprule
\multirow{2}{*}{\begin{tabular}[c]{@{}c@{}}With \\ Cold Start\end{tabular}} & \multicolumn{3}{c|}{Attribute} & \multicolumn{3}{c|}{Position} & \multicolumn{3}{c|}{Interaction} & \multicolumn{3}{c|}{Reasoning} & \multicolumn{3}{c|}{Celebrity} & \multicolumn{3}{c|}{Average} & Rejection \\ \cline{2-20} 
                                                                            & R        & P        & DF1     & R        & P       & DF1     & R         & P        & DF1      & R        & P        & DF1     & R        & P        & DF1      & R        & P       & DF1     & Score     \\ \midrule
No                                                                          & 81.4     & 85.8     & 78.1    & 80.2     & 80.2    & 77.5    & 79.6      & 82.6     & 78.0     & 77.6     & 75.0     & 70.6    & 87.3     & 86.5     & 84.8     & 81.2     & 82.0    & 77.8    & 66.4      \\
Yes                                                                         & \bf 88.5     & \bf 88.7     & \bf 84.1    & \bf 87.2     & \bf 87.1    & \bf 84.6    & \bf 81.5      & \bf 83.5     & \bf 79.1     & \bf 87.7     & \bf 85.4     & \bf 82.3    & \bf 88.0     & \bf 89.3     & \bf 87.2     & \bf 86.6                  & \bf 86.8                 & \bf 83.5                     & \bf 68.2      \\ \bottomrule
\end{tabular}}
  \caption{Ablation on the impact of CoT-based cold start on final performance after GRPO training.}
  \label{tab:cot_ablation}
  \vspace{-1em}
  \end{table*}

\section{Conclusion}
\label{sec:con}
We have presented Rex-Thinker, a novel framework that has reformulated the object referring problem as an explicit Chain-of-Thought reasoning process to achieve grounded and interpretable predictions. Unlike conventional approaches that have treated referring as direct bounding box prediction, our model has first detected candidate objects and then performed step-by-step verification against the referring expression through structured planning-action-summarization reasoning. To support this paradigm, we have constructed HumanRef-CoT, a large-scale dataset with reasoning traces that have enabled learning decomposed and interpretable reasoning patterns. Through a two-stage training approach combining SFT and GRPO-based RL, Rex-Thinker has demonstrated superior performance over prior works in both referring accuracy and rejection.

\paragraph{Limitation} As shown in Table~\ref{tab:HumanRef_benchmark}, our model has exhibited relatively weaker performance in the interaction subset. This limitation has arisen because the CoT reasoning process must simultaneously model relationships and interactions among multiple objects. Errors in this complex inference chain have propagated, leading to misleading final responses. Please refer to the Appendix for further limitation analysis.


{
    \small
    \bibliographystyle{plain}
    \bibliography{main}
}

\newpage
\appendix

\section{Appendix}

\subsection{More Details on Constructing HumanRef-CoT}

\subsubsection{Prompt for GPT-4o}
To annotate HumanRef-CoT dataset using GPT-4o, we designed a two-part prompting strategy that addresses the diverse reasoning requirements across different subsets. This strategy consists of a \textbf{unified system prompt} and a set of \textbf{subset-specific in-context examples}.

The system prompt is shared across all subsets and instructs the model on how to interpret the input, which includes an image, a referring expression, and candidate bounding boxes. It also defines the expected format of the response, including the use of structured reasoning and answer tags. In addition to the system prompt, each of the six subsets in HumanRef-CoT namely attribute, position, interaction, reasoning, celebrity recognition, and rejection, is paired with a collection of in-context examples. These examples are carefully curated to reflect the specific annotation challenges and reasoning patterns required for each subset. They guide GPT-4o in producing chain-of-thought (CoT) rationales that are consistent with human annotations in both style and logic.

In the following sections, we first present the shared system prompt. Then, for each subset, we provide the corresponding in-context examples and visualization results.

\paragraph{Unified System Prompt. }The system prompt instructs the model to perform detailed visual reasoning based on either positional or attribute-based referring expressions. It emphasizes step-by-step analysis, beginning with predefined reasoning steps (first attributes, then orientation), and requires the model to explicitly evaluate each candidate object. Special symbols are also used to denote matching, non-matching, and reference entities during analysis.

\begin{figure*}[h]\centering\vspace{-1em}
\includegraphics[width=1\linewidth]{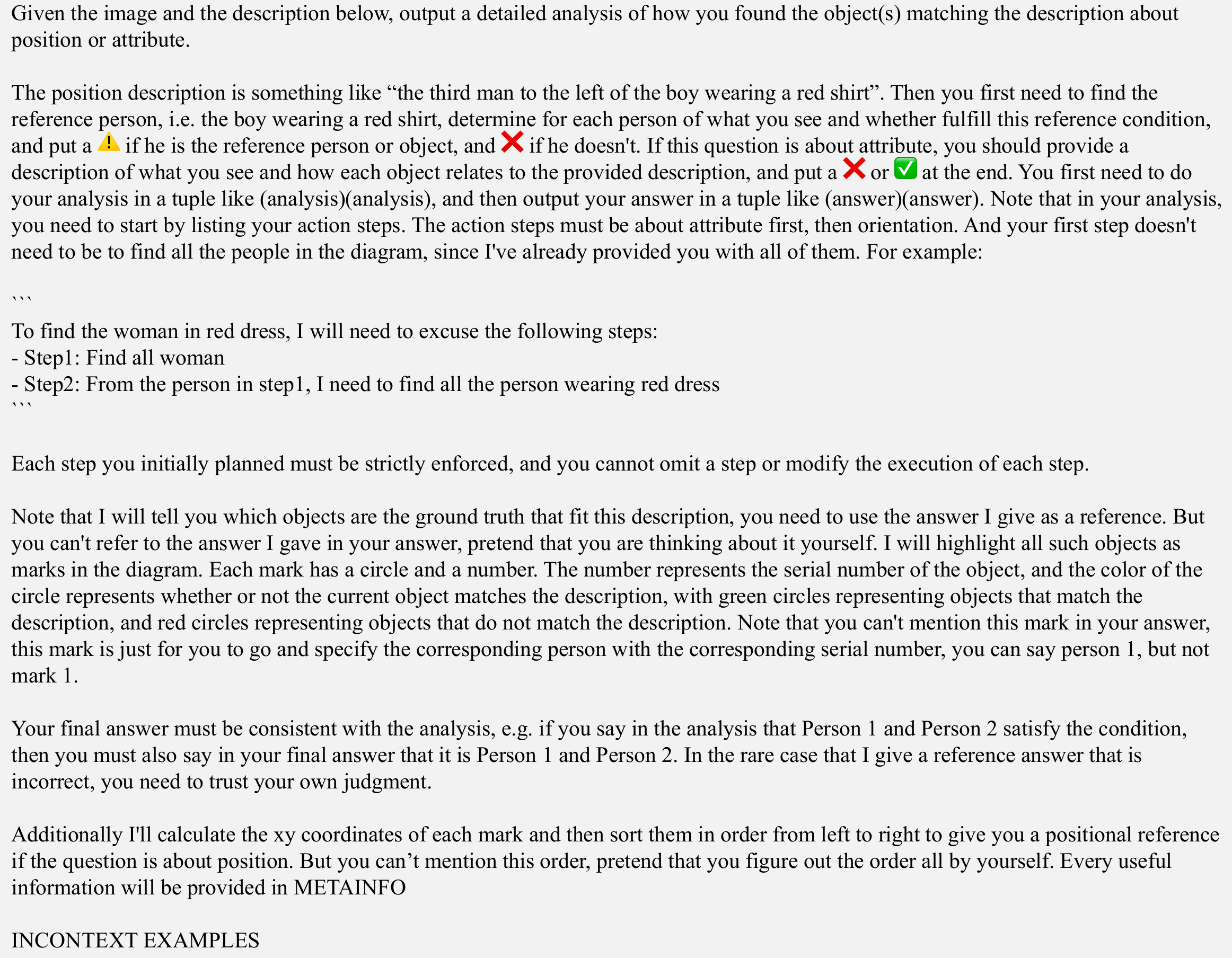}\vspace{-1mm}
\caption{The system prompt used to instruct GPT-4o on visual reasoning for HumanRef-CoT. It specifies output format, reasoning steps, symbol conventions, and the expected alignment between intermediate analysis and final answers.}
\label{fig:system_prompt}
\end{figure*}

\paragraph{Subset-Specific In-Context Examples.} After the system prompt, we provide in-context examples to guide the model toward producing outputs aligned with our CoT structure. These examples help reinforce consistent reasoning patterns. HumanRef-CoT includes six subsets: attribute, position, interaction, reasoning, celebrity recognition, and rejection. Each subset uses its own set of in-context examples tailored to its specific reasoning needs.

We show the in-context prompts used for each subset, along with representative outputs generated by GPT-4o.

\begin{table*}[h]
\centering
  \resizebox{1.0\linewidth}{!}{
    \begin{tabular}{c|c|cc|cc|cc|c|c}
\hline
\multirow{2}{*}{Subset} & attribute                                            & \multicolumn{2}{c|}{position}                                                                                         & \multicolumn{2}{c|}{interaction}                                                                                            & \multicolumn{2}{c|}{reasoning}                                                                                                      & celebrity & rejection \\ \cline{2-10} 
                        & -                                                    & \begin{tabular}[c]{@{}c@{}}inner \\ position\end{tabular} & \begin{tabular}[c]{@{}c@{}}outer \\ position\end{tabular} & \begin{tabular}[c]{@{}c@{}}inner \\ interaction\end{tabular} & \begin{tabular}[c]{@{}c@{}}outer \\ interaction\end{tabular} & \begin{tabular}[c]{@{}c@{}}inner positon\\  reasoning\end{tabular} & \begin{tabular}[c]{@{}c@{}}attribute \\ reasoning\end{tabular} & -         & -         \\ \hline
Prompt                  & Figure \ref{fig:attribute_prompt}  & Figure \ref{fig:inner_position_prompt}  & Figure \ref{fig:outer_position_prompt} & Figure \ref{fig:inner_interaction_prompt}  & Figure \ref{fig:outer_interaction_prompt}                  &   Figure \ref{fig:inner_position_reasoning_prompt}            &  Figure \ref{fig:attribute_reasoning_prompt}     &   Figure \ref{fig:celebrity_prompt}         &  Figure \ref{fig:rejection_prompt}          \\
Example                 & Figure \ref{fig:attribute_example} & Figure \ref{fig:inner_position_example} & Figure \ref{fig:outer_position_example} & Figure \ref{fig:inner_interaction_example}   & Figure \ref{fig:outer_interaction_example} &  Figure \ref{fig:inner_position_reasoning_example}                                                                 &  Figure \ref{fig:attribute_reasoning_example}                                                              &   Figure \ref{fig:celebrity_example}         &  Figure \ref{fig:rejection_example}          \\ \hline
\end{tabular}}
  \caption{Ablation study on the retrieval-based design of our model. We compare performance with and without box hints to assess their impact on referring accuracy.}
  \label{tab:ab1_retrieval_method}
  \vspace{-2mm}
  \end{table*}

\subsubsection{Evaluate GPT-4o on HumanRef}

Since we use GPT-4o to annotate HumanRef-CoT, a natural question is how well GPT-4o performs directly on the HumanRef benchmark when prompted in a similar style. To investigate this, we adopt a setup similar to the annotation phase, using the same SoM-style prompt and a set of visual marks (with all marks shown in red). However, we remove any hint indicating which objects are correct. We then evaluate GPT-4o on the HumanRef-Benchmark without prompting with ground-truth answers. As shown in Table~\ref{tab:gpt4o_performance}, GPT-4o achieves an average DF1 score of 53.2 without any hint supervision. This result suggests that while GPT-4o can be used to generate annotations when given the correct answer as reference, its standalone performance without answer supervision remains limited.

\begin{table*}[h]
\centering
  \resizebox{1.0\linewidth}{!}{
    \begin{tabular}{c|ccccccccccccccc|ccc|c}
\hline
\multirow{2}{*}{Method} & \multicolumn{3}{c}{Attribute} & \multicolumn{3}{c}{Position} & \multicolumn{3}{c}{Interaction} & \multicolumn{3}{c}{Reasoning} & \multicolumn{3}{c|}{Celebrity} & \multicolumn{3}{c|}{Average} & Rejection \\
                        & R        & P        & DF1     & R        & P       & DF1     & R         & P        & DF1      & R        & P        & DF1     & R        & P        & DF1      & R        & P       & DF1     & Score     \\ \hline
GPT-4o-CoT              & 50.2     & 56.2     & 50.9    & 56.1     & 56.8    & 55.1    & 52.8      & 56.8     & 53.2     & 53.3     & 52.9     & 51.1    & 54.9     & 54.3     & 53.2     & 54.3     & 55.2    & 53.2    & 14.8      \\
Rex-Thinker-GRPO         & \bf 88.5     & \bf 88.7     & \bf 84.1    & \bf 87.2     & \bf 87.1    & \bf 84.6    & \bf{81.5}      & \bf 83.5     & \bf 79.1     & \bf{87.7}     & \bf 85.4     & \bf 82.3    & \bf 88.0     & \bf{89.3}     & \bf 87.2     & \bf 86.6                  & \bf 86.8                 & \bf 83.5                     & \bf{68.2}       \\ \hline
\end{tabular}}
  \caption{Evaluation of GPT-4o on the HumanRef-Benchmark test set using SoM-style prompts without answer hints. The model achieves 53.2 average DF1 score, indicating limited standalone performance.}
  \label{tab:gpt4o_performance}
  \vspace{-2mm}
  \end{table*} 
\subsection{Experiment Details}
\subsubsection{CoT SFT Settings}
Table~\ref{tab:cot_sft} summarizes the full training hyperparameters and computational cost used during the CoT SFT stage. These settings were applied in the cold-start phase without prior instruction tuning.

\begin{table*}[h]
\centering
  \resizebox{1.0\linewidth}{!}{
    \begin{tabular}{cc|cc|cc}
\hline
batch size            & 4     & maximum gradient norm   & 1      & precision        & bf16   \\
gradient accumulation & 4     & learning rate scheduler & cosine & epochs           & 2      \\
learning rate         & 2e-5  & max length              & 2048   & times            & 10.1h  \\
optimizer             & AdamW & deepspeed               & zero3  & GPU              & 8xA100 \\
warm up ratio         & 0.03  & weight decay            & 0.01   & trainable module & LLM    \\ \hline
\end{tabular}}
  \caption{Training settings and cost statistics for CoT SFT.}
  \label{tab:cot_sft}
  \vspace{-1em}
  \end{table*} 

\subsubsection{GRPO Settings}
We provide the training configurations used during the GRPO stage in Table~\ref{tab:grpo}. We did not run full GRPO training on the entire HumanRef-CoT dataset. Instead, training was terminated when the reward signal plateaued, indicating convergence.

\begin{table*}[h]
\centering
  \resizebox{1.0\linewidth}{!}{
    \begin{tabular}{cc|cc|cc}
\hline
batch size            & 8     & num of rollout & 8     & precision        & bf16   \\
gradient accumulation & 2     & $\beta$           & 0.04  & epochs       & 0.25      \\
learning rate         & 1e-6  & temperature    & 1.0   & times            & 112h  \\
optimizer             & AdamW & deepspeed      & zero3 & GPU              & 8xA100 \\
warm up ratio         & 0.03  & weight decay   & 0.01  & trainable module & LLM    \\ \hline
\end{tabular}}
  \caption{Hyperparameters used during the GRPO training stage.}
  \label{tab:grpo}
  \end{table*}

\subsubsection{GRPO Training Analysis}

\begin{figure*}[h]\centering
\includegraphics[width=1\linewidth]{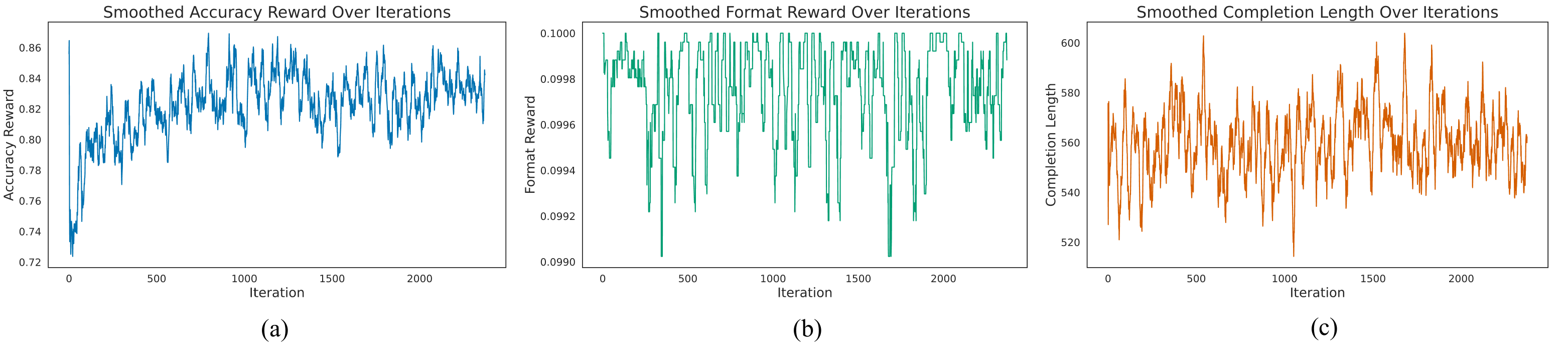}\vspace{-1mm}
\caption{GRPO training curves showing accuracy reward, format reward, and completion length over time.}
\label{fig:grpo_log}
\vspace{-0.5em}
\end{figure*}

We analyze the training logs of the GRPO stage. As shown in Figure~\ref{fig:grpo_log}, we visualize the changes in both reward signals and completion length throughout training.

Thanks to the cold-start CoT initialization, the model achieves a reasonably high accuracy reward at the beginning of GRPO training. At the same time, the format reward is nearly saturated from the start, indicating that the model has already learned to follow the correct output structure after CoT supervision. Meanwhile, the completion length remains stable at around 560 tokens throughout training. We attribute this to the model having already acquired the basic reasoning skills required for the referring task during the CoT fine-tuning phase, resulting in consistent output lengths with minimal fluctuation.

\subsection{Limitations and Broader Impacts}
\subsubsection{Inference Speed}
While the CoT-based design improves both interpretability and performance, it also introduces additional computational overhead at inference time. To quantify this, we randomly selected 100 images from the HumanRef-Benchmark test set and compared the average inference time per image between RexThinker-Plain and RexThinker-GRPO.

All experiments were conducted using the vLLM framework on a single NVIDIA A100 GPU. As shown in Table~3, RexThinker-GRPO exhibits slower inference due to its longer CoT-style outputs. This observation aligns with the general principle of test-time computation, where improved interpretability and accuracy often come at the cost of slower response time.

\begin{table*}[h]
\centering
  \resizebox{0.7\linewidth}{!}{
    \begin{tabular}{c|cc}
\hline
model                  & Rex-Thinker-Plain & Rex-Thinker-GRPO \\ \hline
average inference time & 1.13s                 & 6.68s             \\ \hline
\end{tabular}}
  \caption{Comparison of average inference time between RexThinker-Plain and RexThinker-GRPO. The CoT design in GRPO leads to slower inference.}
  \label{tab:grpo}
  \vspace{-1em}
  \end{table*} 

\subsubsection{Inconsistent Reasoning}
\begin{figure*}[h]\centering
\includegraphics[width=1\linewidth]{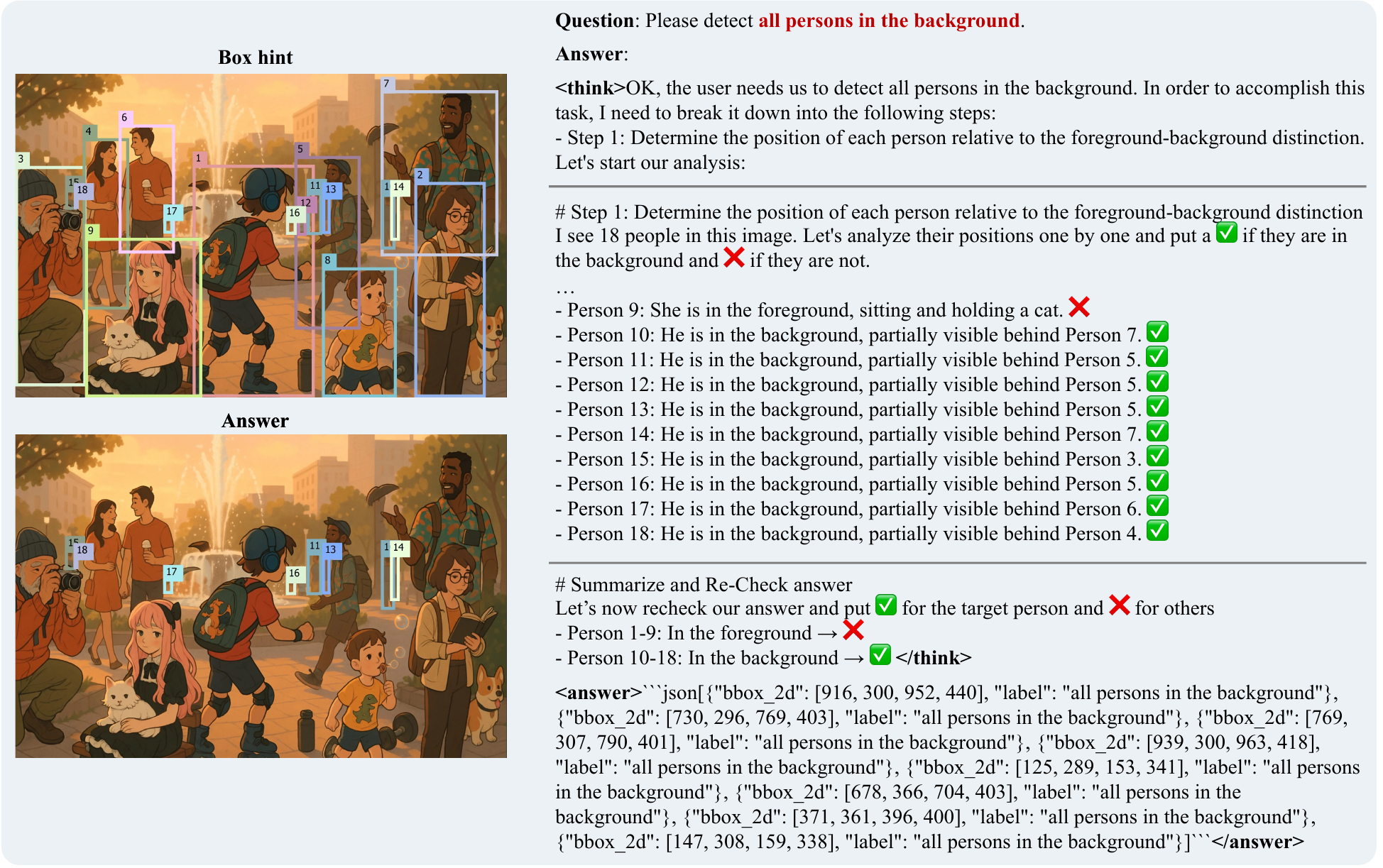}\vspace{-1mm}
\caption{Example of reasoning–answer mismatch. The number of predicted objects differs between reasoning and the final output.}
\label{fig:fail_case}
\vspace{-0.5em}
\end{figure*}

We observe occasional inconsistencies between the model’s reasoning process and its final output. For example, as shown in Figure~\ref{fig:fail_case}, the model identifies nine candidate objects in the reasoning phase but only includes eight in the final predicted coordinates.

We attribute such inconsistencies to the lack of explicit supervision enforcing alignment between the reasoning chain and the final answer during GRPO training. While most predictions remain consistent, these rare cases highlight a potential gap in our current framework. One possible solution is to introduce a consistency reward, which evaluates whether the number of objects summarized in the reasoning matches the number of bounding boxes in the final answer. This direction remains open for future exploration.

\subsubsection{Broader Impacts}
Referring models enable more flexible and natural interaction with visual scenes compared to traditional object detection, as they can understand complex, context-dependent language. This makes them valuable in applications such as surveillance, smart cities, and smart homes.

However, these models also raise concerns. In particular, they may expose privacy risks when deployed in sensitive environments and can inherit biases from training data or user input. Although CoT reasoning improves interpretability, ensuring its consistency and robustness remains an open challenge. Future work should address these risks to ensure safe and responsible deployment.
\subsection{Visualization Results}
\begin{figure*}[t]\centering
\includegraphics[width=1\linewidth]{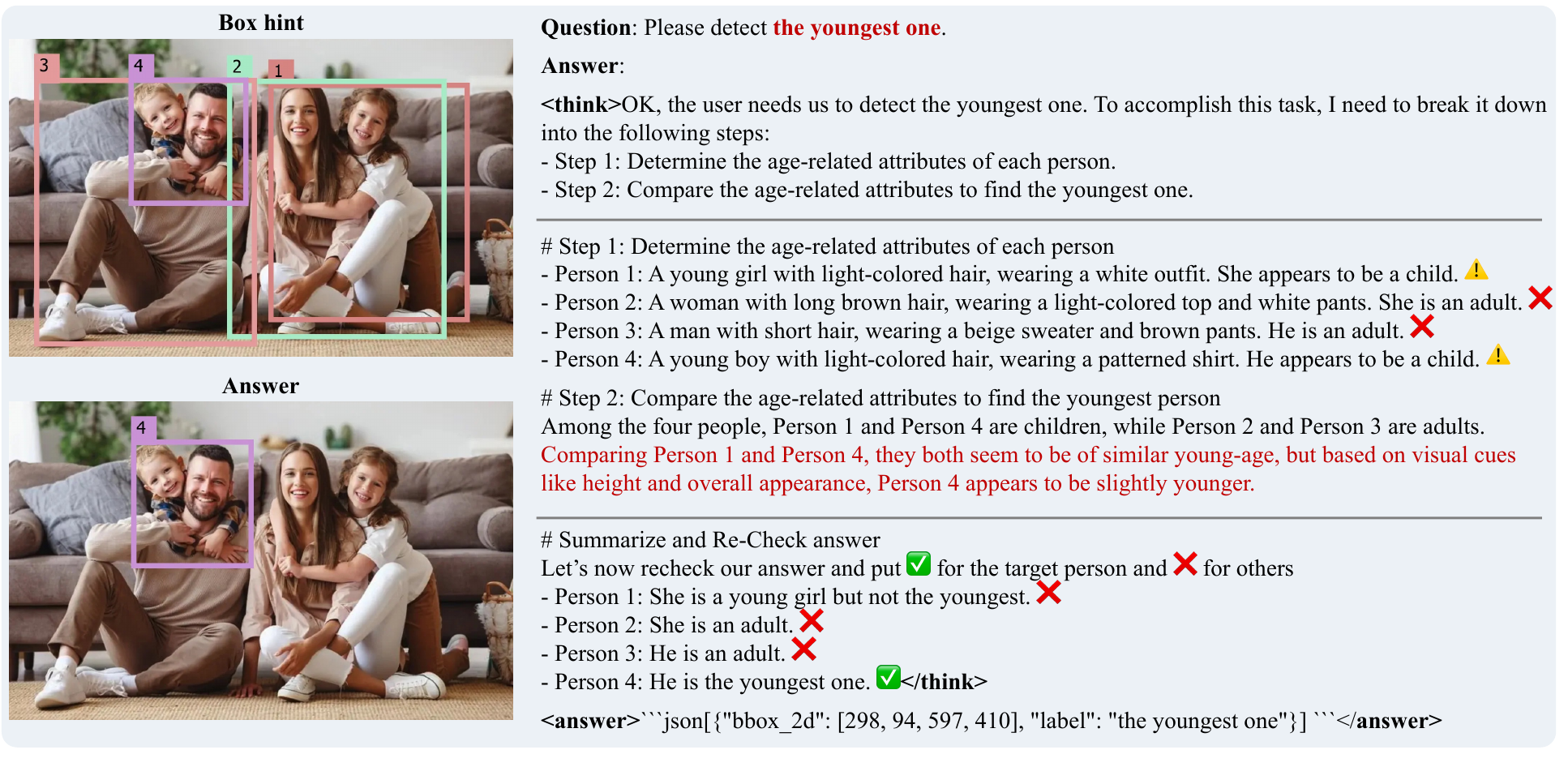}\vspace{-1mm}
\caption{Attribute referring example.}
\label{fig:demo1}
\end{figure*}

\begin{figure*}[t]\centering
\includegraphics[width=1\linewidth]{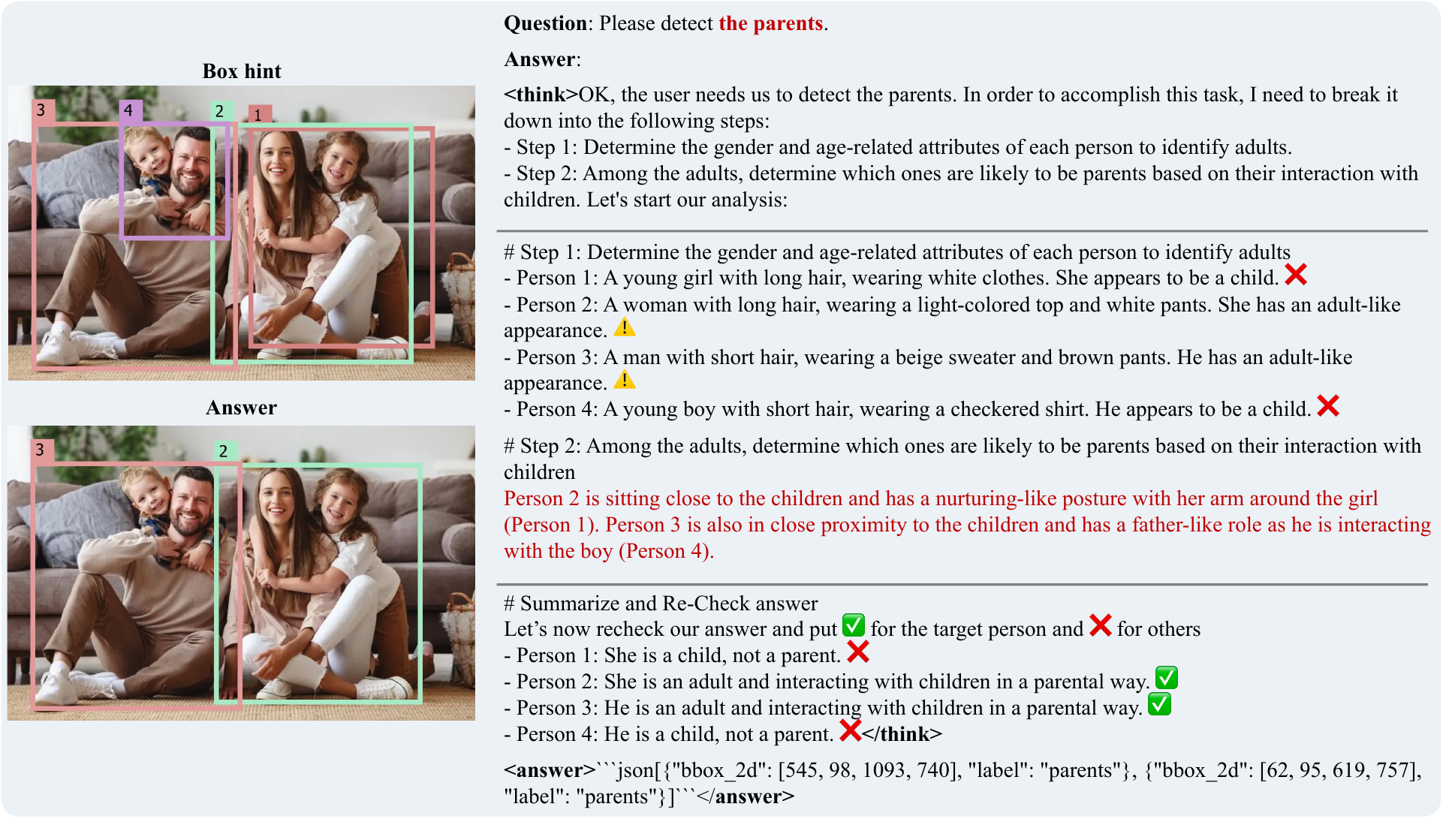}\vspace{-1mm}
\caption{Attribute referring example.}
\label{fig:demo2}
\end{figure*}

\begin{figure*}[t]\centering
\includegraphics[width=1\linewidth]{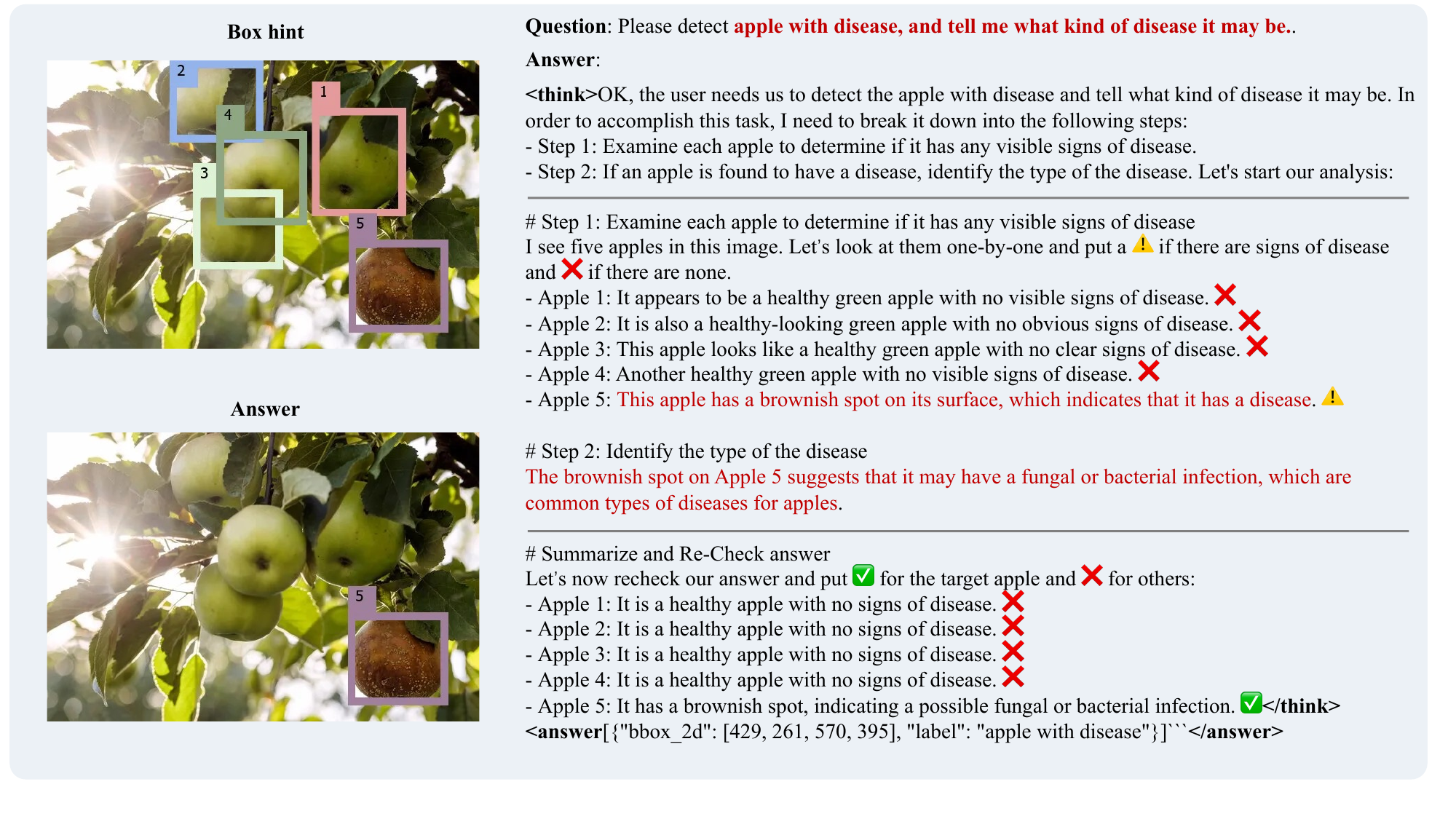}\vspace{-1mm}
\caption{Attribute referring example.}
\label{fig:demo3}
\end{figure*}

\begin{figure*}[t]\centering
\includegraphics[width=1\linewidth]{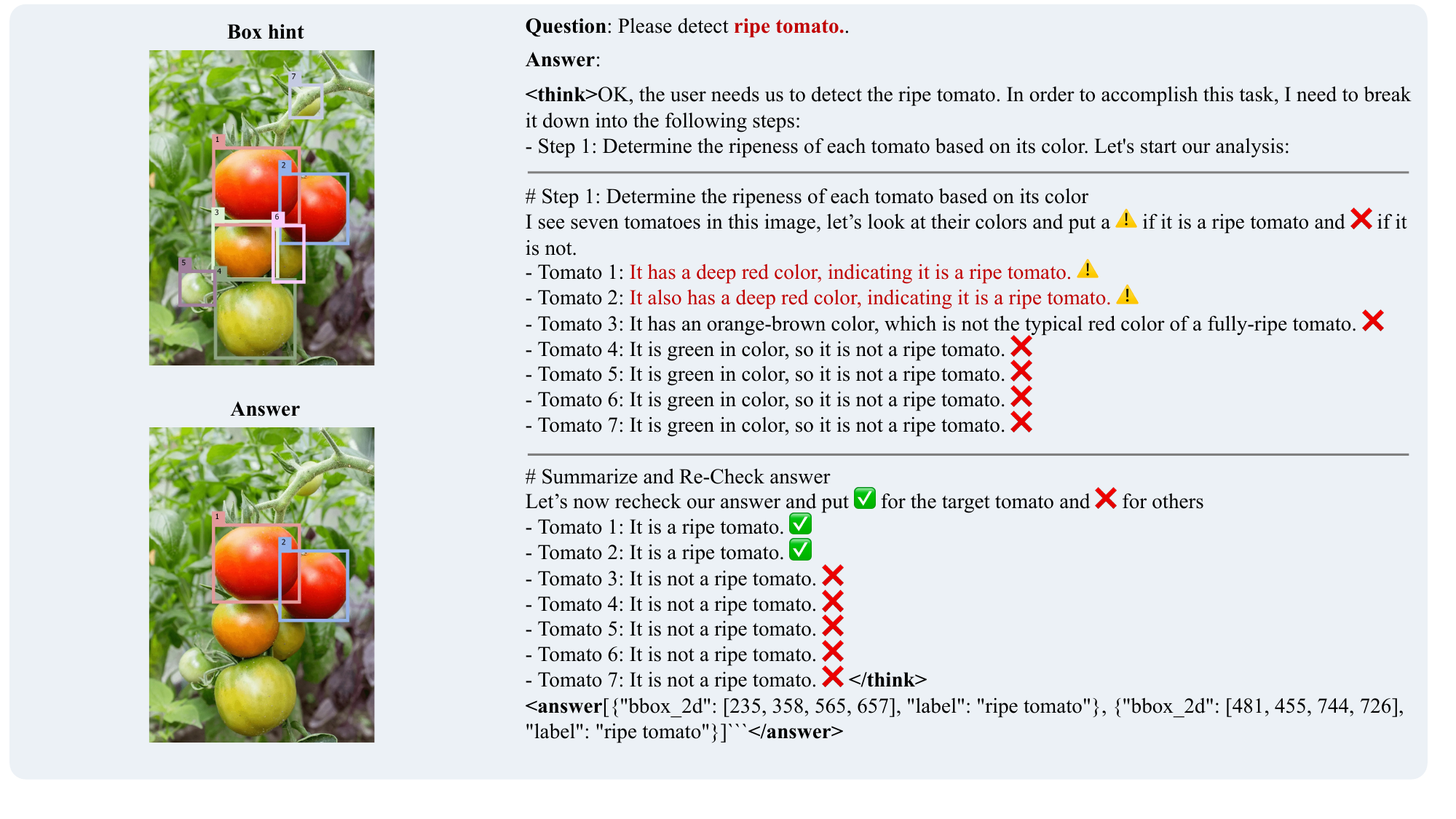}\vspace{-1mm}
\caption{Attribute referring example.}
\label{fig:demo4}
\end{figure*}

\begin{figure*}[t]\centering
\includegraphics[width=1\linewidth]{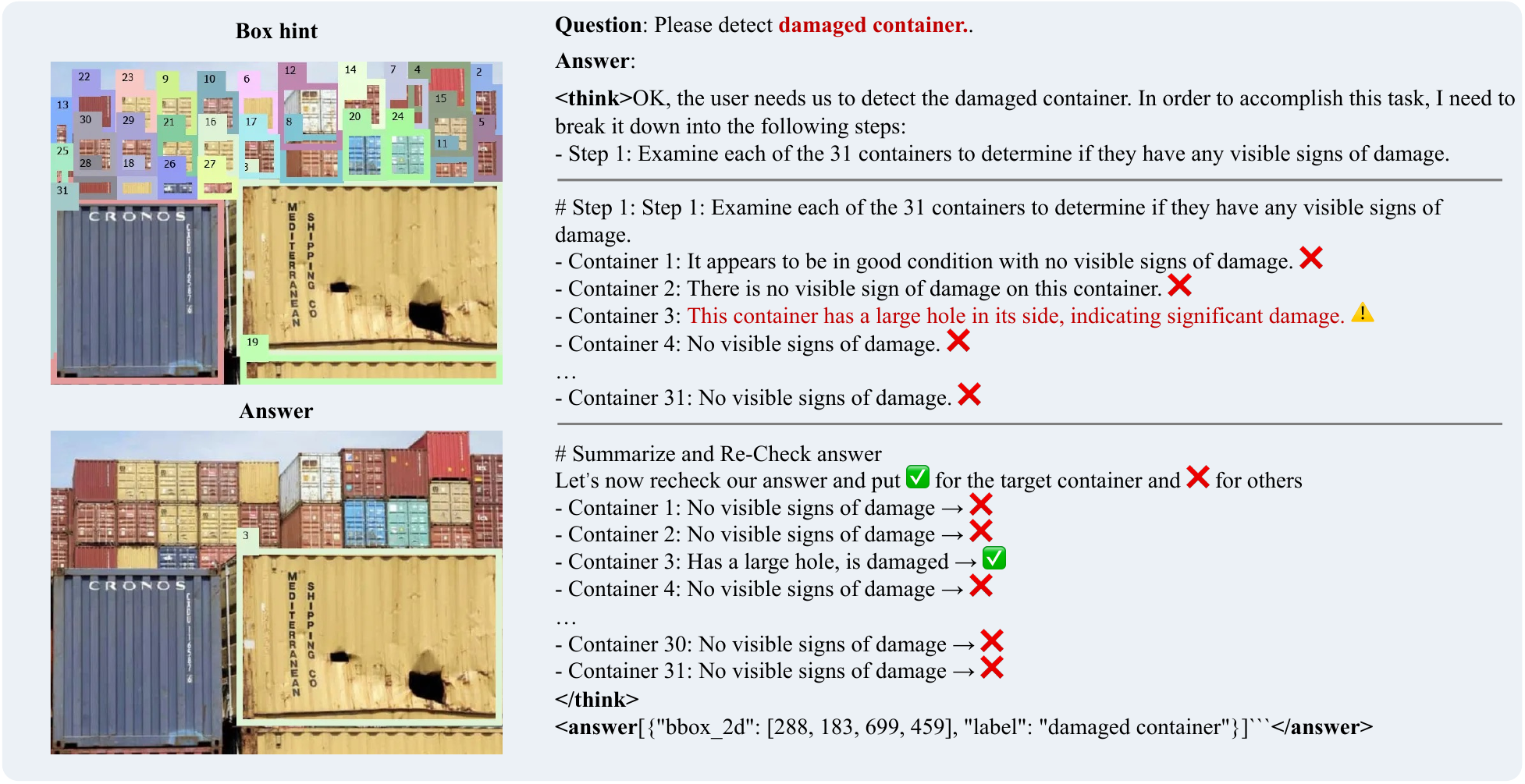}\vspace{-1mm}
\caption{Attribute referring example.}
\label{fig:demo5}
\end{figure*}

\begin{figure*}[t]\centering
\includegraphics[width=1\linewidth]{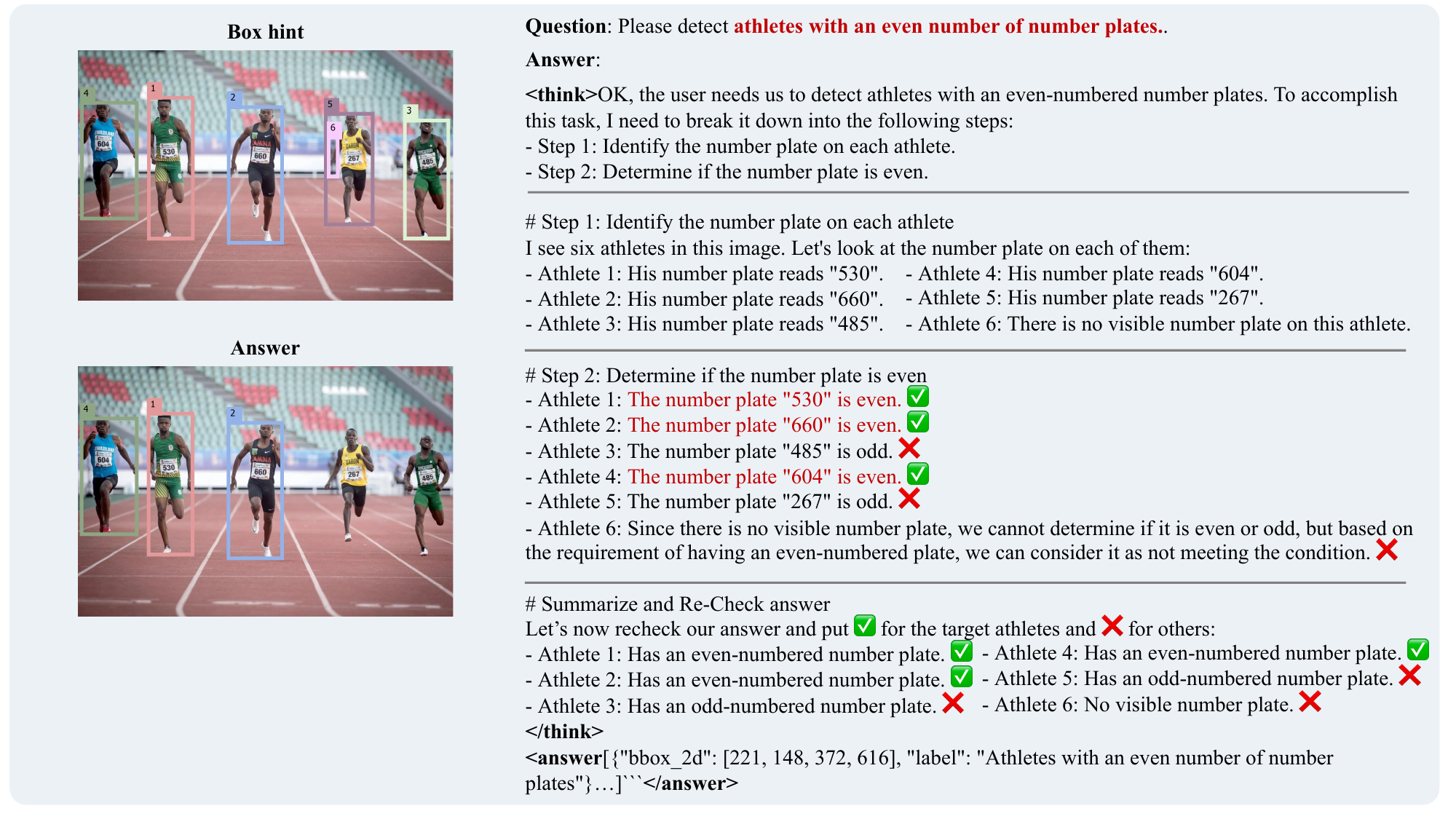}\vspace{-1mm}
\caption{Attribute referring example.}
\label{fig:demo6}
\end{figure*}

\begin{figure*}[h]\centering
\includegraphics[width=1\linewidth]{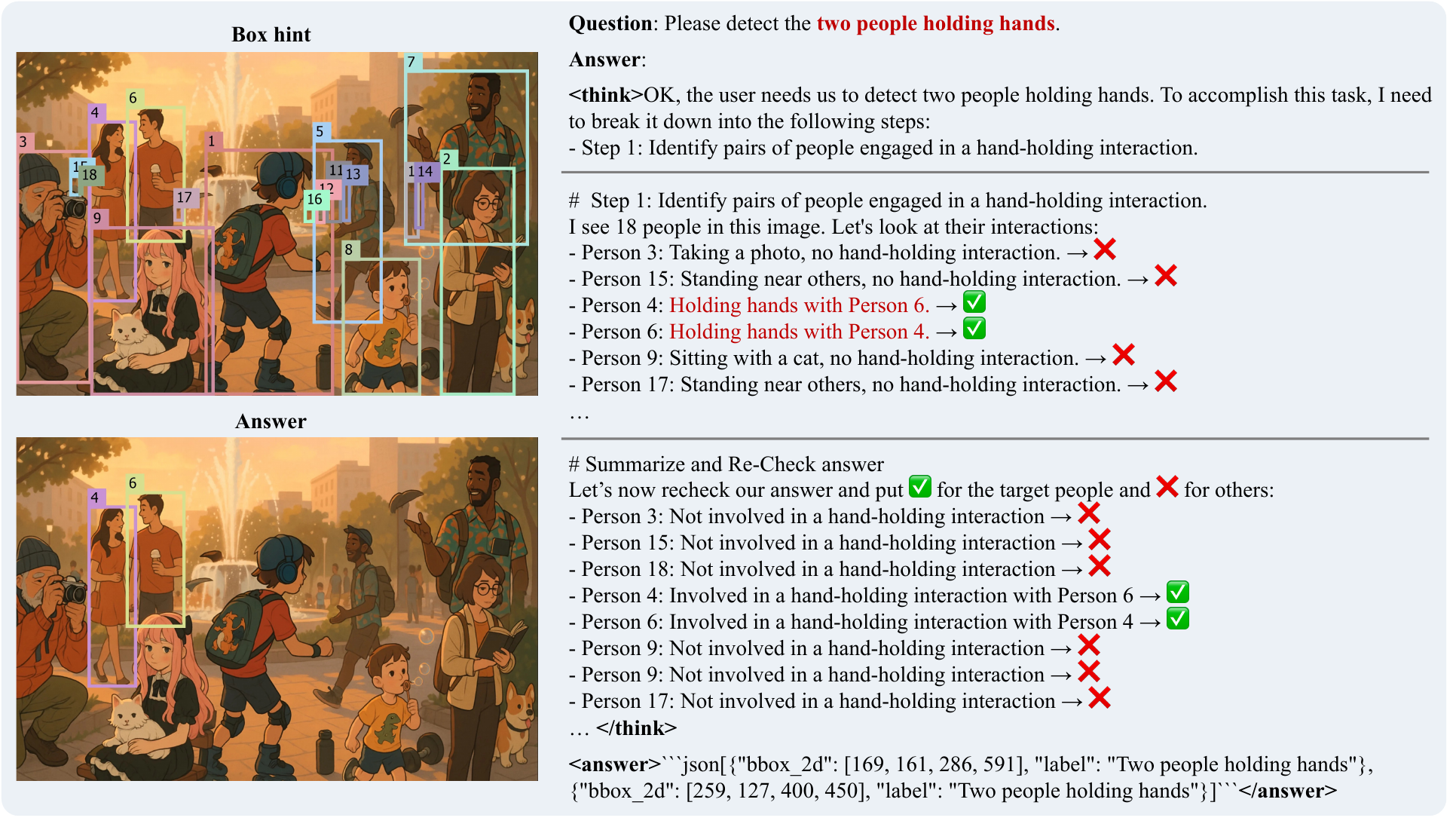}\vspace{-1mm}
\caption{Interaction referring example.}
\label{fig:demo7}
\end{figure*}

\begin{figure*}[h]\centering
\includegraphics[width=1\linewidth]{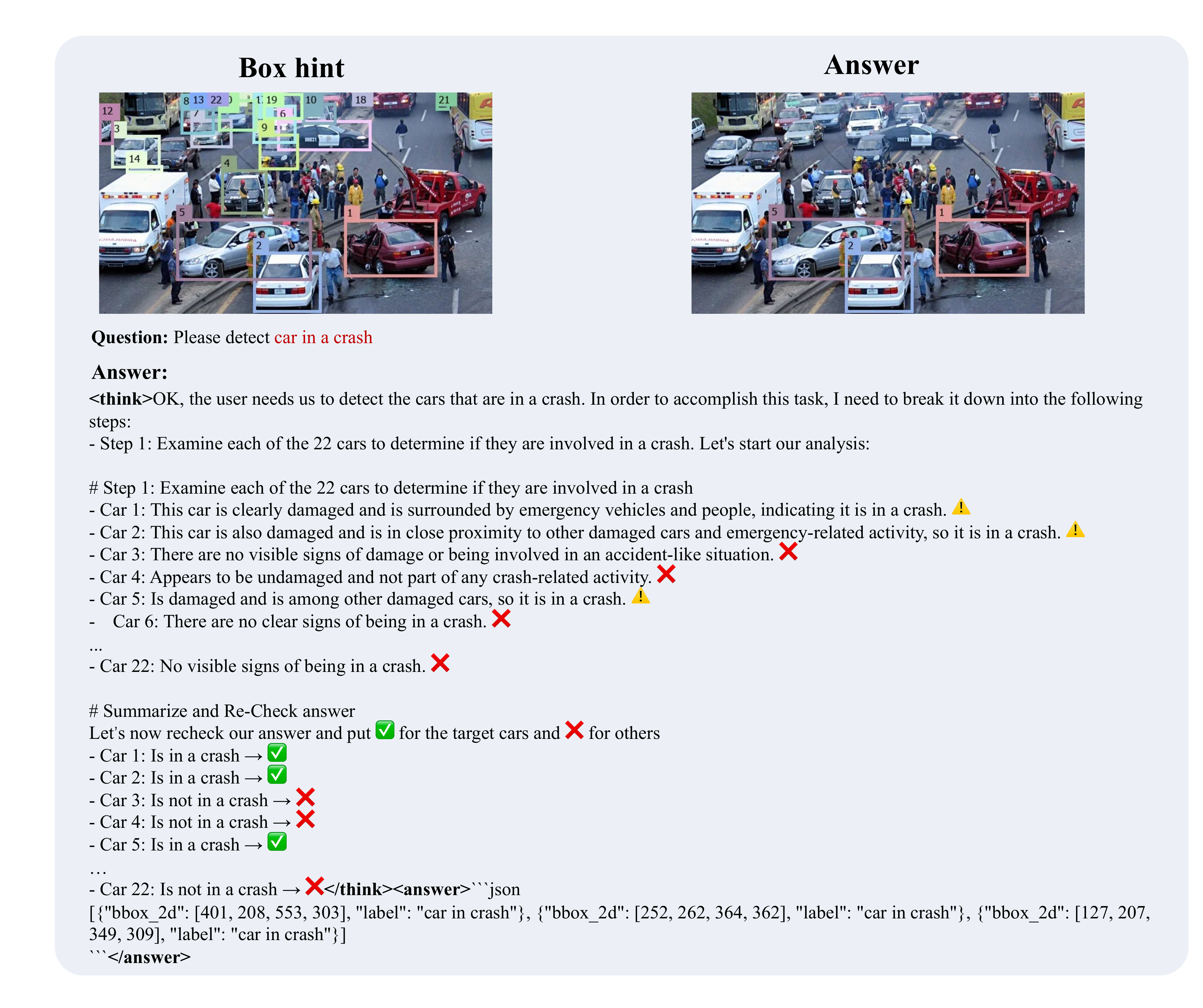}\vspace{-1mm}
\caption{Interaction referring example.}
\label{fig:demo8}
\end{figure*}

\begin{figure*}[h]\centering
\includegraphics[width=1\linewidth]{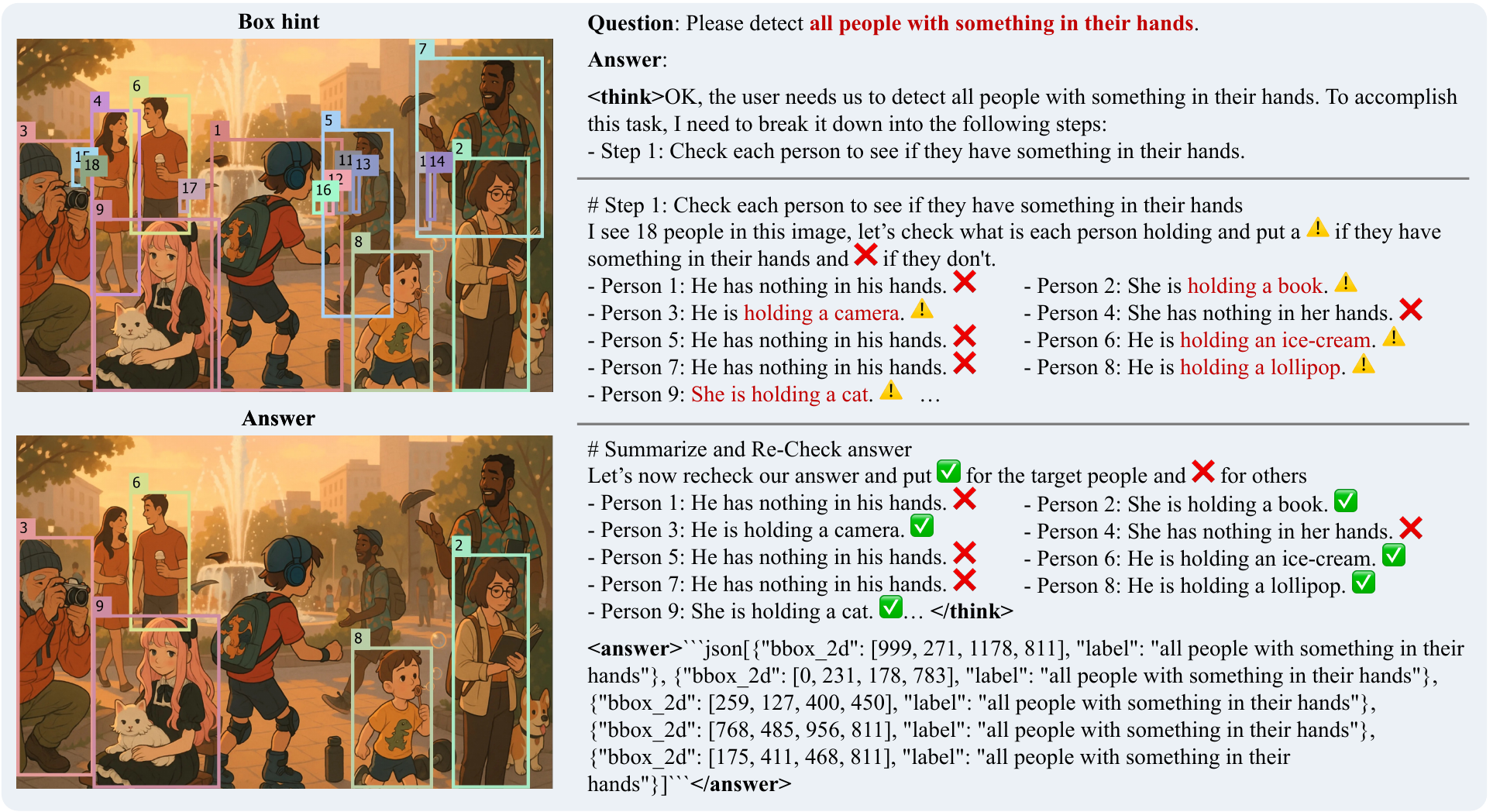}\vspace{-1mm}
\caption{Interaction referring example.}
\label{fig:demo9}
\end{figure*}

\begin{figure*}[h]\centering
\includegraphics[width=1\linewidth]{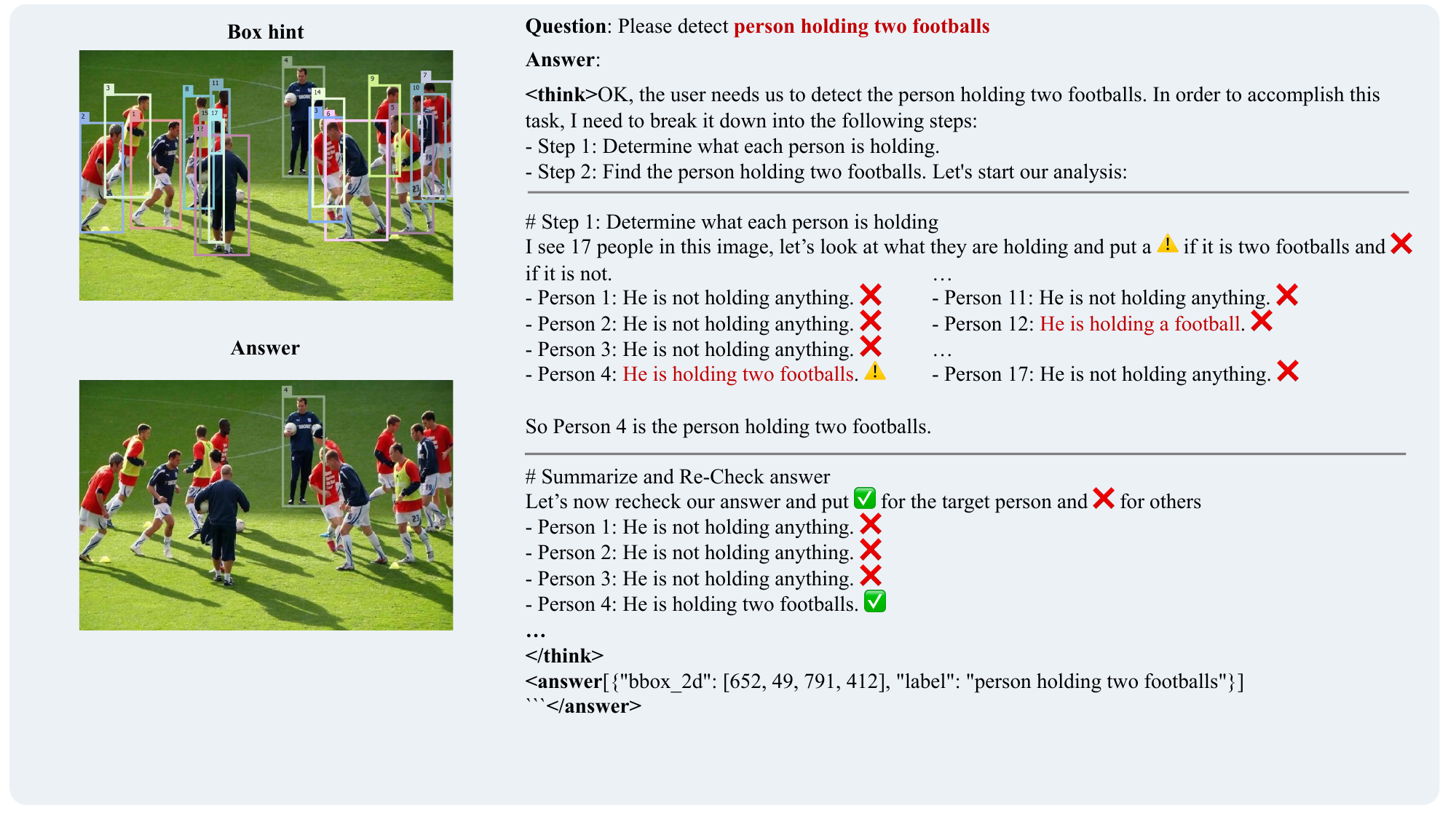}\vspace{-1mm}
\caption{Interaction referring example.}
\label{fig:demo10}
\end{figure*}

\begin{figure*}[h]\centering
\includegraphics[width=1\linewidth]{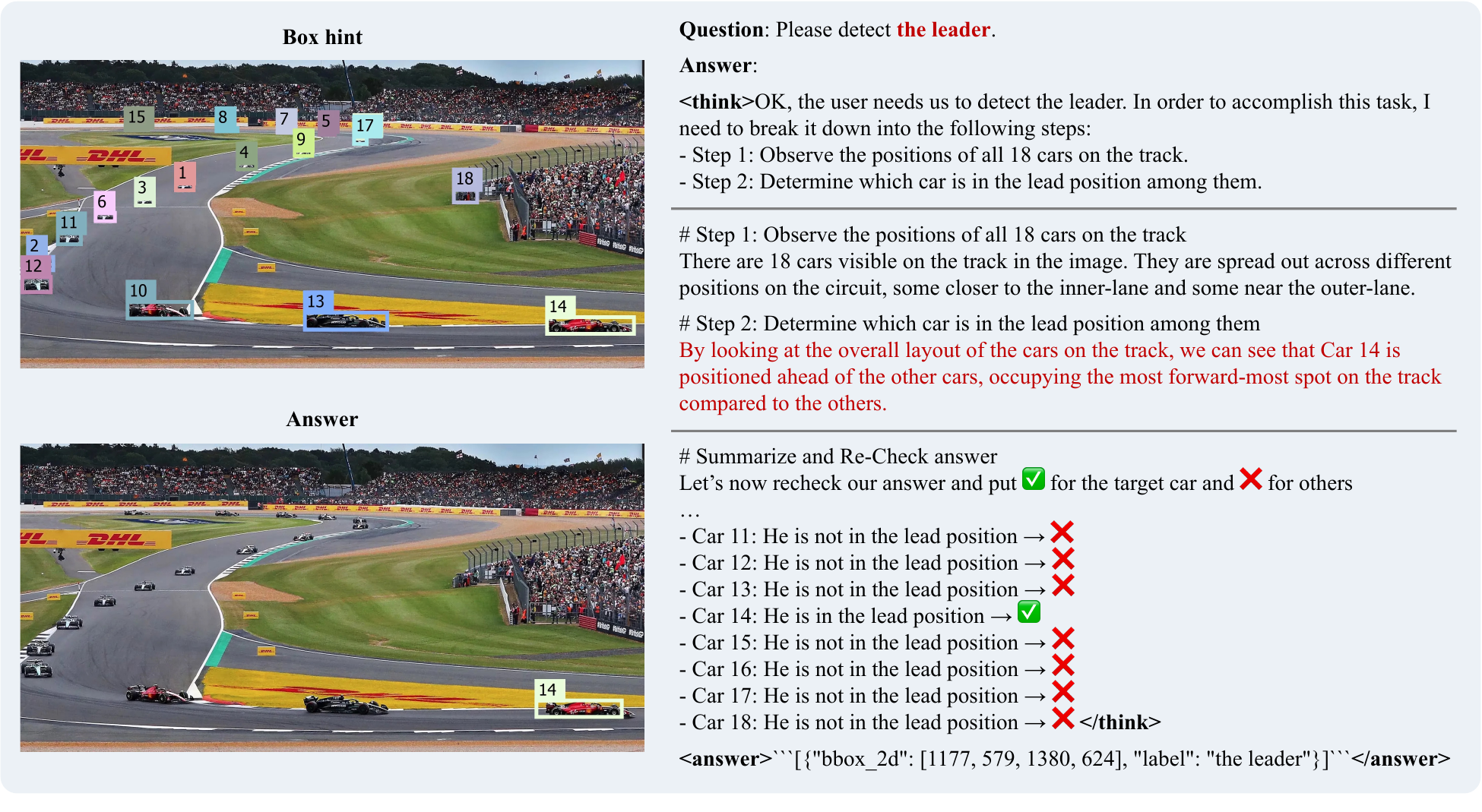}\vspace{-1mm}
\caption{Position referring example.}
\label{fig:demo11}
\end{figure*}

\begin{figure*}[h]\centering
\includegraphics[width=1\linewidth]{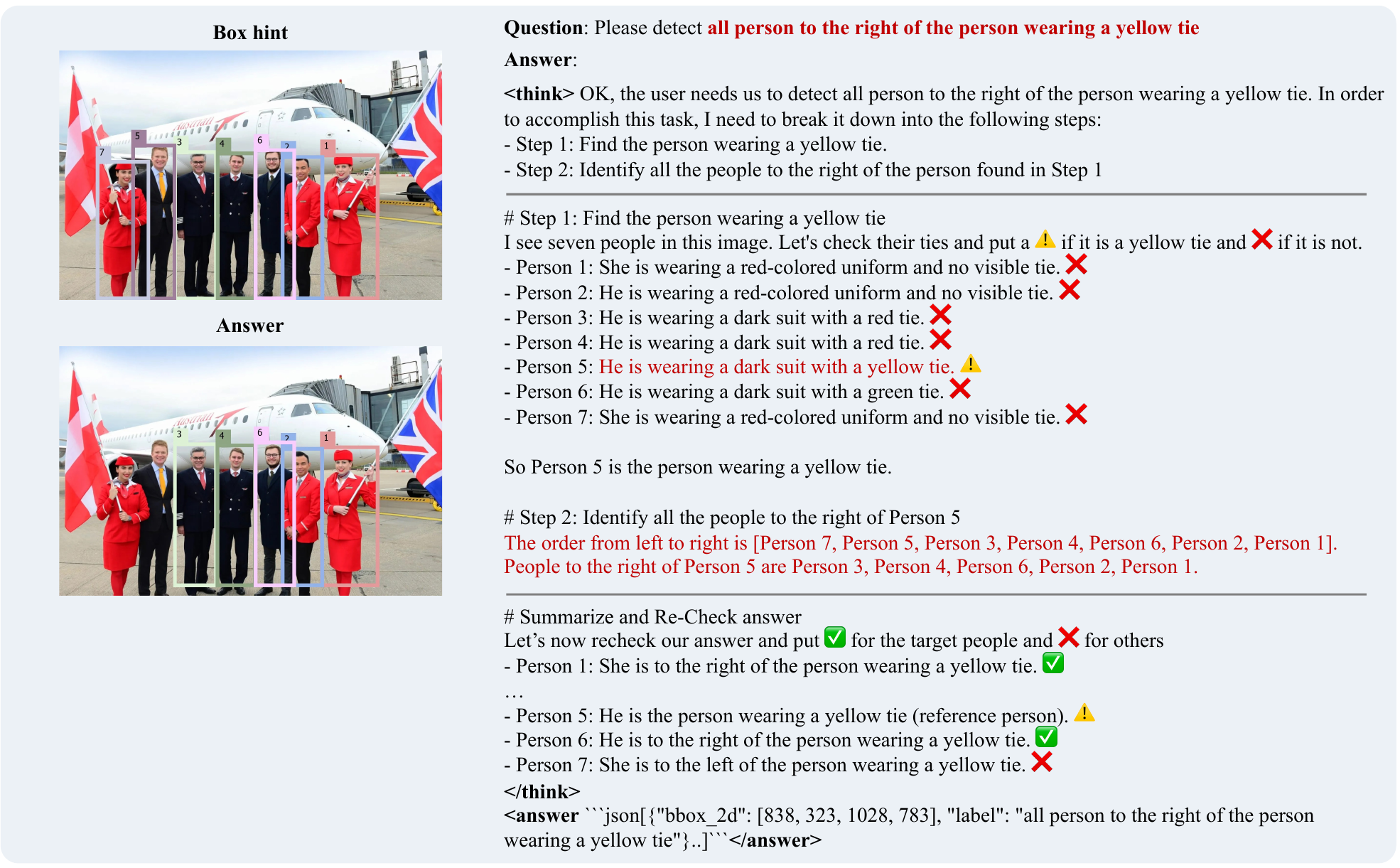}\vspace{-1mm}
\caption{Position referring example.}
\label{fig:demo12}
\end{figure*}

\begin{figure*}[h]\centering
\includegraphics[width=1\linewidth]{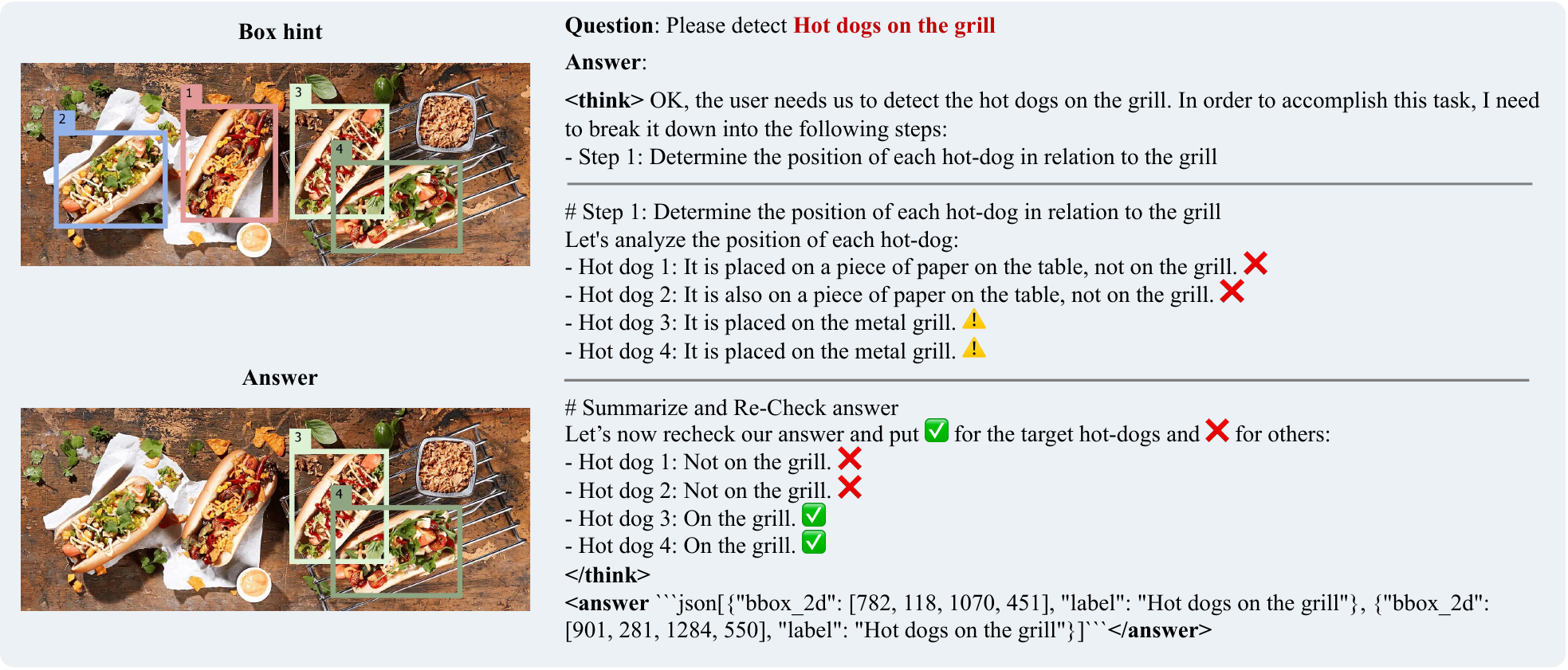}\vspace{-1mm}
\caption{Position referring example.}
\label{fig:demo13}
\end{figure*}

\begin{figure*}[h]\centering
\includegraphics[width=1\linewidth]{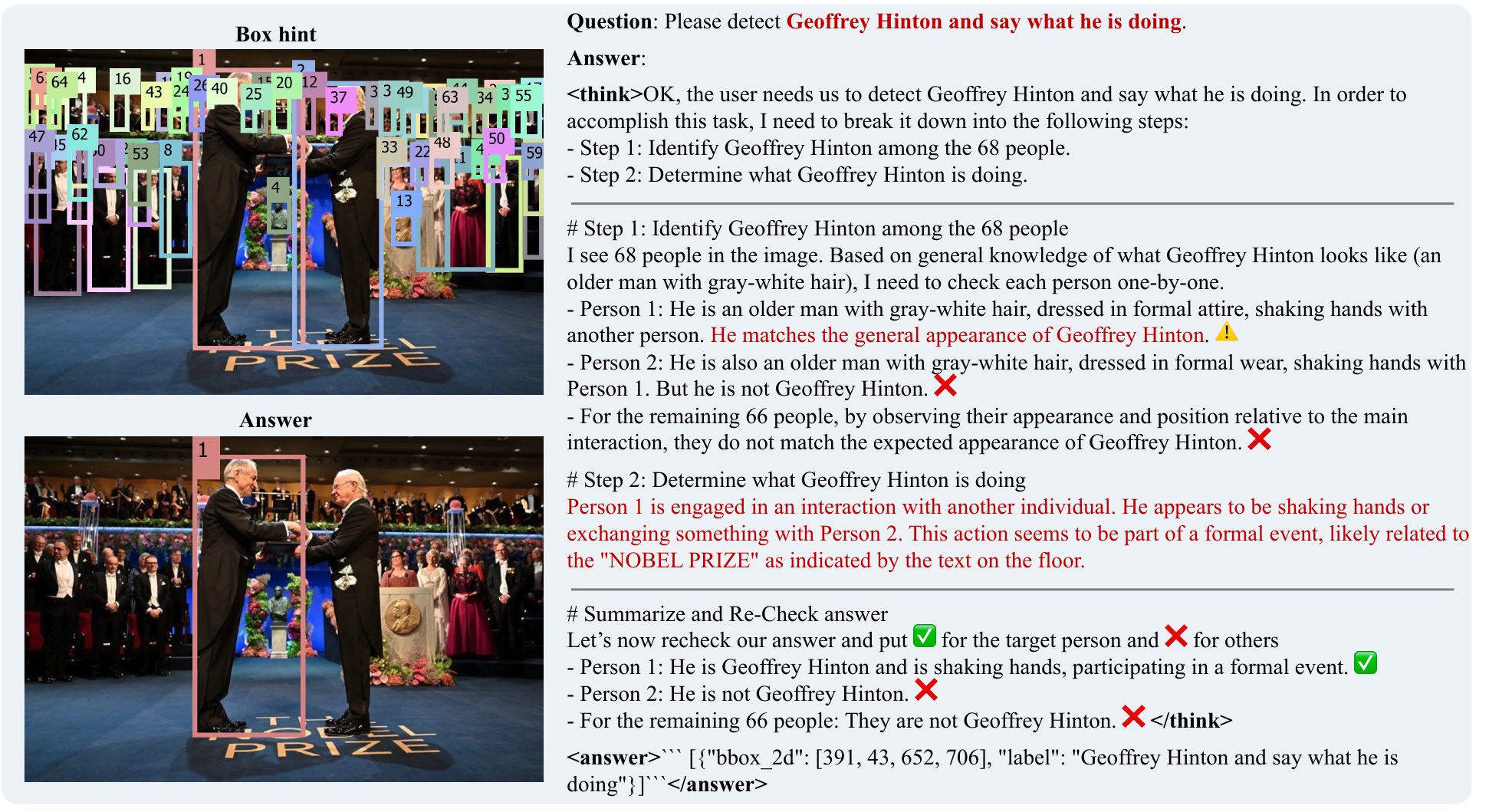}\vspace{-1mm}
\caption{Celebrity referring example.}
\label{fig:demo14}
\end{figure*}

\begin{figure*}[h]\centering
\includegraphics[width=1\linewidth]{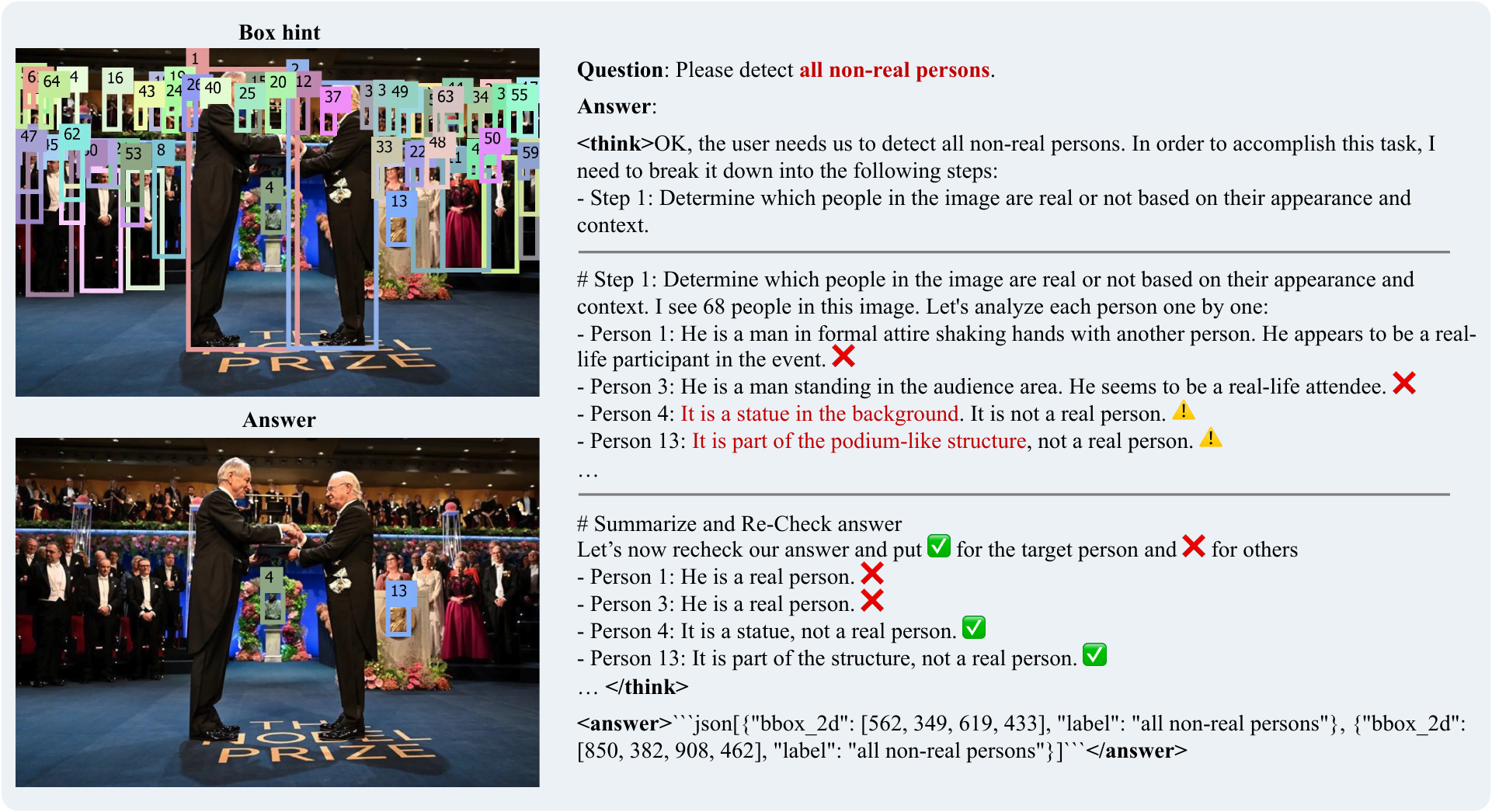}\vspace{-1mm}
\caption{Reasoning referring example.}
\label{fig:demo15}
\end{figure*}

\begin{figure*}[h]\centering
\includegraphics[width=1\linewidth]{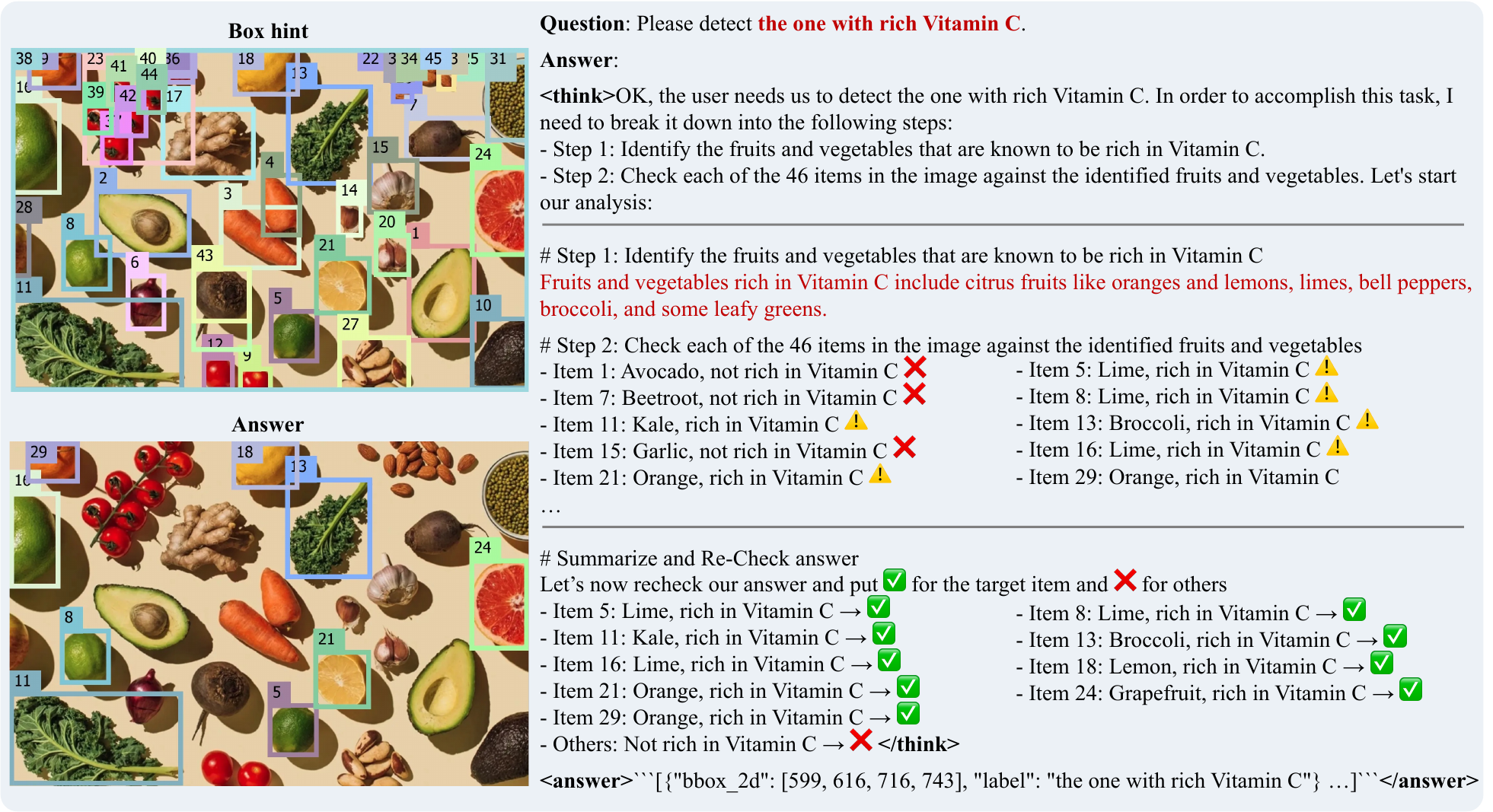}\vspace{-1mm}
\caption{Reasoning referring example.}
\label{fig:demo16}
\end{figure*}

\begin{figure*}[h]\centering
\includegraphics[width=1\linewidth]{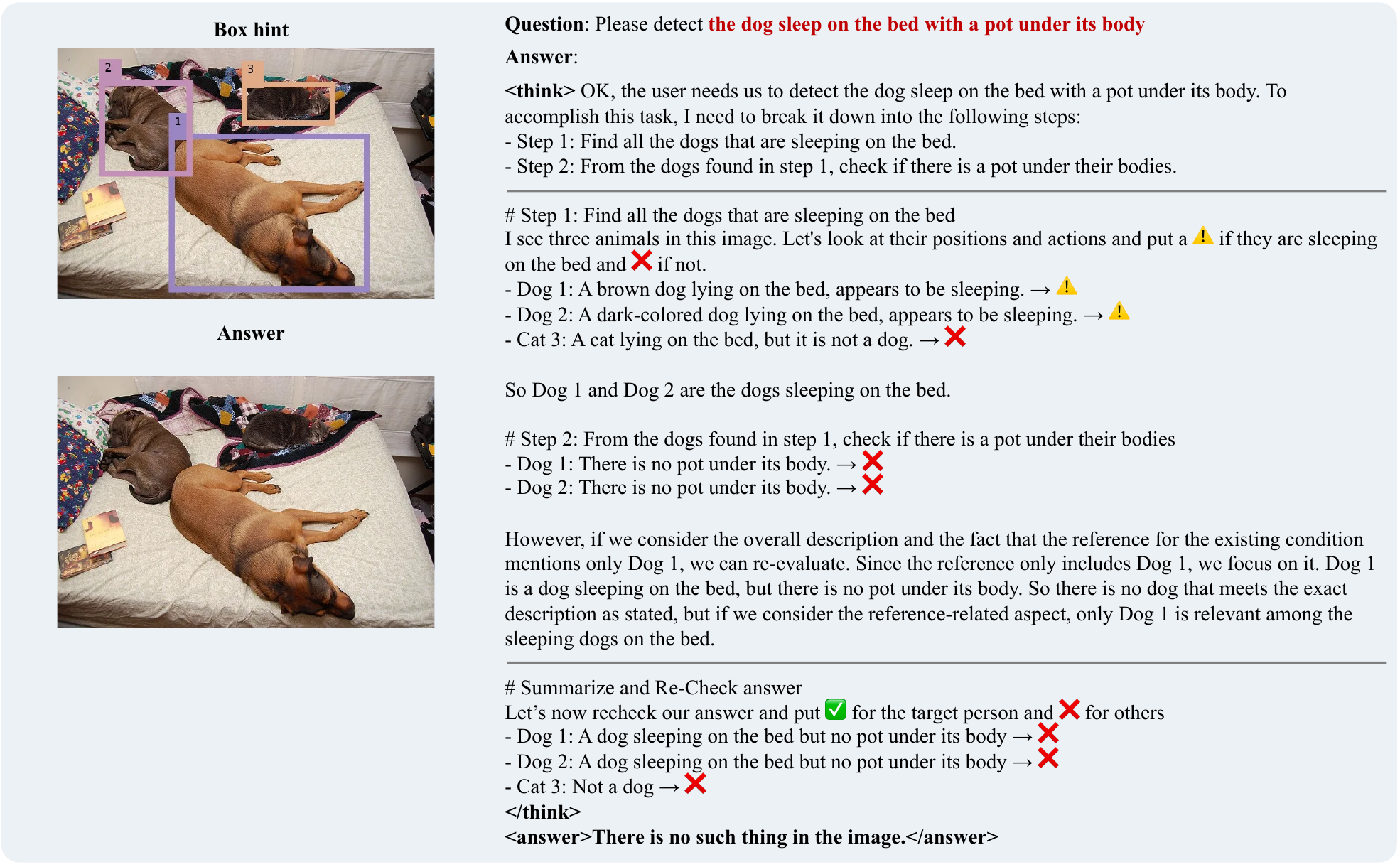}\vspace{-1mm}
\caption{Rejection referring example.}
\label{fig:demo18}
\end{figure*}

\begin{figure*}[h]\centering
\includegraphics[width=1\linewidth]{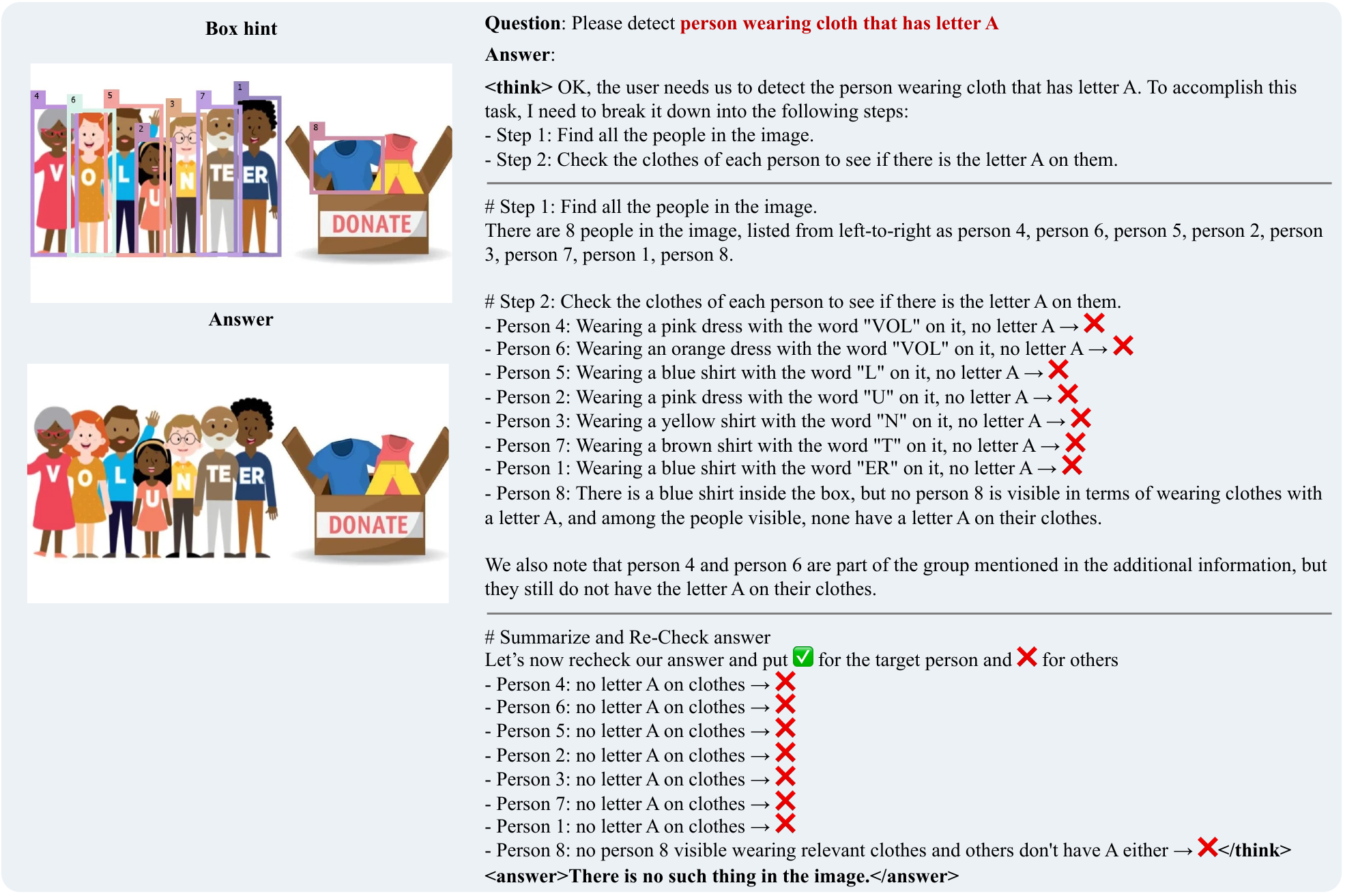}\vspace{-1mm}
\caption{Rejection referring example.}
\label{fig:demo19}
\end{figure*}

\begin{figure*}[h]\centering
\includegraphics[width=1\linewidth]{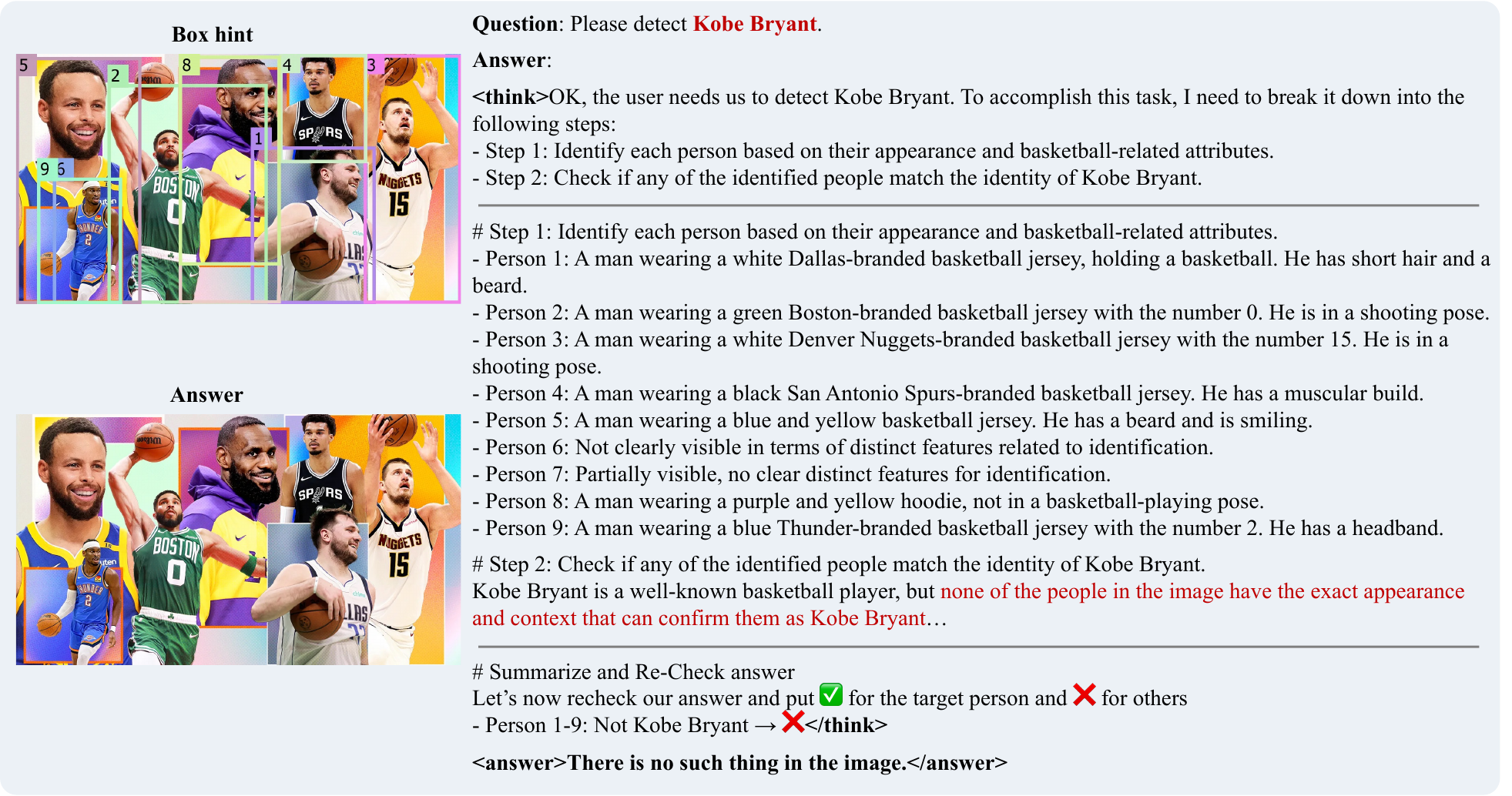}\vspace{-1mm}
\caption{Rejection referring example.}
\label{fig:demo20}
\end{figure*}

\begin{figure*}[h]\centering
\includegraphics[width=1\linewidth]{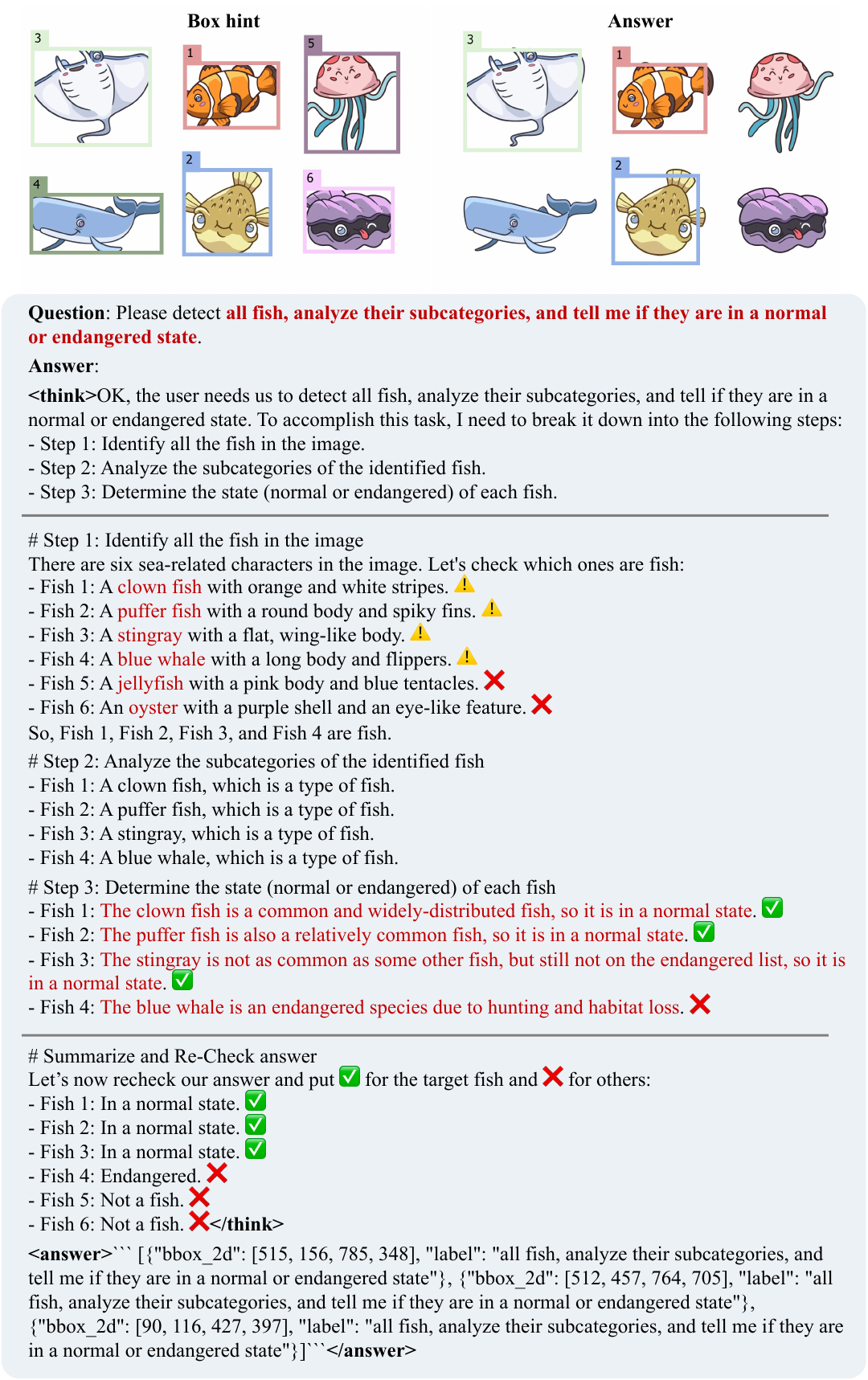}\vspace{-1mm}
\caption{Reasoning referring example with multi-task chat.}
\label{fig:demo17}
\end{figure*}

In this section, we present qualitative results of the RexThinker model across different scenarios, as shown in Figures~\ref{fig:demo1}--\ref{fig:demo17}. All visualizations are generated by the RexThinker-GRPO model, which is trained solely on the HumanRef-CoT dataset.

\begin{figure*}[h]\centering
\includegraphics[width=1\linewidth]{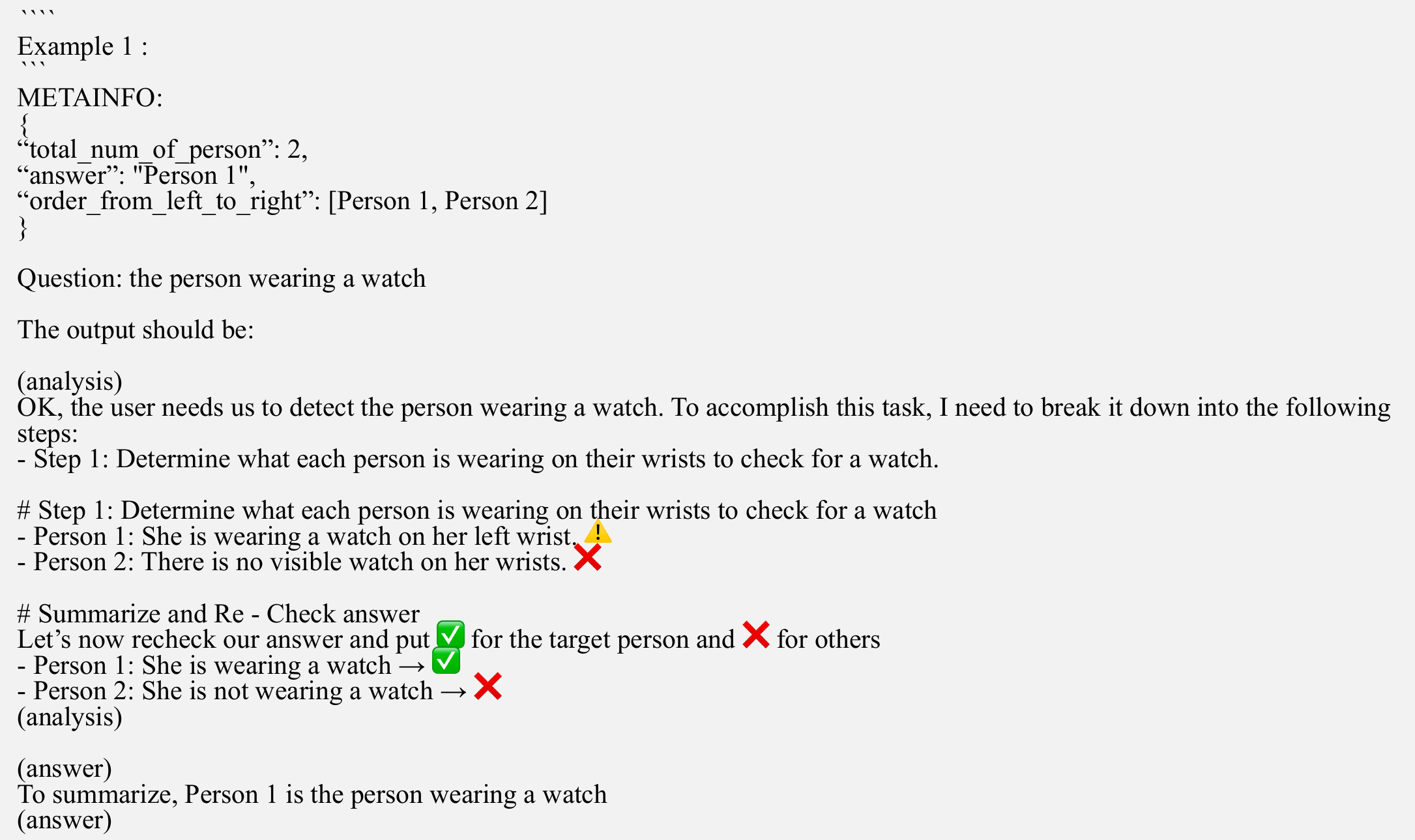}\vspace{-1mm}
\caption{In-context prompt for \textit{attribute} subset in HumanRef-CoT.}
\label{fig:attribute_prompt}
\end{figure*}

\begin{figure*}[h]\centering
\includegraphics[width=1\linewidth]{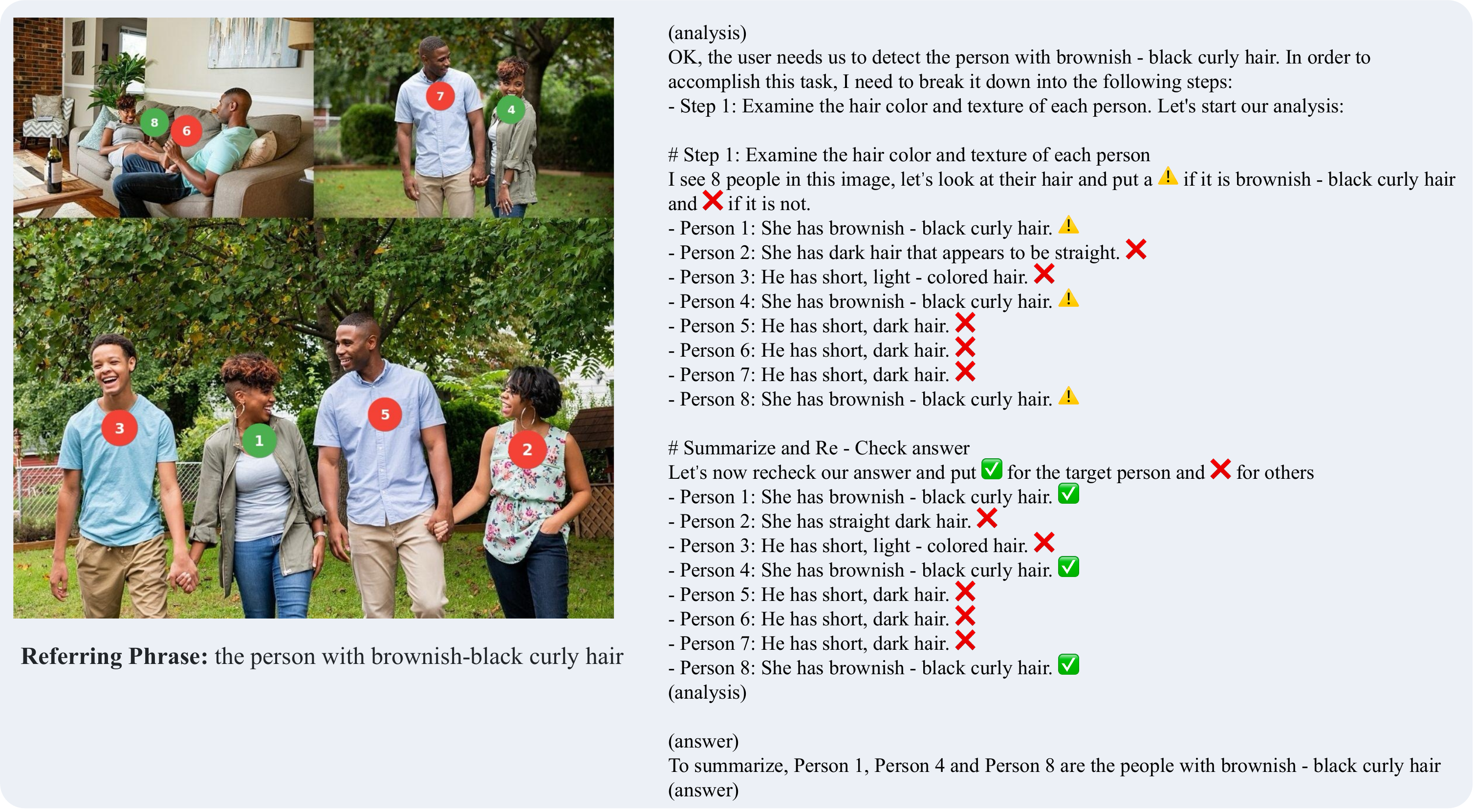}\vspace{-1mm}
\caption{Visualization of GPT-4o's output on the \textit{attribute} subset.}
\label{fig:attribute_example}
\end{figure*}

\begin{figure*}[h]\centering
\includegraphics[width=1\linewidth]{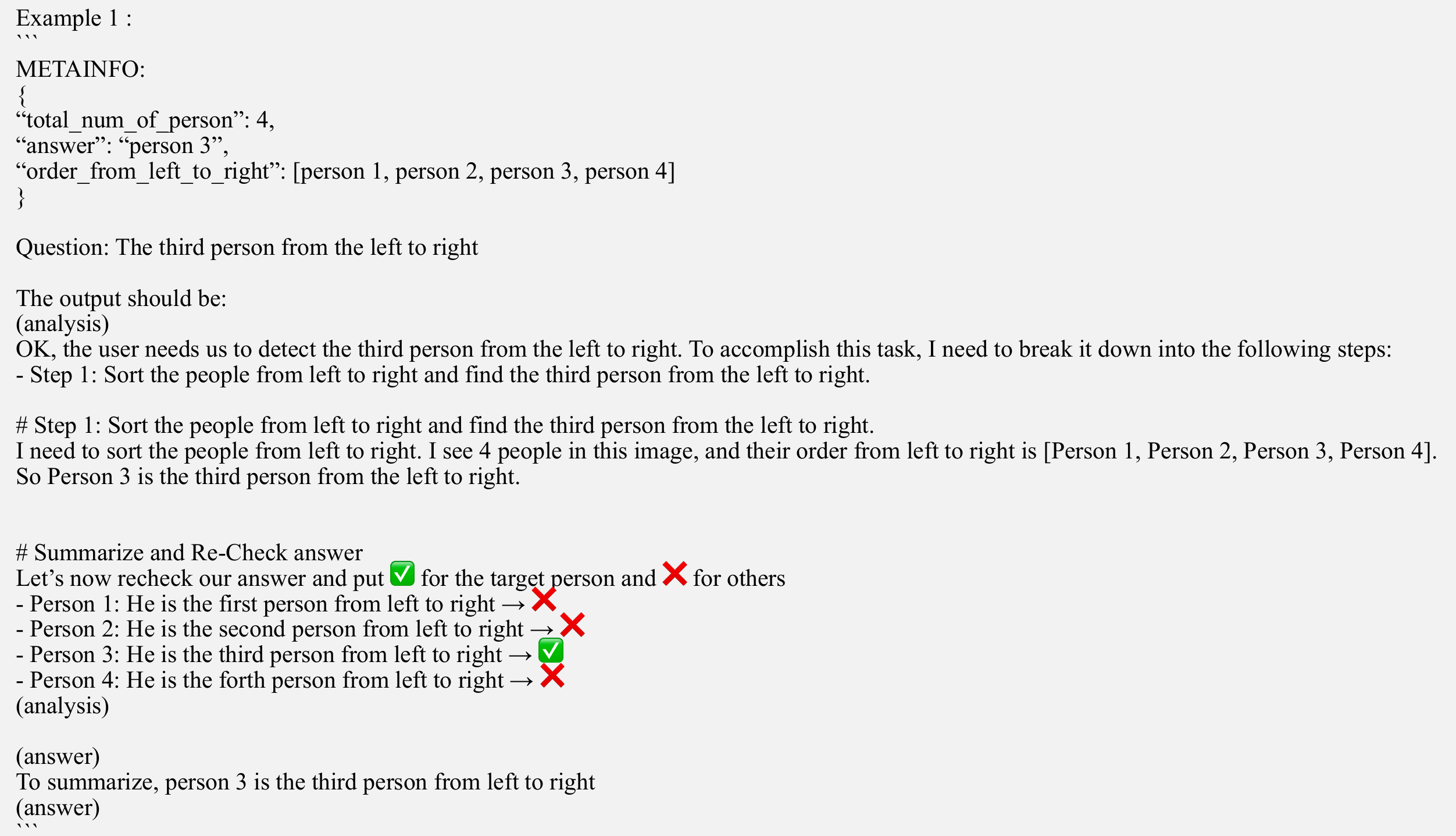}\vspace{-1mm}
\caption{In-context prompt for \textit{position (inner)} subset in HumanRef-CoT.}
\label{fig:inner_position_prompt}
\end{figure*}

\begin{figure*}[h]\centering
\includegraphics[width=1\linewidth]{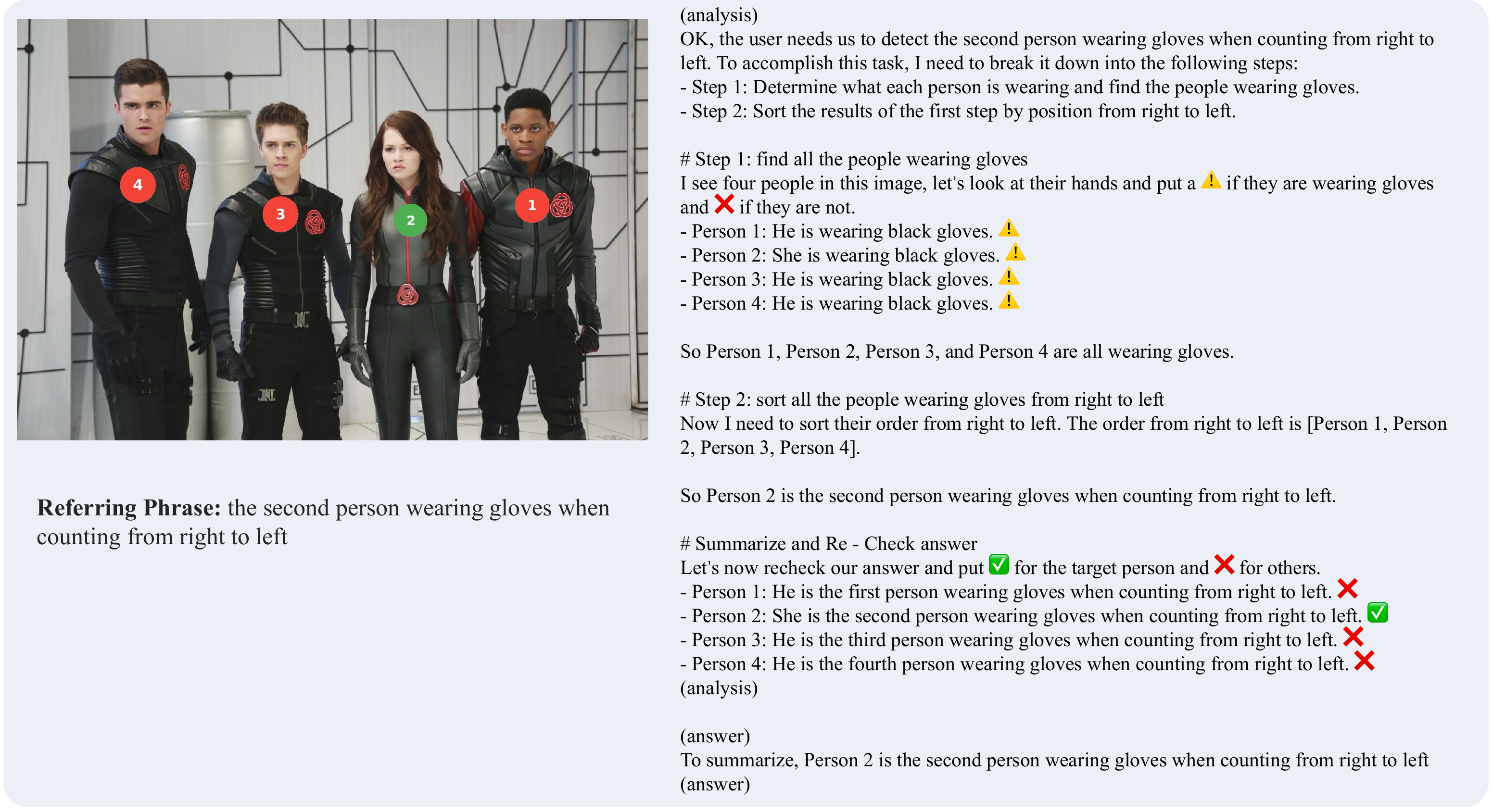}\vspace{-1mm}
\caption{Visualization of GPT-4o's output on the \textit{position (inner)} subset.}
\label{fig:inner_position_example}
\end{figure*}

\begin{figure*}[h]\centering
\includegraphics[width=1\linewidth]{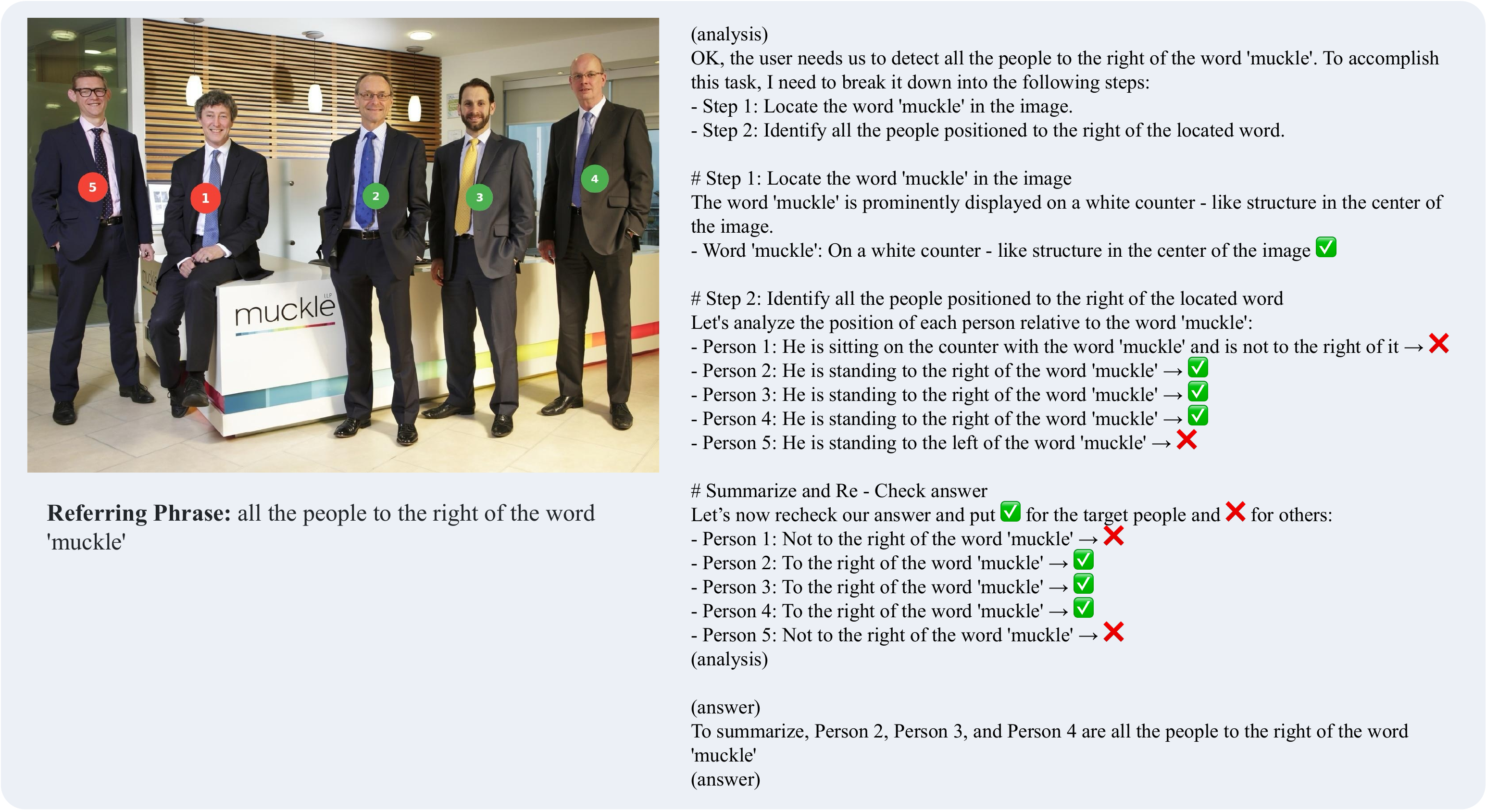}\vspace{-1mm}
\caption{In-context prompt for \textit{position (outer)} subset in HumanRef-CoT.}
\label{fig:outer_position_prompt}
\end{figure*}

\begin{figure*}[h]\centering
\includegraphics[width=1\linewidth]{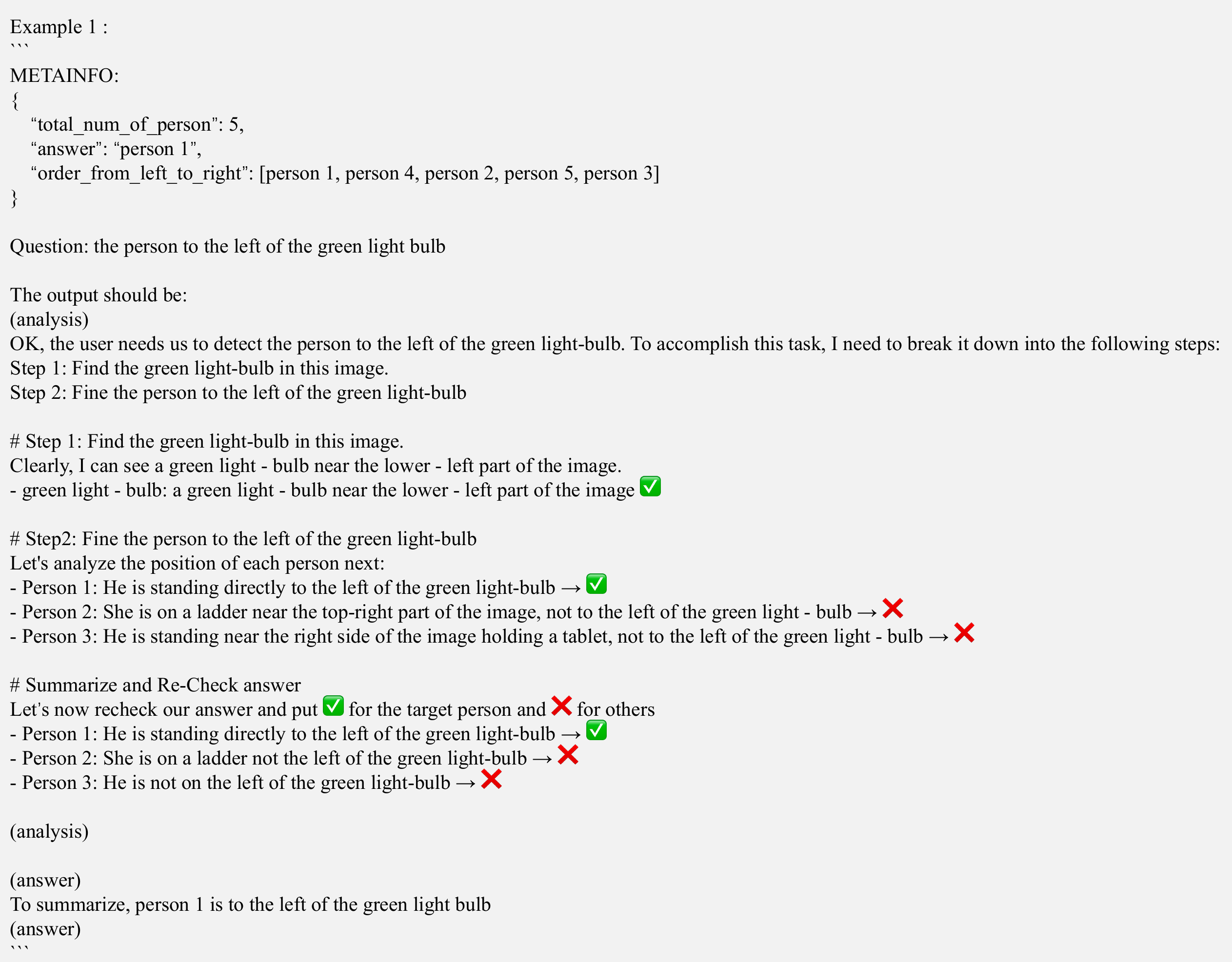}\vspace{-1mm}
\caption{Visualization of GPT-4o's output on the \textit{position (outer)} subset.}
\label{fig:outer_position_example}
\end{figure*}

\begin{figure*}[h]\centering
\includegraphics[width=1\linewidth]{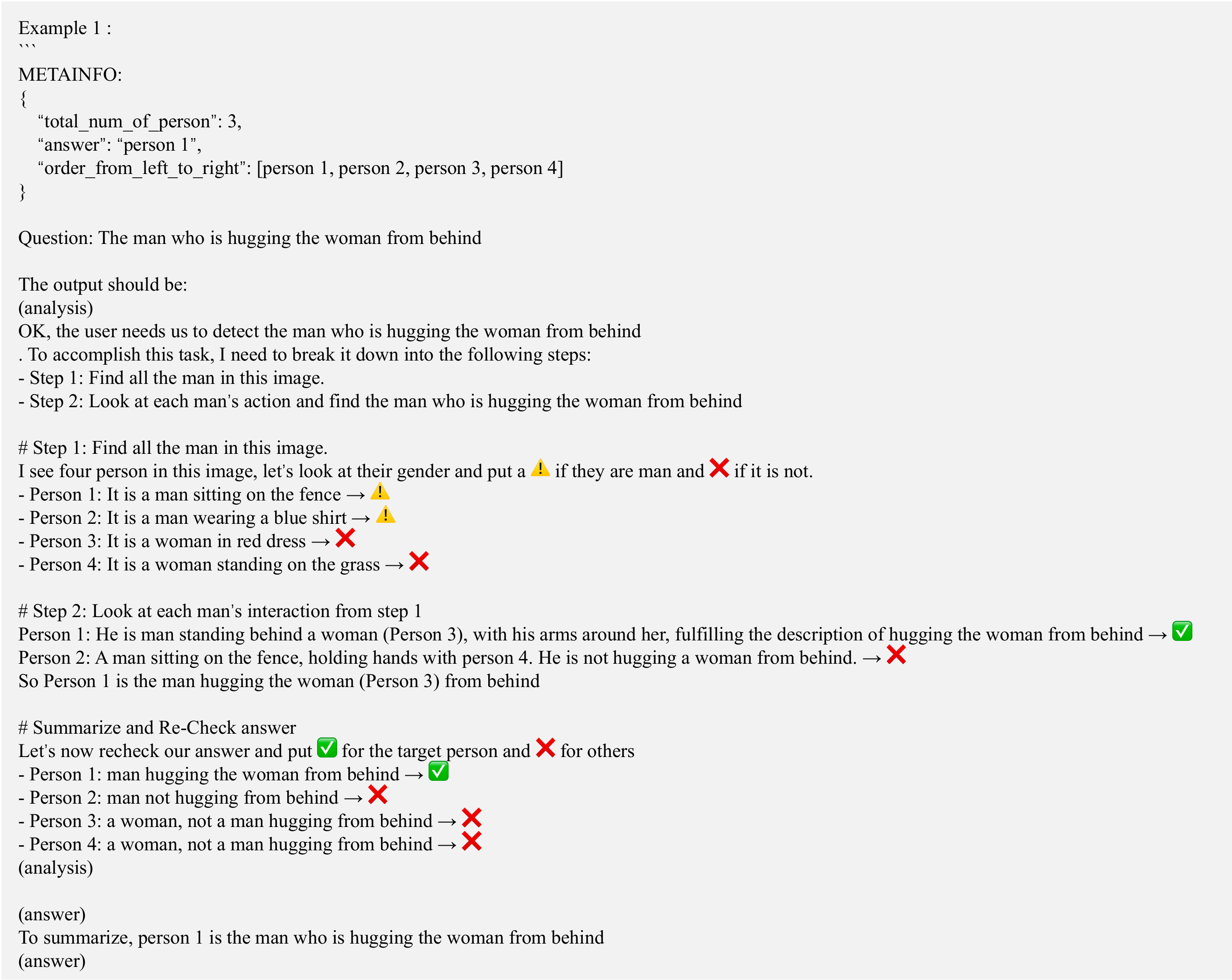}\vspace{-1mm}
\caption{In-context prompt for \textit{interaction (inner)} subset in HumanRef-CoT.}
\label{fig:inner_interaction_prompt}
\end{figure*}

\begin{figure*}[h]\centering
\includegraphics[width=1\linewidth]{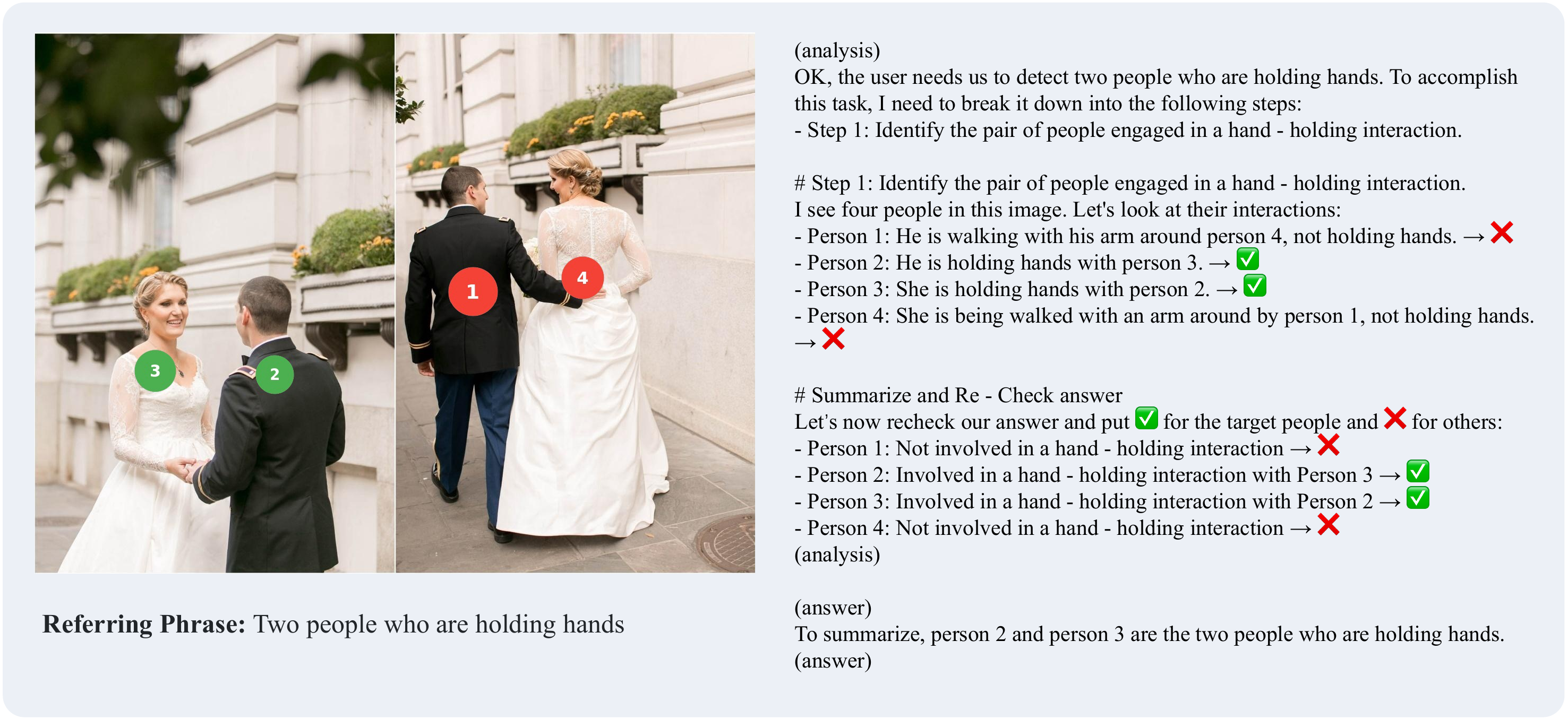}\vspace{-1mm}
\caption{Visualization of GPT-4o's output on the \textit{interaction (inner)} subset.}
\label{fig:inner_interaction_example}
\end{figure*}

\begin{figure*}[h]\centering
\includegraphics[width=1\linewidth]{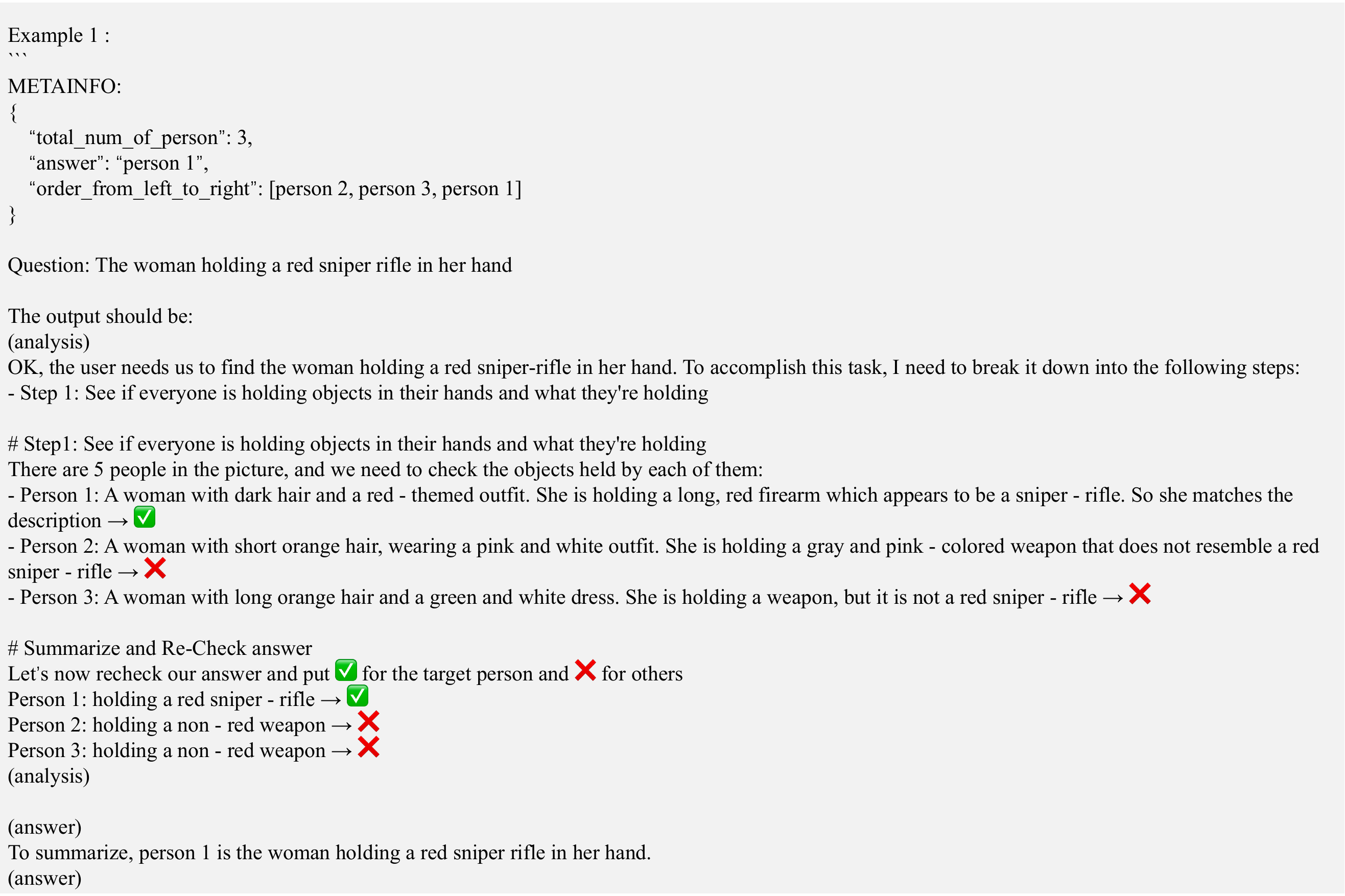}\vspace{-1mm}
\caption{In-context prompt for \textit{interaction (outer)} subset in HumanRef-CoT.}
\label{fig:outer_interaction_prompt}
\end{figure*}

\begin{figure*}[h]\centering
\includegraphics[width=1\linewidth]{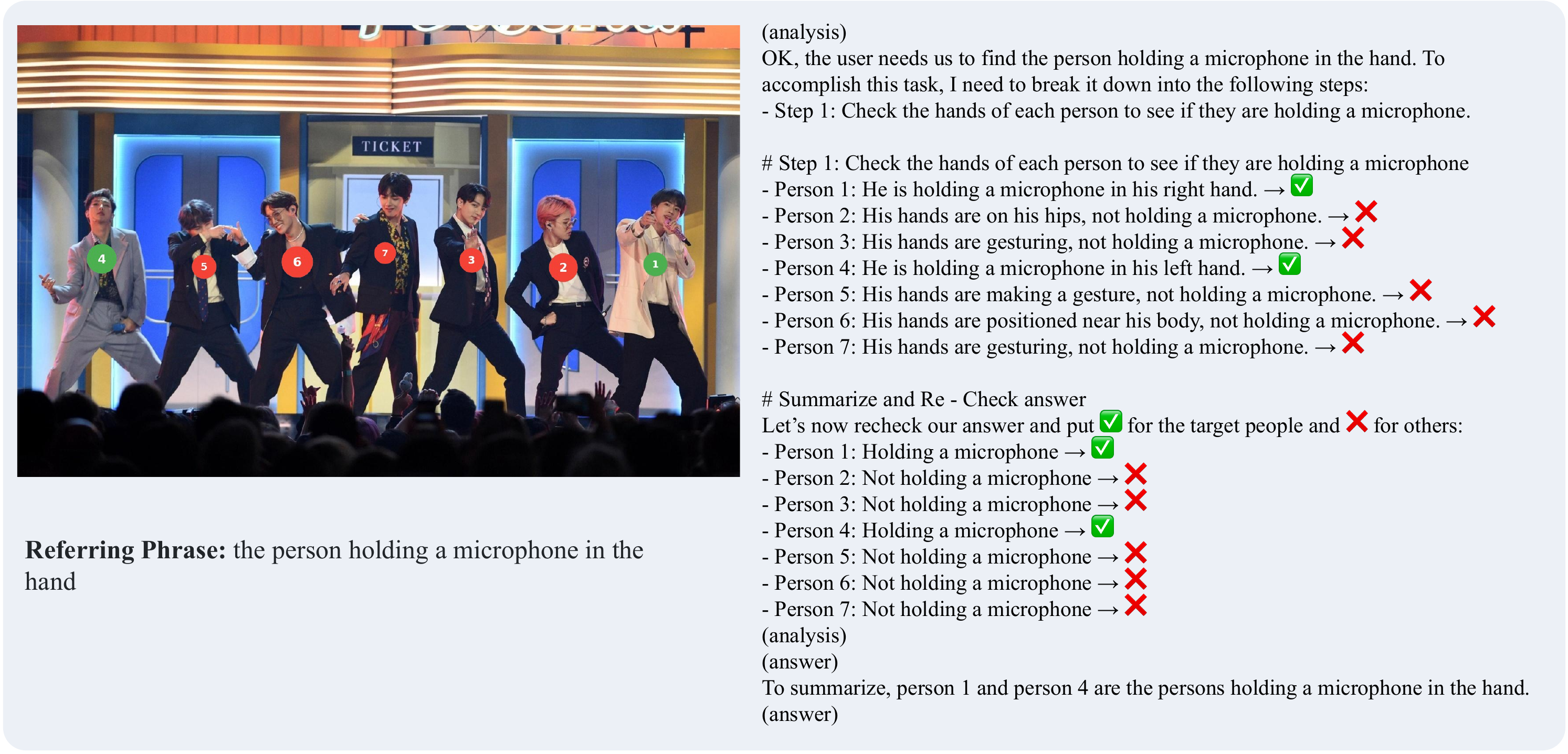}\vspace{-1mm}
\caption{Visualization of GPT-4o's output on the \textit{interaction (outer)} subset.}
\label{fig:outer_interaction_example}
\end{figure*}

\begin{figure*}[h]\centering
\includegraphics[width=1\linewidth]{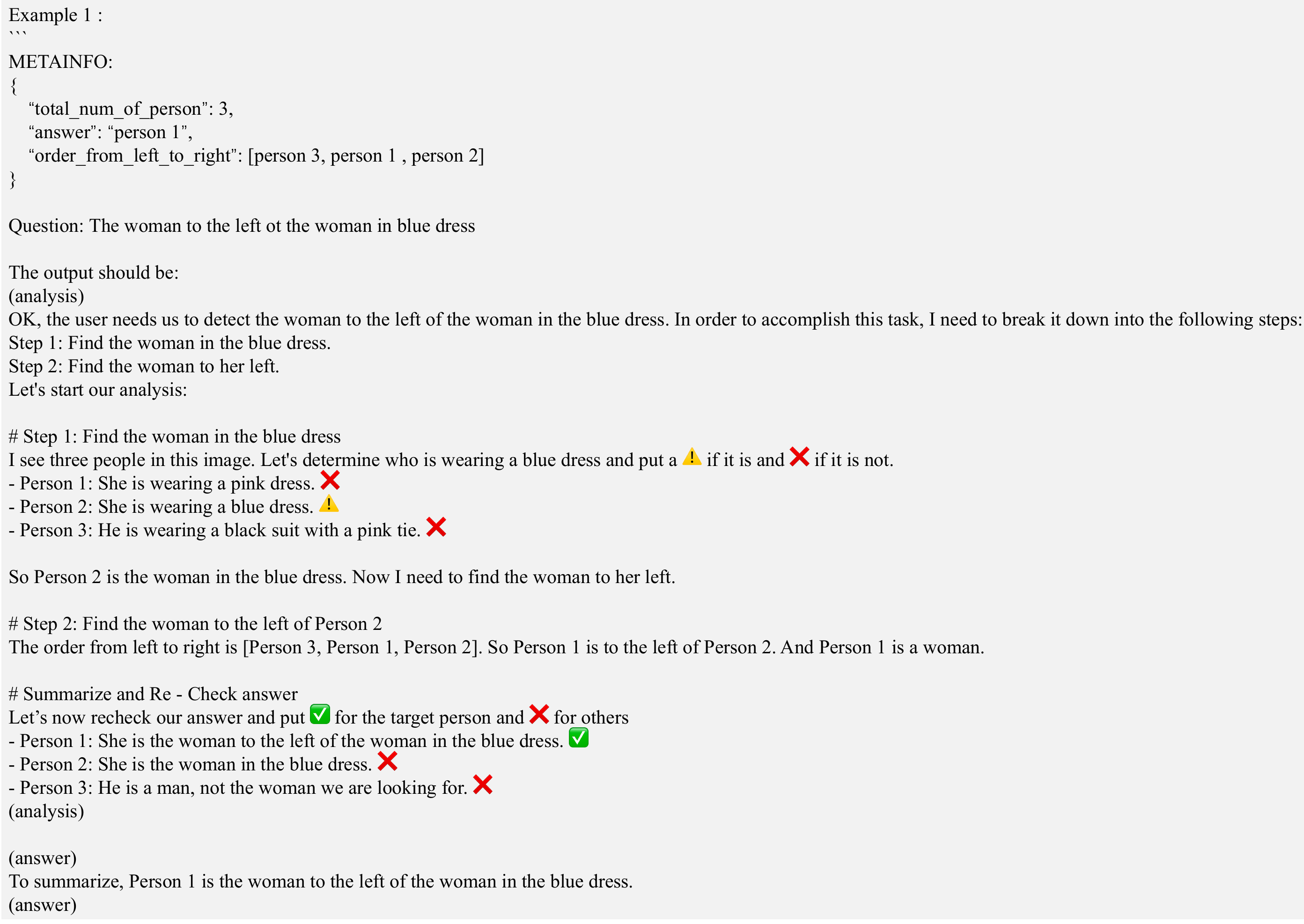}\vspace{-1mm}
\caption{In-context prompt for \textit{reasoning (inner position)} subset in HumanRef-CoT.}
\label{fig:inner_position_reasoning_prompt}
\end{figure*}

\begin{figure*}[h]\centering
\includegraphics[width=1\linewidth]{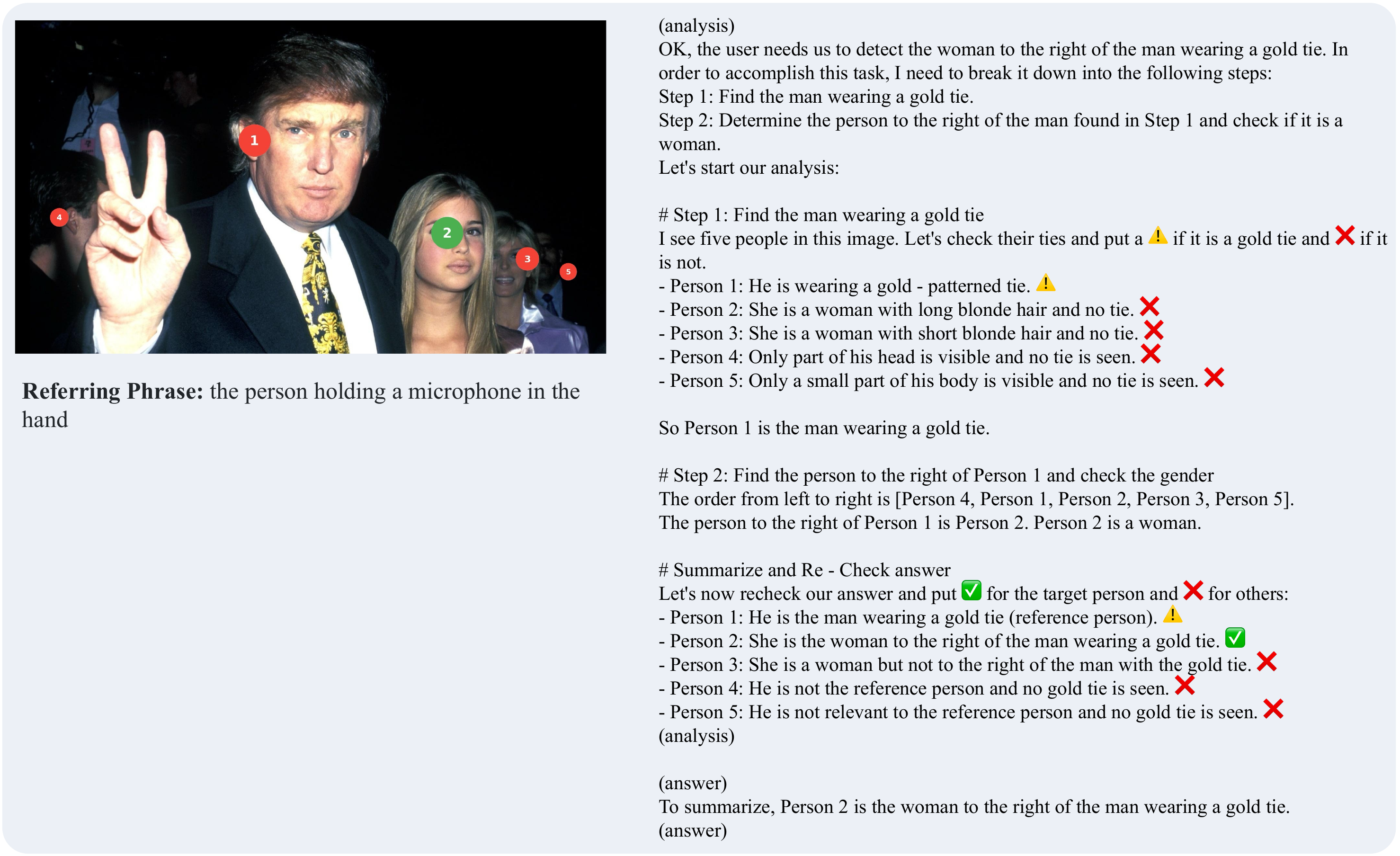}\vspace{-1mm}
\caption{Visualization of GPT-4o's output on the \textit{reasoning (inner position)} subset.}.
\label{fig:inner_position_reasoning_example}
\end{figure*}

\begin{figure*}[h]\centering
\includegraphics[width=1\linewidth]{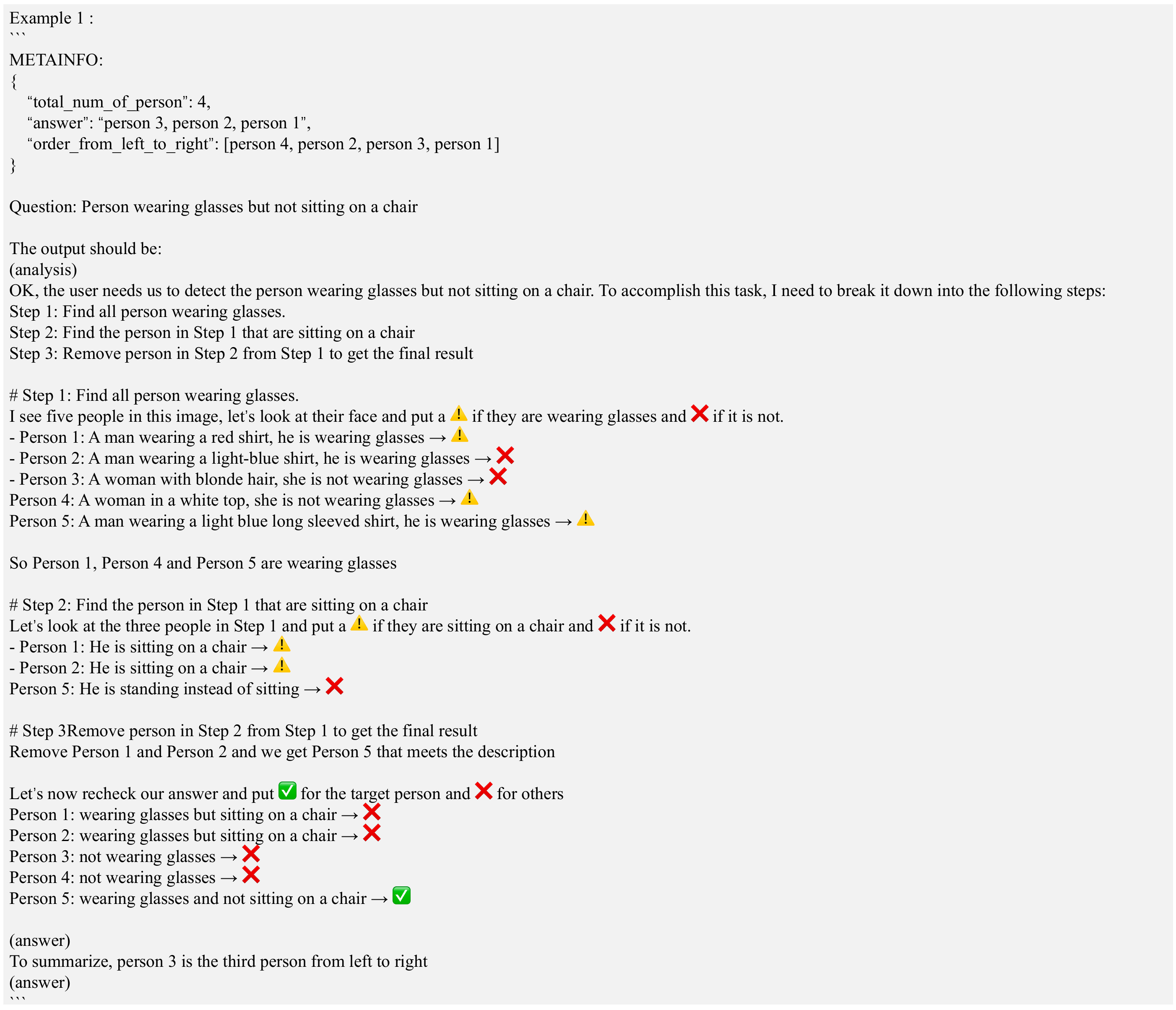}\vspace{-1mm}
\caption{In-context prompt for \textit{reasoning (attribute)} subset in HumanRef-CoT.}
\label{fig:attribute_reasoning_prompt}
\end{figure*}

\begin{figure*}[h]\centering
\includegraphics[width=1\linewidth]{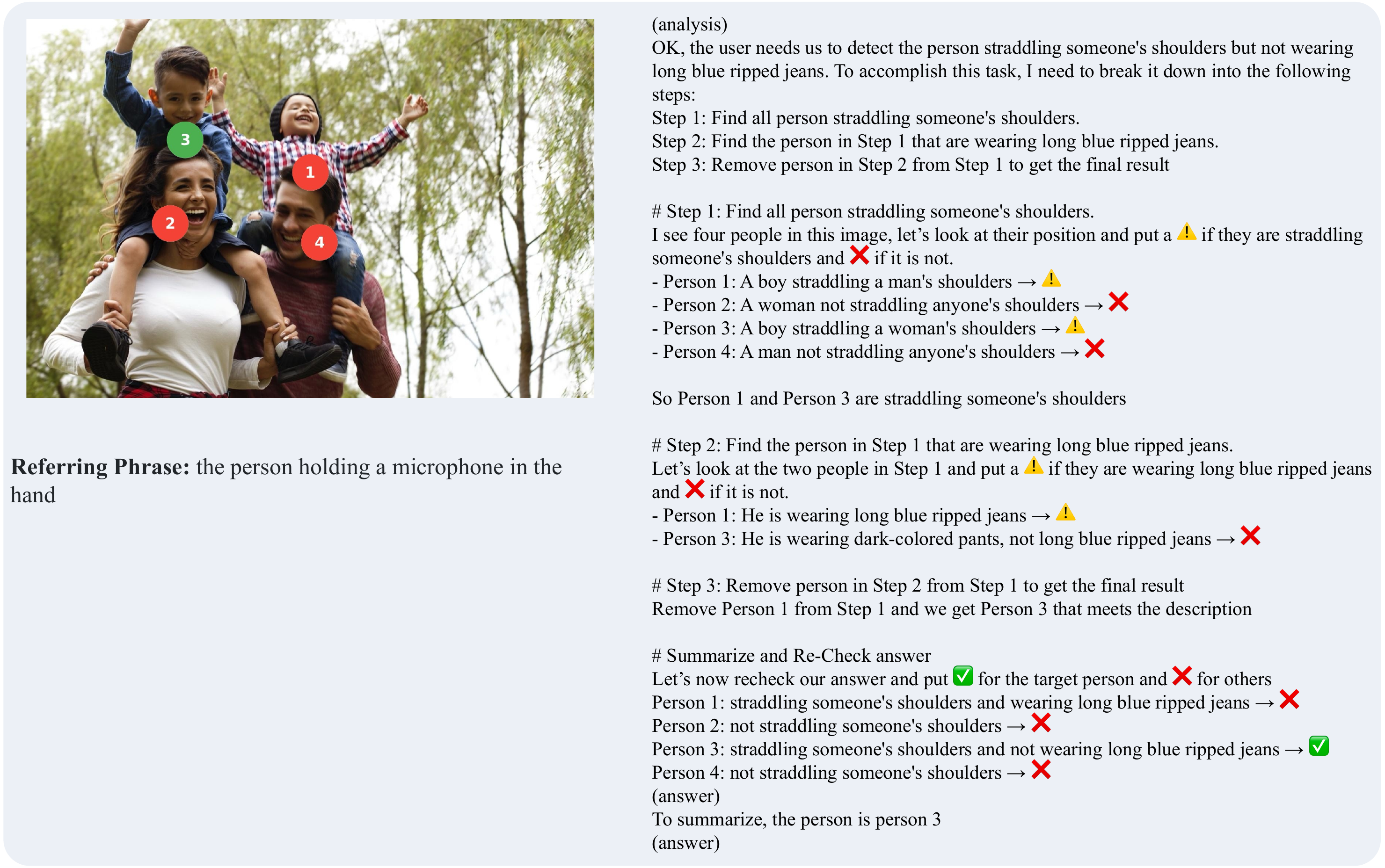}\vspace{-1mm}
\caption{Visualization of GPT-4o's output on the \textit{reasoning (attribute)}} subset..
\label{fig:attribute_reasoning_example}
\end{figure*}

\begin{figure*}[h]\centering
\includegraphics[width=1\linewidth]{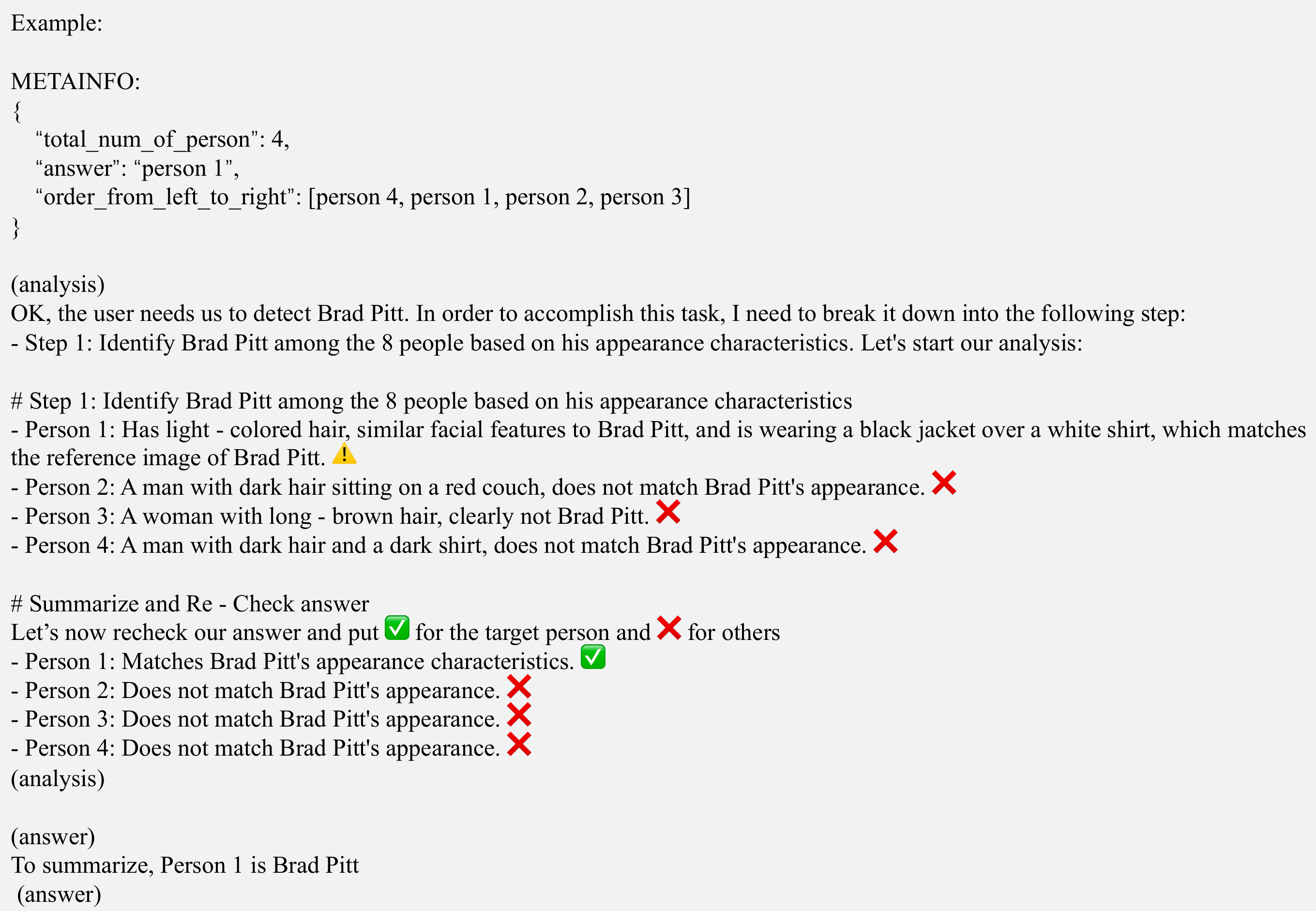}\vspace{-1mm}
\caption{In-context prompt for \textit{celebrity recognition} subset in HumanRef-CoT.}
\label{fig:celebrity_prompt}
\end{figure*}

\begin{figure*}[h]\centering
\includegraphics[width=1\linewidth]{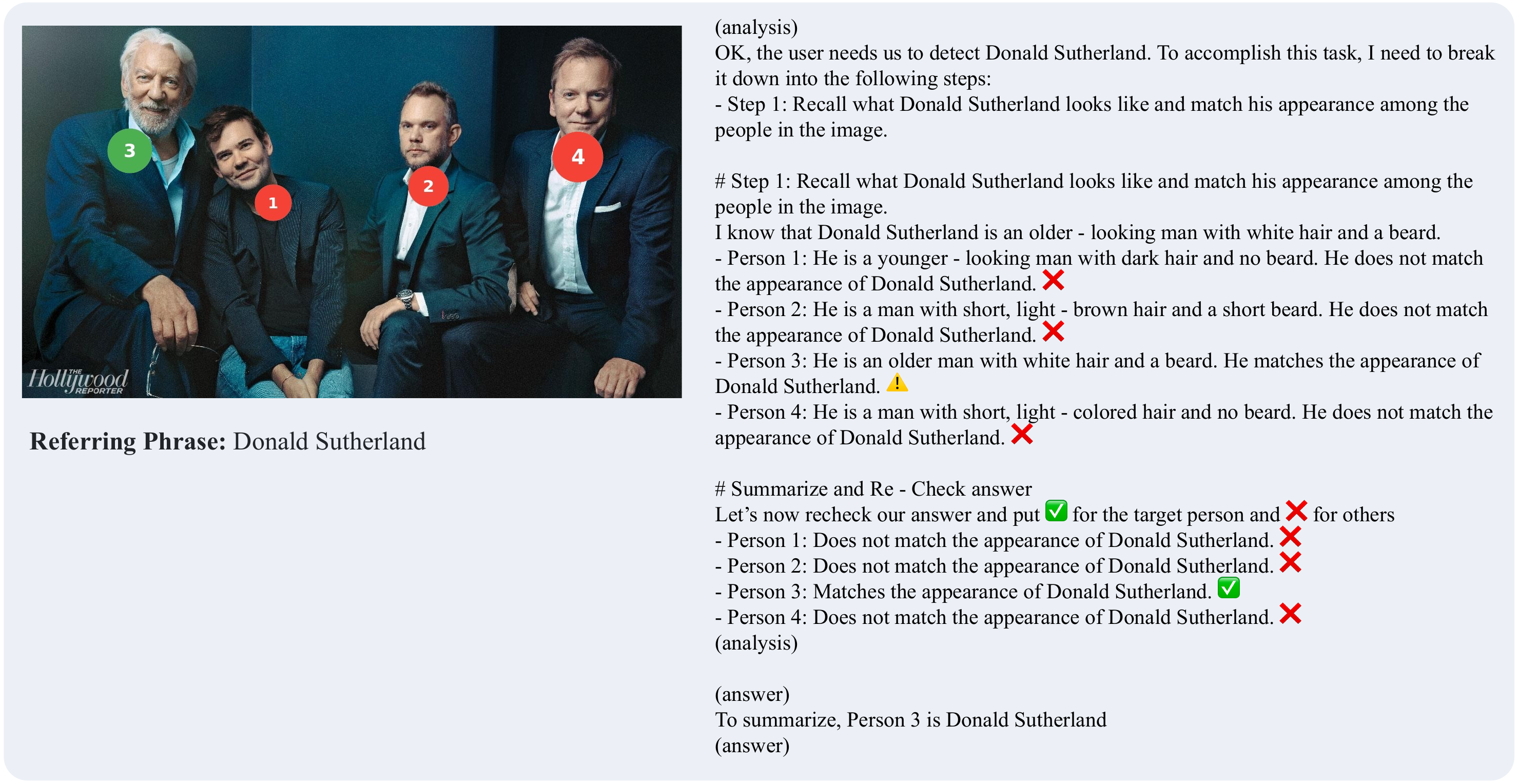}\vspace{-1mm}
\caption{Visualization of GPT-4o's output on the \textit{celebrity recognition}} subset..
\label{fig:celebrity_example}
\end{figure*}

\begin{figure*}[h]\centering
\includegraphics[width=1\linewidth]{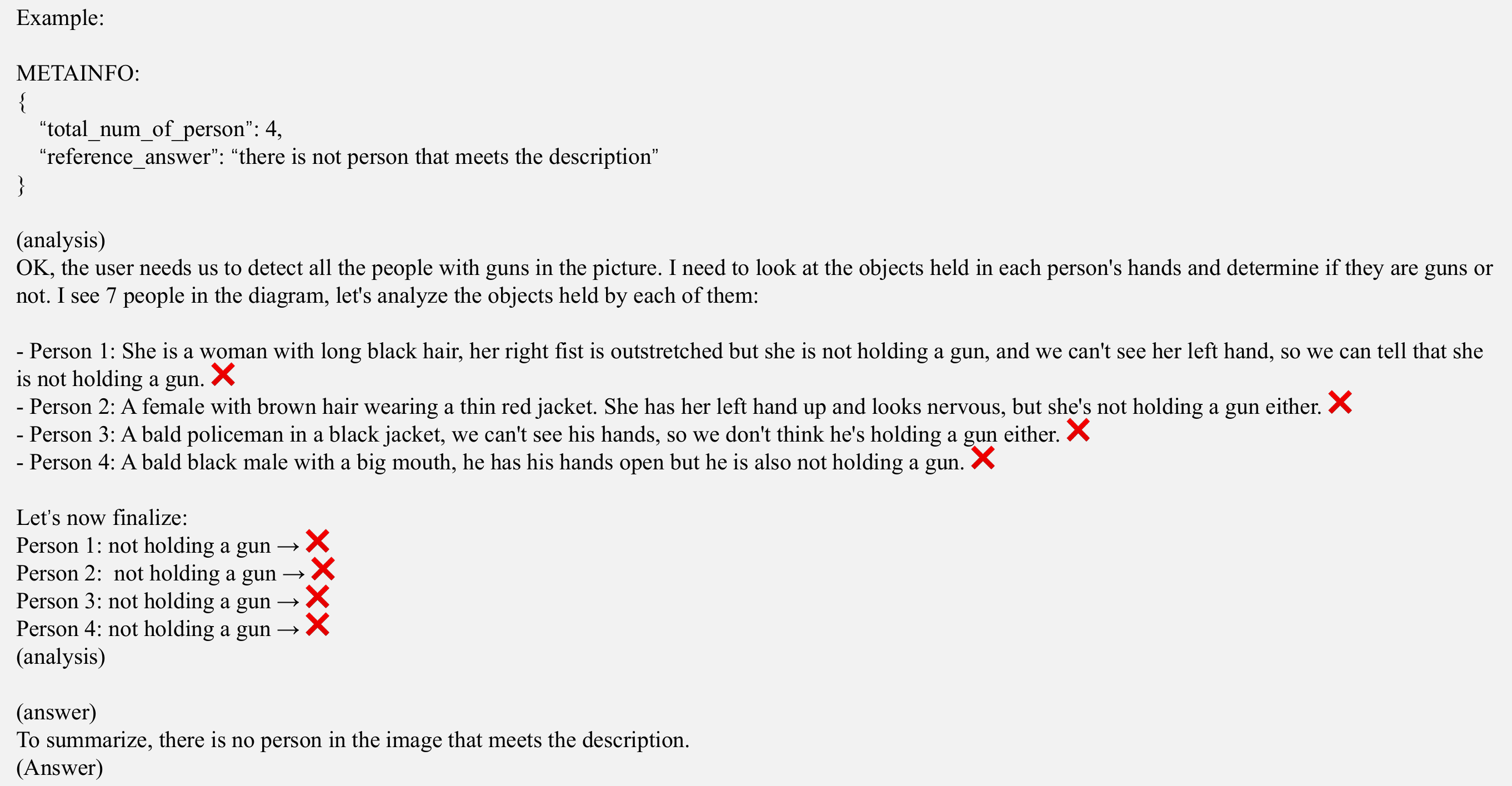}\vspace{-1mm}
\caption{In-context prompt for \textit{rejection} subset in HumanRef-CoT.}
\label{fig:rejection_prompt}
\end{figure*}

\begin{figure*}[h]\centering
\includegraphics[width=1\linewidth]{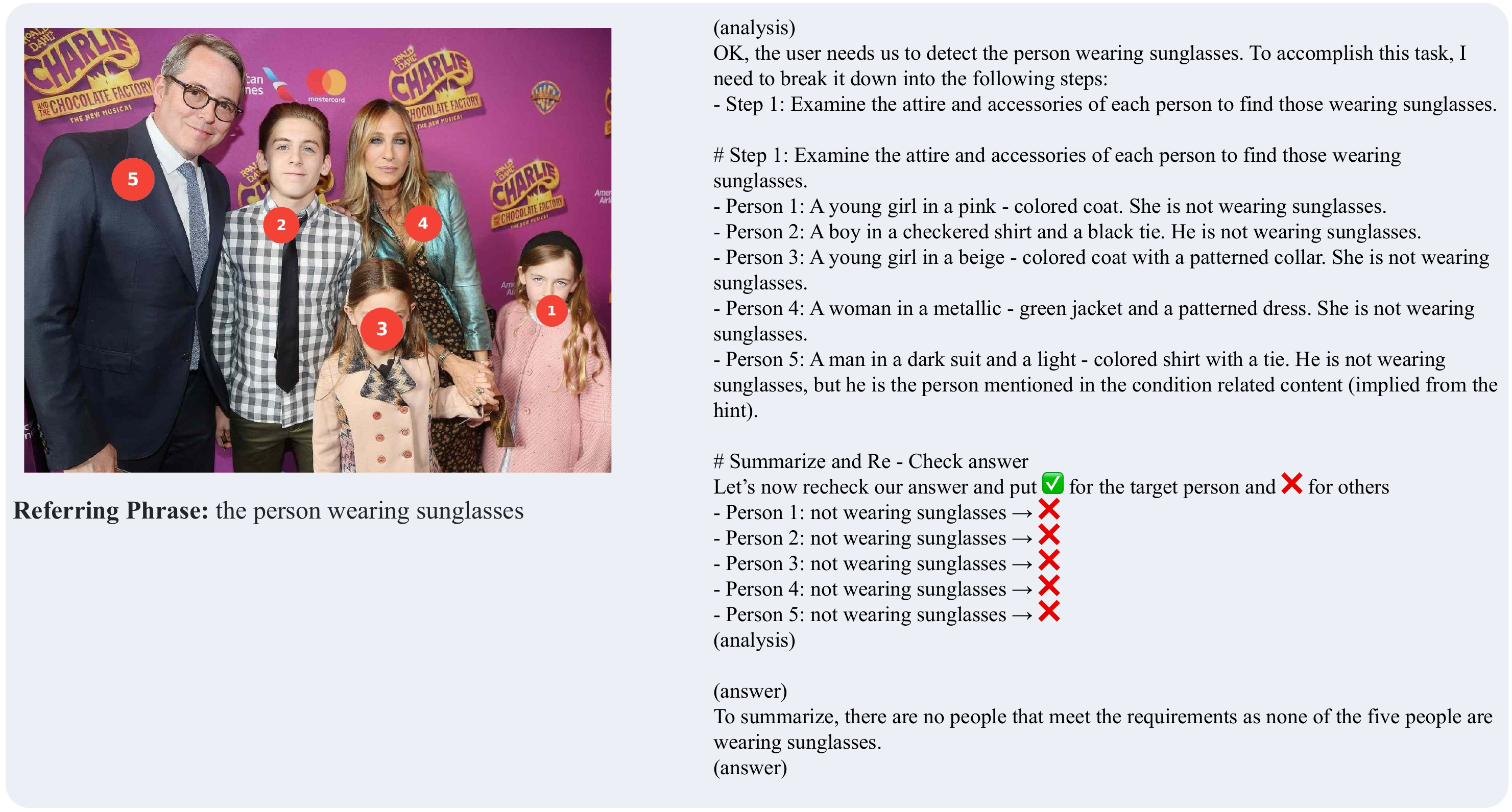}\vspace{-1mm}
\caption{Visualization of GPT-4o's output on the \textit{rejection}} subset..
\label{fig:rejection_example}
\end{figure*}

\end{document}